\pdfoutput=1
\documentclass[10pt,twocolumn,letterpaper]{article}

\usepackage{iccv}
\usepackage{times}
\usepackage{epsfig}
\usepackage{graphicx}
\usepackage{amsmath}
\usepackage{amssymb}

\usepackage[pagebackref=true,breaklinks=true,letterpaper=true,colorlinks,bookmarks=false]{hyperref}

\iccvfinalcopy 

\usepackage[colorlinks]{hyperref}
\usepackage{multirow}
\usepackage{times}
\usepackage{amsmath,amssymb}
\usepackage{algorithm}
\usepackage{epstopdf}
\usepackage{subfigure}
\usepackage[dvipsnames]{xcolor}
\usepackage{algorithm}
\usepackage{algorithmic}
\usepackage{cite}
\usepackage{enumitem}

\usepackage{bbm}

{\begin{list}               
    {$\bullet$ \hfill}{
        \setlength{\leftmargin}{\parindent}
        \setlength{\parsep}{0.04\baselineskip}
        \setlength{\itemsep}{0.5\parsep}
        \setlength{\labelwidth}{\leftmargin}
        \setlength{\labelsep}{0em}}
    }
{\end{list}}

\providecommand{\eref}[1]{\eqref{#1}}  
\providecommand{\cref}[1]{Chapter~\ref{#1}}

\providecommand{\fref}[1]{Figure~\ref{#1}}

\providecommand{\R}{\ensuremath{\mathbb{R}}}

\providecommand{\calA}{\mathcal{A}}

\providecommand{\calX}{\mathcal{X}}

\newcommand{\minimize}[1]{\mathop{\underset{#1}{\mathrm{minimize}}}}

\begin{document}

\title{Detecting and Segmenting Adversarial Graphics Patterns from Images}

\author{Xiangyu Qu, Stanely H. Chan\\
School of Electrical and Computer Engineering, Purdue University, West Lafayette, Indiana USA\\
{\tt\small \{qu27, stanchan\}@purdue.edu}
}

\maketitle

\begin{abstract}
Adversarial attacks pose a substantial threat to computer vision system security, but the social media industry constantly faces another form of ``adversarial attack'' in which the hackers attempt to upload inappropriate images and fool the automated screening systems by adding artificial graphics patterns. In this paper, we formulate the defense against such attacks as an artificial graphics pattern segmentation problem. We evaluate the efficacy of several segmentation algorithms and, based on observation of their performance, propose a new method tailored to this specific problem. Extensive experiments show that the proposed method outperforms the baselines and has a promising generalization capability, which is the most crucial aspect in segmenting artificial graphics patterns.
\end{abstract}

\section{Introduction}
\begin{figure}[t]
\setlength\tabcolsep{1pt}
\renewcommand{\arraystretch}{0.5}
\centering
\begin{tabular}{cc}
\includegraphics[width=0.5\linewidth,height=0.3\linewidth]{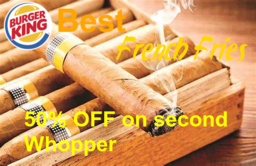}&\includegraphics[width=0.5\linewidth,height=0.3\linewidth]{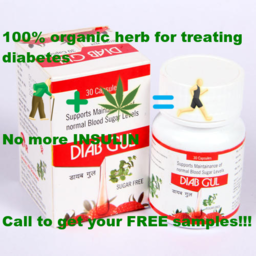}\\
\includegraphics[width=0.5\linewidth,height=0.3\linewidth]{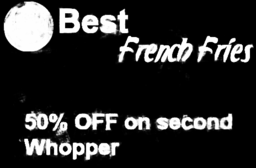}&\includegraphics[width=0.5\linewidth,height=0.3\linewidth]{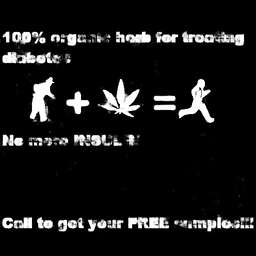}\\
(a) cigar Ad & (b) illicit medicine Ad \\
\end{tabular}
\vspace{2ex}
\caption[]{[Top row] Example unlawful/policy-violating advertisement images with adversarially added graphics patterns for bypassing screening systems\protect\footnotemark. [Bottom row] The masks generated by the proposed method. We use one network for all these types of perturbations. Detailed results can be found in the experiment section.}
\label{fig: figure 1}
\end{figure}

Social media enables people to share their lives with the world through images and videos. However, some people use these platforms for more sinister purposes: spreading illegal content, selling drugs, and making illicit advertisements \cite{illicit_online_drug_sale, porn_national_survey, image_substance_use_on_social_media} as shown in \fref{fig: figure 1}. Most of today's social media platforms use automated screening systems for filtering these inappropriate contents by using a variety of computer vision tools \cite{dnn_for_fighting_child_porn, porn_detection_review}. However, these automated screening systems are not robust as the offenders have managed to breach them by altering the policy-violating content. Their approaches are sometimes simple, e.g, drawing patterns or adding texts. Although not as powerful as adversarial attacks in terms of minimizing the visual difference between the original and the altered images, adding artificial graphics patterns is easily doable by a layperson and these simple artificial graphics patterns can be effective. As a simple toy example, we conducted an experiments on ImageNet with pre-trained ResNet-101 and Mobilenet-v3large provided by PyTorch. The accuracy of each classifier drops from 76.9\% to 54.6\% and 72.1\% to 41.8\%  respectively, although the randomly added graphics patterns occupy less than $4\%$ of the image area.

\footnotetext{Due to privacy of the real data, the examples shown here are created by the authors using Photoshop for illustration purposes only.}

At the first glance, adversarial training would seem be to an effective solution. However, considering the set of all possible graphics patterns, deploying adversarial training at a large-scale is challenging. Inspired by \cite{dropblock}, in which the authors show that a classifier can be trained and performs equally well during testing even if large connected regions are masked, we argue that a more practical solution is to disentangle the task of identifying added graphics patterns and the task of screening. This way, all screening classifiers can be trained on fixed masked images and one only needs to re-train or fine-tune the graphics pattern detector whenever a novel adversarial pattern appears. In this paper, we focus on detecting and segmenting adversarially added \textit{artificial graphics patterns}, as illustrated in \fref{fig: figure 1}. Given the context of the problem, we limit the scope of artificial graphics patterns to \textit{any} relatively simple patterns that are drawn using computer graphics software (e.g. Adobe Photoshop, Microsoft Paint), such as stickers, text, lines, shapes, etc. As we show in our experiments, by leveraging multi-scale modeling and mining hard samples (i.e. more complicated patterns with less foreground-background visual difference), models can generalize well to out-of-distribution samples even if the training set is tiny and the variety of patterns in the training set is limited.

Our contribution is summarized as following:
\begin{itemize}
\setlength\itemsep{0ex}
    \item We take a novel perspective in handling adversarial attacks by artificial graphics patterns that are prevalent in industry and establish baseline results for various SOTA segmentation systems. 
    \item We propose a data synthesis scheme to systematically simulate training images. Our synthesis method adjusts the adversarial patterns so that the visual features are similar to the local neighborhood and enables hard sample mining.
    \item We demonstrate that utilization of multi-scale information and lower level features is crucial for generalizing to out-of-distribution patterns. Based on our observations, we propose a novel cascade network with custom training policy and show that it achieves better generalization to unseen patterns and more consistent performance across pattern size.
\end{itemize}

\section{Related Work}
The problem studied in this paper is about segmenting patterns created by human using computer graphics software. To our knowledge, previous literature does not have a solution to this specific problem. Subjects such as image forensics \cite{Deep-Cascade, Huh_2018_ECCV, 8014966, SALLOUM2018201, Caldelli_2017_CVPR_Workshops, 1381775} are traditionally framed under unsupervised learning or single-class classification settings. Splice detection methods such as \cite{Splicebuster} focus on detecting splicing between two natural images and therefore have a different data domain. Watermark and logo detection \cite{Dekel_2017_CVPR, deep_logo_detection} are related tasks but SOTA solutions that generate segmentation masks are similar to those in saliency detection and semantic segmentation.

\subsection{Semantic Segmentation and Saliency Detection}
Since the goal of this paper is about segmentation, the most natural comparisons would be semantic segmentation and saliency detection, summarized as follows.

\textbf{Semantic segmentation}. Semantic segmentation aims at giving each pixel a categorical label. Traditional approaches focus on using hierarchical graph models of superpixels \cite{HarmonyPotentials, AssociativehierarchicalCRFsforobjectclassimagesegmentation, PylonModelforSemanticSegmentation}. Convolutional networks outperform these methods \cite{FCN}. However, early works usually have high model capacity, only produce low-resolution output due to large receptive field, and do not use multi-scale information well. Improvements have been proposed by, for example, Farabet et al.  \cite{LearningHierarchicalFeaturesforSceneLabeling}, Chen et al. \cite{DeepLab}, Zhao et al. \cite{PSPNet}, and Ronneberger et al. \cite{Unet}. Recent works leverage a combination of these improved techniques \cite{DLV3, AttentiontoScale:Scale-AwareSemanticImageSegmentation, Feedforwardsemanticsegmentationwithzoom-outfeatures, ContextEncodingforSemanticSegmentation, RefineNet, EfficientPiecewiseTrainingofDeepStructuredModelsforSemanticSegmentation, Chen_2019_CVPR, upernet, panopticfpn}.

While semantic segmentation can be considered as a parent task of our problem, the diversity of the data domain makes our problem different. In general, data from the same class in a traditional semantic segmentation task (e.g. scene parsing and face parsing) share a common global context (e.g. shape) and have a relatively fixed feature-to-scale correspondence. Our problem does not share these characteristics, and therefore many generic semantic segmentation methods fail. In the experiment section, we will use DeepLabv3 (DLV3) \cite{DLV3} and UNet \cite{Unet} as baselines for comparison with our proposed method.

\textbf{Saliency detection}. The goal of saliency detection is to locate the most informative region in a natural image. Arguably, one can translate the problem to ours by declaring that an artificial pattern region is important and natural regions are not. However, when framed in this context, to our knowledge, there is no off-the-shelf solution presented, although many lessons in saliency detection can be learned.

Existing saliency detection methods range from using hand-crafted features and graphical models \cite{GlobalContrastBasedSalientRegionDetection, SubmodularSalientRegionDetection, SaliencyDetectionviaaMultipleSelf-WeightedGraph-BasedManifoldRanking} to the recent deep learning approaches \cite{PyramidFeatureAttentionNetworkforSaliencyDetection, DeeplySupervisedSalientObjectDetectionwithShortConnections, Non-localDeepFeaturesforSalientObjectDetection, SalientObjectDetectionwithRecurrentFullyConvolutionalNetworks, Amulet:AggregatingMulti-levelConvolutionalFeaturesforSalientObjectDetection, BASNet:Boundary-AwareSalientObjectDetection, Visualsaliencybasedonmultiscaledeepfeatures, Saliencydetectionbymulti-contextdeeplearning, DHSNet:DeepHierarchicalSaliencyNetworkforSalientObjectDetection, DeepContrastLearningforSalientObjectDetection, DeepSaliencywithEncodedLowLevelDistanceMapandHighLevelFeatures, AStagewiseRefinementModelforDetectingSalientObjectsinImages}. For example, modeling local and global context by a two-stream network \cite{Saliencydetectionbymulti-contextdeeplearning}, exploiting both pixel-level prediction and region-level information \cite{DeepContrastLearningforSalientObjectDetection}, making a coarse prediction and then gradually refine the prediction with recurrent network \cite{DHSNet:DeepHierarchicalSaliencyNetworkforSalientObjectDetection}. Recent works focus on combining hand-crafted features with deep network \cite{DeepSaliencywithEncodedLowLevelDistanceMapandHighLevelFeatures}, better multi-scale and multi-level joint modeling \cite{DeepContrastLearningforSalientObjectDetection, DeeplySupervisedSalientObjectDetectionwithShortConnections, Amulet:AggregatingMulti-levelConvolutionalFeaturesforSalientObjectDetection, Non-localDeepFeaturesforSalientObjectDetection, PyramidFeatureAttentionNetworkforSaliencyDetection}, or more accurate segmentation along boundaries \cite{BASNet:Boundary-AwareSalientObjectDetection, AStagewiseRefinementModelforDetectingSalientObjectsinImages, SalientObjectDetectionwithRecurrentFullyConvolutionalNetworks}. In the experiment section, we will compare with the BASNet \cite{BASNet:Boundary-AwareSalientObjectDetection} and PFANet \cite{PyramidFeatureAttentionNetworkforSaliencyDetection}, which are state-of-the-art saliency detection methods.

\begin{figure*}[t]
\centering
\begin{tabular}{cc}
\includegraphics[height = 5cm]{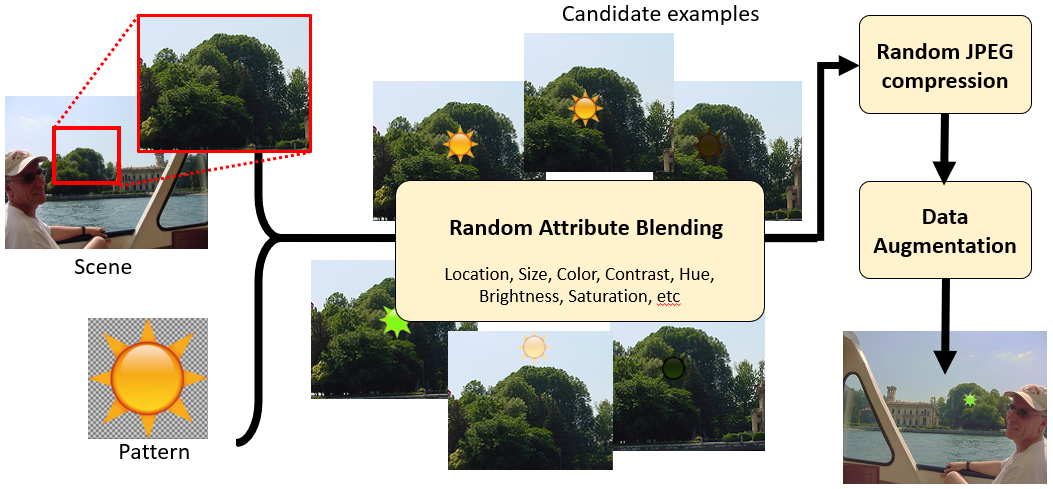}&
\hspace{-2ex}
\includegraphics[height = 5cm, width=0.35\linewidth]{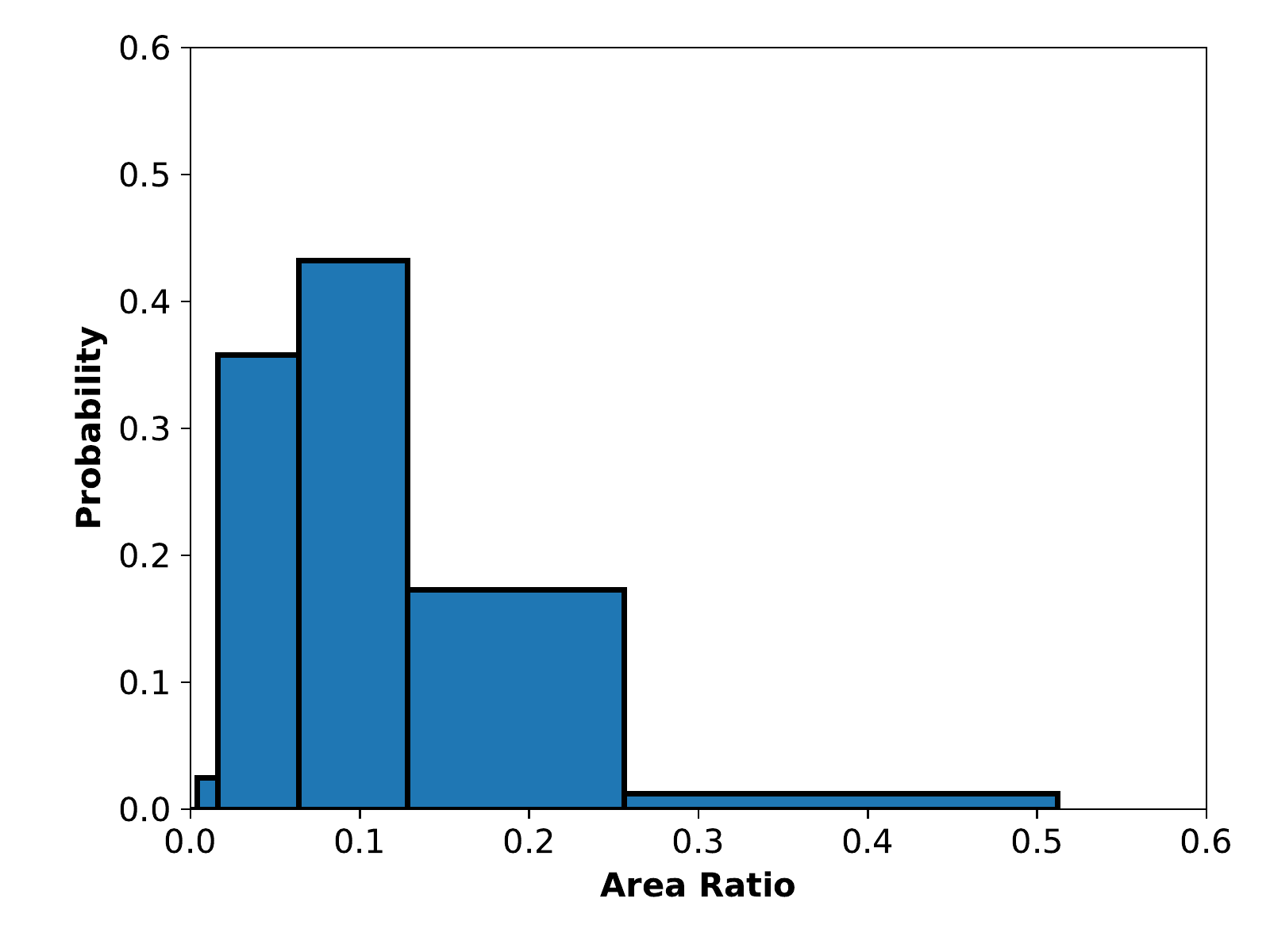}\\
(a) Proposed data synthesis pipeline & (b) Empirical distribution of sizes
\end{tabular}
\vspace{2ex}
\caption{(a) The proposed data synthesis pipeline: We first draw random images and artificial patterns from their respective datasets. For each random location we choose, we adjust the pattern until the attributes match with those in the local neighborhood. JPEG compression is added to improve robustness against compression artifacts. (b) Empirical distributions of the sizes of the artificial patterns in a dataset containing 100 images downloaded from popular social media websites. Following this analysis, we configure the size of our synthesized pattern to approximately 0.1\%-25\% of the image.}
\label{fig: data generation}
\end{figure*}

\subsection{Cascade Models}
The backbone of our proposed solution is a cascade network. This idea is inspired by the Viola-Jones detection framework \cite{viola-jones}. In Viola-Jones, cascading is used for computational efficiency. In our case, we use cascading to induce implicit attention and save the model capacity to resolve regions that cannot be resolved at shallower layers. Previous CNN based works have explored similar ideas in other areas of computer vision such as face landmark prediction \cite{DeepConvolutionalNetworkCascadeforFacialPointDetection}, pose estimation \cite{DeepPose}, face detection \cite{Aconvolutionalneuralnetworkcascadeforfacedetection}, classification \cite{DeepDecisionNetworkforMulti-classImageClassification}, and perceptual edge detection \cite{He_2019_CVPR}. For image segmentation, there are attempts for coarse-to-fine class segmentation \cite{10.1007/978-3-319-46723-8_48}, multi-stream fusion \cite{ADE20K}, and class-agnostic segmentation refinement \cite{Cheng_2020_CVPR}. Finally, our proposed method shares some common ideas with a work earlier by Li et al. \cite{NotAllPixelsAreEqual}. The main differences are twofold: (1) our proposed architecture operates on multi-scale input with multiple entry points and is different from theirs. (2) We train our multi-scale cascade network bottom-up at multiple stages so that coarser-scale network guide the training of the finer-scale network, rather than training all sub-networks jointly in two stages.

\subsection{Multi-scale Methods}
The proposed method uses a multi-scale framework to model and extract features across scale variations. While this idea has been used in many prior work, e.g., \cite{Visualsaliencybasedonmultiscaledeepfeatures, Saliencydetectionbymulti-contextdeeplearning, Feedforwardsemanticsegmentationwithzoom-outfeatures, EfficientPiecewiseTrainingofDeepStructuredModelsforSemanticSegmentation,panopticfpn,AttentiontoScale:Scale-AwareSemanticImageSegmentation, LearningHierarchicalFeaturesforSceneLabeling}, the specific usage of the scale information is quite different when we compare the two perspectives in Table~\ref{tab:multiscale-comparison}: 1) At which stage is multi-scale modeling constructed: explicitly extract features from image at different scales (explicit) or treat feature maps from different levels of a network as multi-scale (implicit). 2) How information from different scales is used to generate final predictions: aggregate features or aggregate predictions.

\begin{table}[ht]
\centering
\begin{tabular}{ccc}
& feature aggreg. & prediction aggreg. \\
\hline
\hline
explicit & \cite{Visualsaliencybasedonmultiscaledeepfeatures, Saliencydetectionbymulti-contextdeeplearning, Feedforwardsemanticsegmentationwithzoom-outfeatures, EfficientPiecewiseTrainingofDeepStructuredModelsforSemanticSegmentation} & \cite{AttentiontoScale:Scale-AwareSemanticImageSegmentation, LearningHierarchicalFeaturesforSceneLabeling}, ours \\ \hline
implicit & \cite{DeepContrastLearningforSalientObjectDetection, PSPNet, PyramidFeatureAttentionNetworkforSaliencyDetection, Non-localDeepFeaturesforSalientObjectDetection} & \cite{Amulet:AggregatingMulti-levelConvolutionalFeaturesforSalientObjectDetection, DeeplySupervisedSalientObjectDetectionwithShortConnections, panopticfpn} \\ 
& \cite{BASNet:Boundary-AwareSalientObjectDetection, DHSNet:DeepHierarchicalSaliencyNetworkforSalientObjectDetection, upernet} & \\
\hline
\end{tabular}
\caption{Taxonomy of multi-scale modeling by semantic segmentation and saliency detection methods}
\label{tab:multiscale-comparison}
\end{table}

Most of prediction aggregation methods in the segmentation literature take the weight average of the predictions from multiple scales as the final prediction, where the weighting is implemented as a learn-able layer. In contrast, our model takes the conjunction of predictions from different scales for each pixel. This will be illustrated in the experiment section when we compare with the panoptic feature pyramid network (SFPN) by Kirillov et al. \cite{panopticfpn}.

\section{Method}
We discuss the three ideas of this paper: a data synthesis pipeline, a multi-scale network, and a training scheme. 

\subsection{Data Synthesis}
\label{sec:data_synthesis}

Since the actual process of adding artificial patterns is relatively simple to emulate, we synthesize the data during training on-the-fly.

\begin{figure*}[t]
\centering
\includegraphics[width=0.85\linewidth]{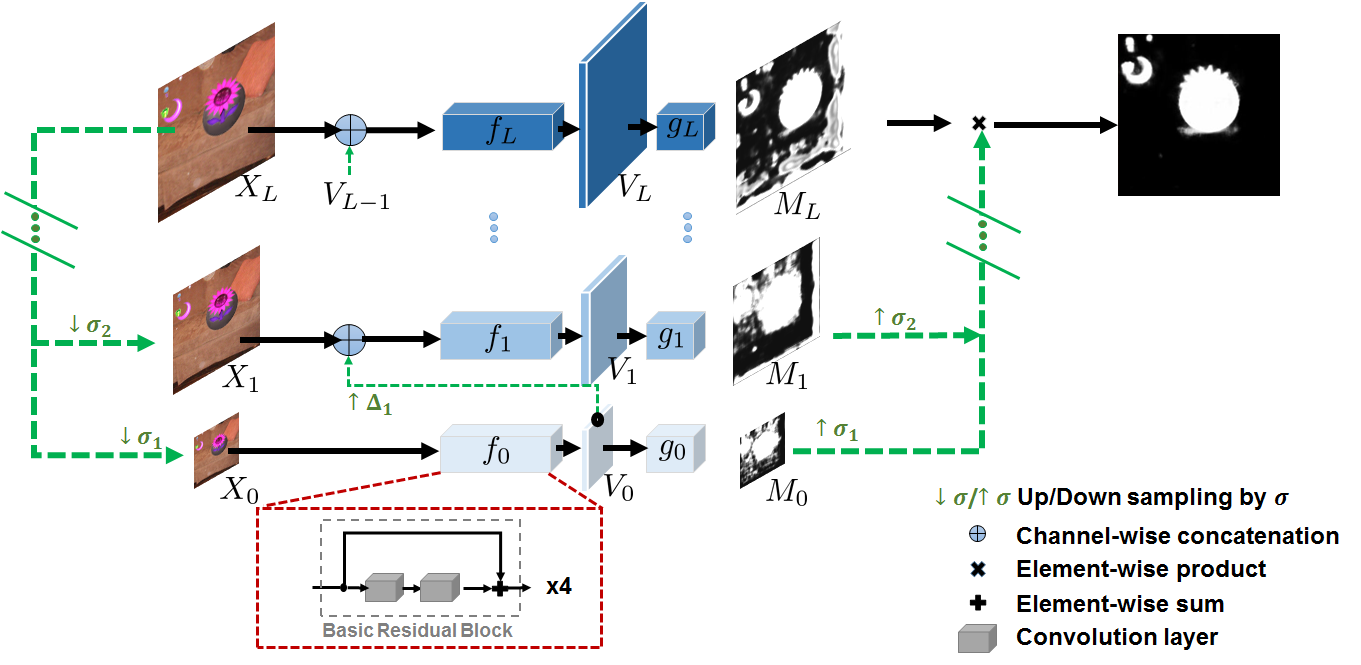}
\caption{The proposed multi-scale cascade network. In our model, we use a cascade of $L$ levels of sub-networks. The goal of each sub-network is to segment the graphics patterns in their respective scales. Information at the current level is propagated to the next level, so that we can accumulate the prediction results. Under such a design, the final mask is generated by taking the intersection of the patterns of all the scales. Note how prediction from each sub-network complement each other and allows false positive predictions in other sub-networks}
\label{fig: multis-scale}
\end{figure*}

\textbf{Procedure for synthesizing training data}. Our data generation pipeline is as illustrated in \fref{fig: data generation}(a). We identify four categories of artificial content: stickers (e.g. emoji), text, lines/stripes, and digital logos. We construct a set $\calA$ of these canonical patterns, which consists of 381 samples of high-resolution patterns. The size of these added patterns consumes approximately from 0.1\% to 25\% of the size of the images, with a higher concentration in the bottom quantile. \fref{fig: data generation}(b) shows an empirical distribution of 100 randomly downloaded images from the internet.

Given an input image $X \in \R^{H \times W \times 3}$ drawn from a dataset $\calX$ of natural images, we randomly draw $K$ artificial patterns $A_1, ..., A_K$ from the canonical pattern set $\calA$. We resize these $K$ patterns so that the sum of the area of all $K$ patterns reach a target proportion, e.g., 10\% of the image size. For each pattern $A_k$ ($k = 1,\ldots,K$), we randomly put it at a location in the image $X$. Afterward, we perform JPEG compression with a random quality factor between 70 and 100 to the resulting image. This step is to ensure that any compression artifacts are learned.

In order to make sure that the manipulated image does not have obvious features for saliency detection (so that the training data is not too easy), each pattern $A_k$ is pre-processed before added to the image. This observation is confirmed by Jiang et al. \cite{SubmodularSalientRegionDetection, SaliencyDetectionviaaMultipleSelf-WeightedGraph-BasedManifoldRanking} who showed that brightness, local and global contrast, hue, color saturation are important cues for saliency detection. Our pre-processing involves comparing these attributes (brightness, local and global contrast, hue, color saturation) of artificial patterns with the neighbors at the target location of $X$. We randomly adjust the attributes of the added patterns so they match or mis-match with those of the neighbors.

\subsection{Multi-scale Cascade Network}

\textbf{Proposed architecture}. We propose a multi-scale cascade network to encourage generalization to unseen patterns and explicitly handle the huge size variation across the patterns. The overall network is shown in \fref{fig: multis-scale}. The network consists of multiple sub-networks that operate on inputs at different scales. Each sub-network consists of two elements:
\begin{itemize}
    \item \textbf{Backbone} for extracting features: Each backbone sub-network is composed of 1 convolution layer with $C$ 3-by-3 kernels that transforms the input feature to a $C$-channel feature, and 4 basic blocks from the Resnet. 
    \item \textbf{Regression head} for generating segmentation mask: The regression head of each sub-network is composed of 1 convolution layer with 3-by-3 kernels that reduce the number of channels by half, 1 convolution layer with a 1-by-1 kernel that collapses channel number to 1, and a sigmoid function that normalizes the score map to the range $[0, 1]$.
\end{itemize}

We denote $\ell \in \{0, 1, ..., L\}$ as the scale indices from coarse to fine. Let $f_\ell$, $g_\ell$ be the backbone and regression head at scale $\ell$, respectively. Given an input image  $X \in \R^{H \times W \times 3}$, we low-pass filter it and down-sample it to obtain coarser scale images $X_\ell \in \R^{\frac{H}{\sigma_\ell} \times \frac{W}{\sigma_\ell} \times 3}$, where $\sigma_\ell$'s are the scale factors (e.g., $\sigma_\ell = \sqrt{2}$). Features extracted at the $\ell$-th layer are denoted as $V_{\ell}$'s. 

At the coarsest scale $\ell = 0$, the sub-network at that scale extracts features and makes a prediction
\begin{equation}
    \underset{\textcolor{blue}{\text{mask}}}{\underbrace{ \;\; M_\ell \;\;}} = 
    \underset{\textcolor{blue}{\text{regression}}}{\underbrace{\;\; g_\ell \;\; }}(
    \underset{\textcolor{blue}{\text{feature extract}}}{\underbrace{\;\; f_\ell(X_\ell) \;\; }}), \qquad \text{for} \; \ell = 0 \; \text{only},
\end{equation}
where $M_\ell \in \R^{\frac{H}{\sigma_\ell} \times \frac{W}{\sigma_\ell} \times 3}$ is the predicted mask by the $\ell$-th sub-network. 

In our proposed network, high-level features extracted from the coarser levels are re-used at the finer scales as a global context, and the predicted mask from the coarser scale is used to filter out regions that are difficult to make a correct prediction based on local context and features at the fine scale. Therefore, at scales $\ell = 1, \ldots, L$, the sub-network leverages extracted high-level features from the previous scale as well as the image at that scale

\begin{equation}
    M_\ell = 
    g_\ell
    \bigg(f_\ell
    \bigg(
    \overset{
    \begin{subarray}{c}
    \textcolor{blue}{\text{concatenation of}} \\
    \textcolor{blue}{\text{image and feature}}
    \end{subarray}
    }{\overbrace{
     X_\ell \;\;\; \oplus 
    }}
    \; 
    \overset{
    \begin{subarray}{c}
    \textcolor{blue}{\text{upsample of }} \\
    \textcolor{blue}{\text{$(\ell-1) ^{\text{th}}$ feature}}
    \end{subarray}
    }{\overbrace{
    \big[ \; V_{\ell-1} \; \big]_{\uparrow \Delta_\ell}}}
    \bigg)
    \bigg),
\end{equation}
where $\oplus$ denotes concatenation of along channel dimension, $[\;\; \cdot \;\; ]_{\uparrow\Delta}$ denotes the bi-linear upsampling with a factor of $\Delta$. In our problem, $\Delta_\ell = \sigma_\ell / \sigma_{\ell-1}$. 

The stage-wise output mask $\widetilde{M}_{\ell} \in \R^{H \times W \times 3}$ is the intersection of the previous stage masks. It is defined as the pixel-wise multiplication from all scales below $\ell$ 
\begin{equation}
  \widetilde{M}_{\ell}
  =
  \underset{
  \begin{subarray}{c}
  \textcolor{blue}{\text{intersection}}\\
  \textcolor{blue}{\text{of all stages}}
  \end{subarray}
  }{\underbrace{
  \prod_{i=0}^{\ell}}}
  \qquad
  \underset{
  \begin{subarray}{c}
  \textcolor{blue}{\text{upsample of}}\\
  \textcolor{blue}{\text{previous masks}}
  \end{subarray}
  }{\underbrace{
  \big[ \;\; M_i \;\; \big]_{\uparrow{\sigma_i}}
  }
  }.
\end{equation}
Our cascade scheme allows sub-networks to complement each other and therefore, each sub-network is encouraged to be "forgiving"; a sub-network can make false positive predictions at regions that are rejected by other networks.

\subsection{Customized Stage-wise Training}
\begin{figure}[h]
\centering
\includegraphics[width=\linewidth]{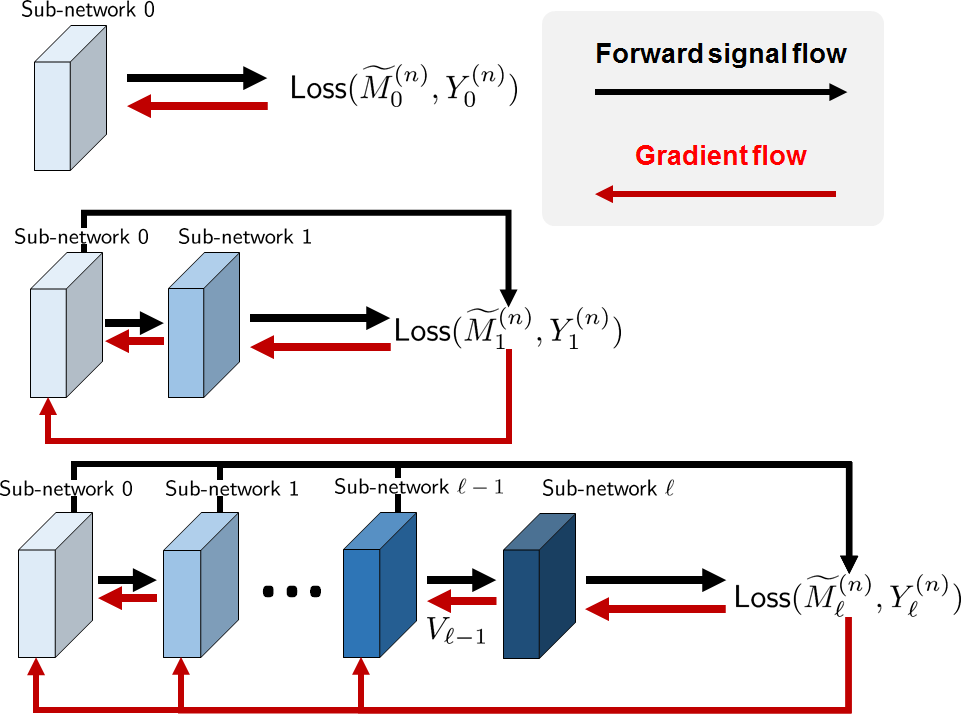} 
\caption{Multi-stage training for the proposed network. We train the sub-networks sequentially by initializing with the previous stage outputs.}
\label{fig: traininig procedure}
\end{figure}

Training of the proposed multi-scale cascade network requires a stage-wise training scheme. The training strategy is illustrated in \fref{fig: traininig procedure}. We discuss the rationale and the procedure as follows. 

\textbf{Why stage-wise training?} A stage-wise training strategy for the proposed architecture has two advantages: 
\begin{enumerate}[label=(\roman*)]
\setlength\itemsep{0ex}
    \item It allows us to explicitly specify the minimum precision ratio and the desired recall ratio for each cascading level. See our optimization below.
    \item The learned coarser scale sub-networks can guide the finer scale networks to focus on regions where positive samples might exist, and neglect regions that are already considered as hard negative (gradient to finer scale networks is masked). This encourages fine-scale classifiers to look at different features.
\end{enumerate}
If the entire network is trained jointly, the benefit of coarse-scale sub-network guidance is invalid, since the mask from a lower scale is not meaningful at that stage. We show the ablation study of multi-stage training and simultaneous joint training in supplementary material.

\textbf{Training procedure}.  We denote $(\cdot)^{(n)}$ as the $n$-th training sample. At every scale $\ell$, we consider an estimate mask $\widetilde{M}_\ell$ and a ground truth binary mask $Y_\ell$. Using the cross-entropy as the training loss, we formulate the following optimization \footnote{For notation simplicity we omit the averaging over all pixels of an image in the constraints.}:
\begin{align}
\minimize{f_\ell, \; g_\ell} &  \;\; 
-\frac{1}{N}\sum_{n=1}^N 
\bigg\{
Y_\ell^{(n)}\log \widetilde{M}_\ell^{(n)} \notag \\
&\qquad + 
(1-Y_\ell^{(n)})\log (1-\widetilde{M}_\ell^{(n)})
\bigg\} \label{eq: optimization} \\
\text{subject to} & \;\;
\underset{\text{\textcolor{blue}{average precision using threshold $\tau$}}}{\underbrace{
\frac{1}{N}\sum_{n=1}^N \text{Precision}\left(\widetilde{M}_\ell^{(n)}, Y_\ell^{(n)}, \tau \right)}} \ge P_{\text{min}},
\notag \\
& \;\;
\underset{\text{\textcolor{blue}{average recall using threshold $\tau$}}}{\underbrace{\frac{1}{N}\sum_{n=1}^N \text{Recall} \left(\widetilde{M}_\ell^{(n)}, Y_\ell^{(n)}, \tau \right)}} \ge R_{\text{min}}. \notag
\end{align}

In this optimization problem, the functions $\text{Precision}(\cdot)$ and $\text{Recall}(\cdot)$ computes the precision-recall values of the predicted mask $\widetilde{M}_{\ell}$ relative to the true mask $Y_\ell$, at a given threshold $\tau$. 

The parameters $P_{\text{min}}$ and $R_{\text{min}}$ are the minimum precision and recall levels we desire. A pixel is considered to be positive (artificial) in the final prediction by the entire network only if all sub-networks predict it to be positive. Therefore, the minimum recall $R_{\text{min}}$ at each stage must be high (so that we allow more false positives to go through). This is not a problem, because even if the false positive rate at each sub-network is high (e.g. 40\% false positive rate or 60\% precision), after cascading the final false positive rate will be low (e.g. $40\%$ false positive rate at each level will be $6.4\%$ after 3 levels of cascading). Therefore, we could have a relatively loose precision $P_{\text{min}}$ at each stage of training. In our implementation, we set $R_{\text{min}}$ and $P_{\text{min}}$ by checking final loss on validation set.

\subsection{Data Balancing}
Unlike a generic semantic segmentation problem where the sizes of the objects do not have a substantial impact on the performance  \cite{FCN}, the same issue is significantly more important in our problem. Specifically, since the added artificial contents usually only occupy a small portion of the foreground, there is a significant data imbalance between the positive and the negative samples. 

Addressing the data balancing issue is typically done by (1) drawing samples according to the desired percentage, or (2) adjusting the training loss. Our approach is based on the latter. Considering the cross-entropy loss in \eref{eq: optimization}, we introduce a per-image weighing constant $\alpha_\ell^{(n)}$ so that the loss becomes
\begin{align*}
\minimize{f_\ell, \; g_\ell} &  \;\; 
-\frac{1}{N}\sum_{n=1}^N 
\bigg\{
\textcolor{blue}{\frac{1}{\alpha_\ell^{(n)}}} Y_\ell^{(n)}\log \widetilde{M}_\ell^{(n)} \notag \\
&\qquad + 
\textcolor{blue}{\frac{1}{(1-\alpha_\ell^{(n)})}} (1-Y_\ell^{(n)})\log (1-\widetilde{M}_\ell^{(n)})
\bigg\}
\end{align*}
Here, the weighing constant is defined according to the proportion of the added content relative to the image:
\begin{equation}
    \alpha_\ell^{(n)} = \frac{1}{HW} \sum_{\text{all pixels}} Y_\ell^{(n)},
\end{equation}
where $H$ and $W$ are the number of rows and columns of the label mask $Y_{\ell}^{(n)}$.

\section{Experiments}
\subsection{Implementation and Datasets}
\textbf{Training and validation}. For the training data, we synthesize the perturbed images on-the-fly to minimize overfitting. As discussed in Section~\ref{sec:data_synthesis}, the data synthesis procedure draws clean and natural images from a dataset. Then it renders artificial patterns and adds them to the images. The clean images we use are the PASCAL VOC 2012 training set. We randomly crop and resize them to $256\times256$ and send through the data synthesis pipeline. The artificial patterns we use are manually selected from the emoji dataset used by Apple's platform. We use this small collection of Apple platform emoji rather than a larger variety of artificial contents because the goal of our experiment is to test whether a model can generalize. We further break down the collected emoji images to the training set and validation set by the emoji category. Detailed training and validation set statistics are as shown in table \ref{tab:train-val-set-stats}.

We implemented in PyTorch. We used Adam \cite{Kingma2015AdamAM} for optimizing sub-network parameters at each stage with initial learning rate 1e-3 and $\beta's$ in (0.9, 0.999).  

\begin{table}[h]
\centering
\begin{tabular}{c|c|c|c}
\hline
\hline
\multicolumn{2}{c}{Clean images} & \multicolumn{2}{c}{Total = 5717} \\ \hline
\multirow{4}{*}{Training emoji} & animals & 22 & \multirow{4}{*}{156} \\ \cline{2-3}
 & food & 36 &  \\ \cline{2-3}
 & nature & 44 &  \\ \cline{2-3}
 & objects & 54 &  \\ \hline
\multirow{3}{*}{Validation emoji} & people & 79 & \multirow{3}{*}{225} \\ \cline{2-3}
 & face & 109 &  \\ \cline{2-3}
 & hand signs & 37 &  \\ \hline
\end{tabular}
\caption{Statistics of the emoji we use to train our model.}
\label{tab:train-val-set-stats}
\end{table}

\textbf{Testing set}. To quantitatively evaluate the generalization capability of each segmentation model, we create a fixed synthetic testing dataset using a large collection of commonly seen artificial patterns:  \textbf{stickers/emoji, texts, lines/curves, and logos}. (Note that this is for testing. \textbf{We only train on a small collection of 156 Apple emoji}. Our goal is to show that our model generalizes.) Two large public emoji sets (noto-emoji by Google and twemoji by Twitter, a total of more than 6000 emoji) are used. These emoji have drastically different visual and textural appearances as compared to the small Apple emoji set. They are also widely adopted by various software/websites due to their open-source nature. For the logo patterns, we use Large Logo Dataset (LLD) by Sage et al. \cite{sage2017logodataset}. Since the size of these artificially added patterns has a significant impact on the segmentation quality, for each type of artificial content, we group the synthesized test images into three subsets (\textbf{small, medium, large}). In total, our test set has 12000 images (1000 for each category and size level). Detailed description and statistics of testing set is in supplementary material.

\begin{table*}[]
\resizebox{\textwidth}{!}{%
\begin{tabular}{c|cccc|cccc|cccc|cccc}
\hline
\hline
 & \multicolumn{4}{c|}{\textbf{Stickers}} & \multicolumn{4}{c|}{\textbf{Lines}} & \multicolumn{4}{c|}{\textbf{Texts}} & \multicolumn{4}{c}{\textbf{Logos}} \\ 
 & mIoU & MAE & $F_{0.3}$ & $F_{2}$ & mIoU & MAE & $F_{0.3}$ & $F_{2}$ & mIoU & MAE & $F_{0.3}$ & $F_{2}$ & mIoU & MAE & $F_{0.3}$ & $F_{2}$ \\ \hline
\textbf{Ours} & \textbf{0.818} & \textbf{0.011} & \textbf{0.933} & \textbf{0.894} & \textbf{0.846} & 0.022 & \textbf{0.930} & \textbf{0.942} & \textbf{0.645} & 0.017 & \textbf{0.793} & \textbf{0.826} & \textbf{0.673} & \textbf{0.038} & \textbf{0.878} & \textbf{0.822} \\ 
\textbf{Unet} & 0.512 & 0.035 & 0.713 & 0.691 & 0.690 & 0.046 & 0.843 & 0.828 & 0.321 & 0.040 & 0.440 & 0.536 & 0.398 & 0.061 & 0.680 & 0.625 \\
\textbf{DLV3} & 0.368 & 0.051 & 0.519 & 0.509 & 0.352 & 0.111 & 0.574 & 0.458 & 0.245 & 0.039 & 0.357 & 0.445 & 0.368 & 0.062 & 0.508 & 0.498 \\ 
\textbf{BASNet} & 0.702 & 0.015 & 0.827 & 0.800 & 0.819 & \textbf{0.020} & \textbf{0.930} & 0.908 & 0.589 & \textbf{0.013} & 0.716 & 0.740 & 0.579 & 0.044 & 0.752 & 0.684 \\ 
\textbf{PFANet} & 0.689 & 0.024 & 0.826 & 0.839 & 0.735 & 0.031 & 0.896 & 0.883 & 0.411 & 0.023 & 0.538 & 0.630 & 0.491 & 0.059 & 0.746 & 0.663 \\
\textbf{SFPN} & 0.630 & 0.033 & 0.816 & 0.778 & 0.599 & 0.064 & 0.793 & 0.770 & 0.315 & 0.042 & 0.451 & 0.531 & 0.562 & 0.048 & 0.780 & 0.720 \\
\hline\hline
\end{tabular}%
}
\caption{Quantitative comparison of our proposed model and baseline models across different image with artificial pattern test data}
\label{tab:comparison}
\end{table*}

\begin{figure*}[t]
\setlength\tabcolsep{1pt}
\renewcommand{\arraystretch}{0.5}
\centering
\begin{tabular}{cccc}
\includegraphics[width=0.25\linewidth]{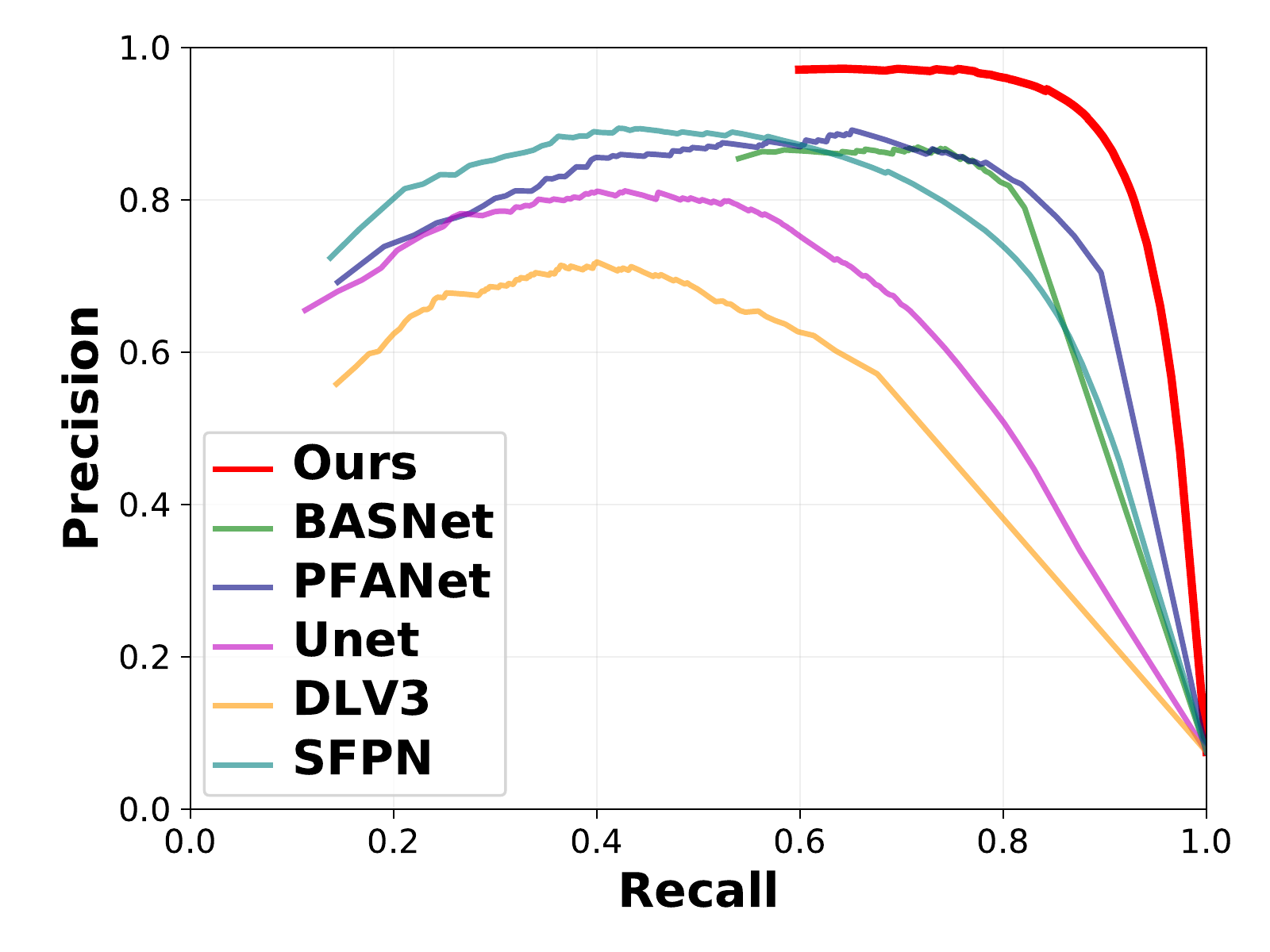} &
\includegraphics[width=0.25\linewidth]{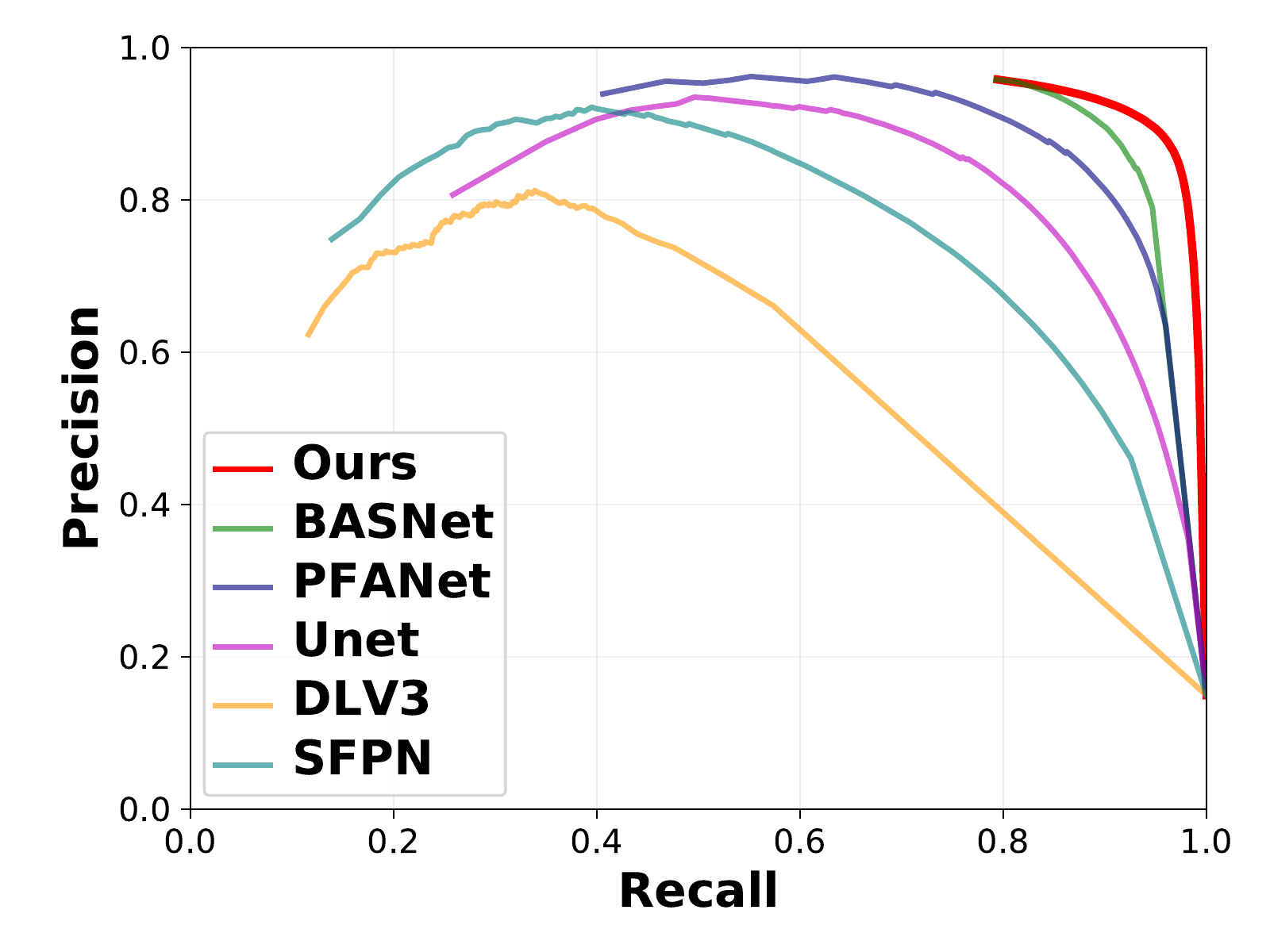} &
\includegraphics[width=0.25\linewidth]{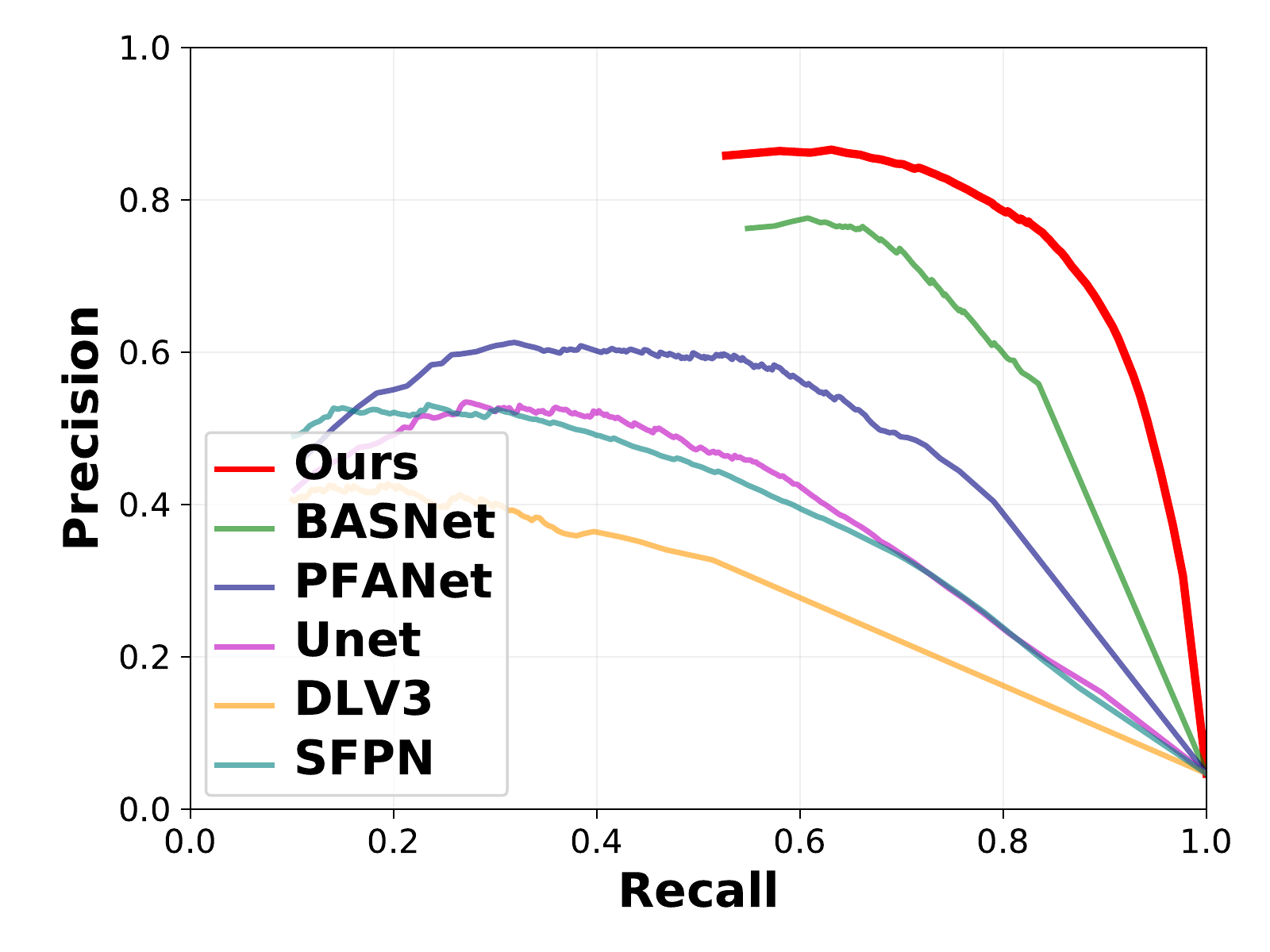} &
\includegraphics[width=0.25\linewidth]{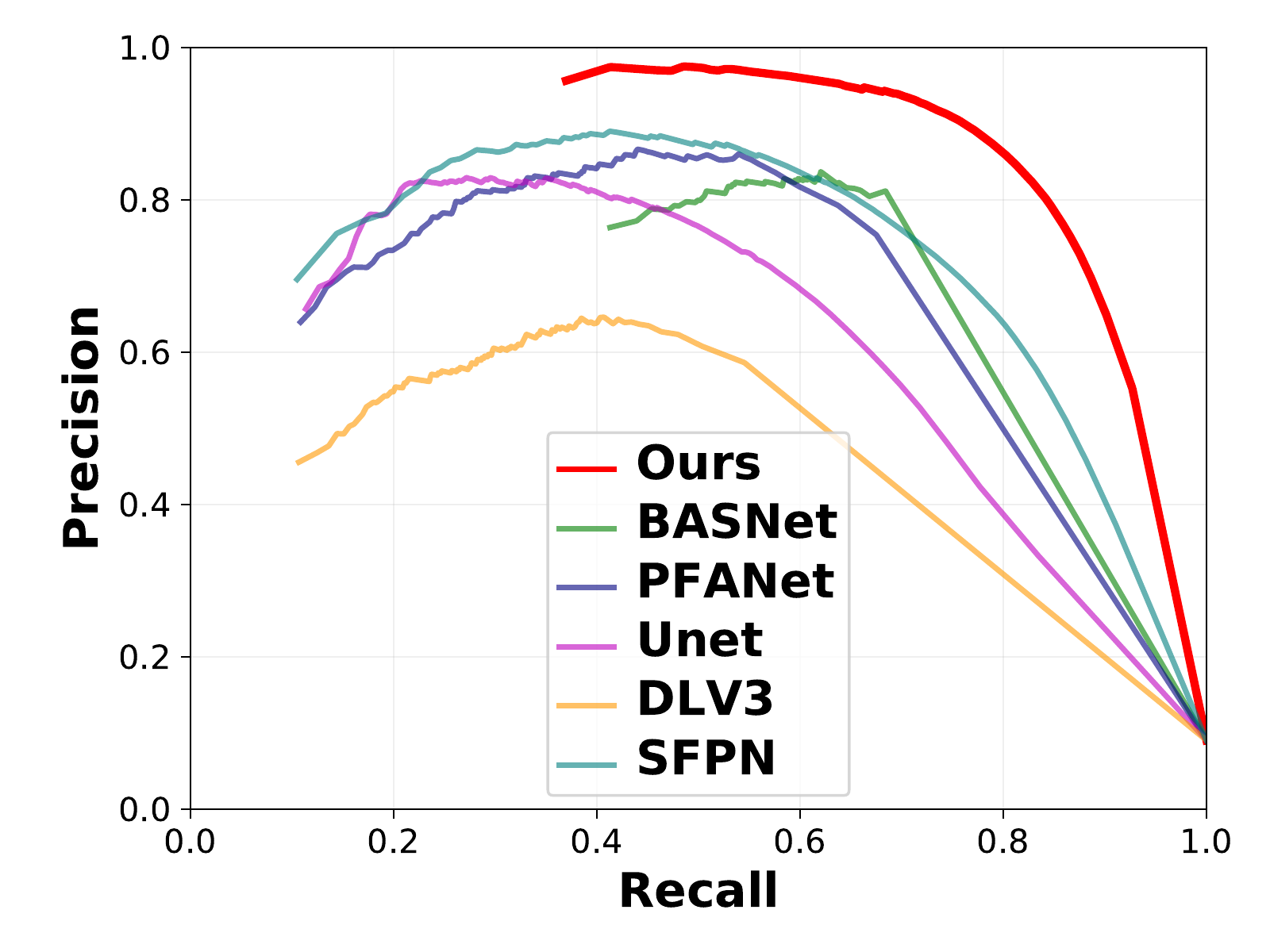} \\
\multicolumn{4}{c}{(a) Precision-recall curves} \\
\includegraphics[width=0.25\linewidth]{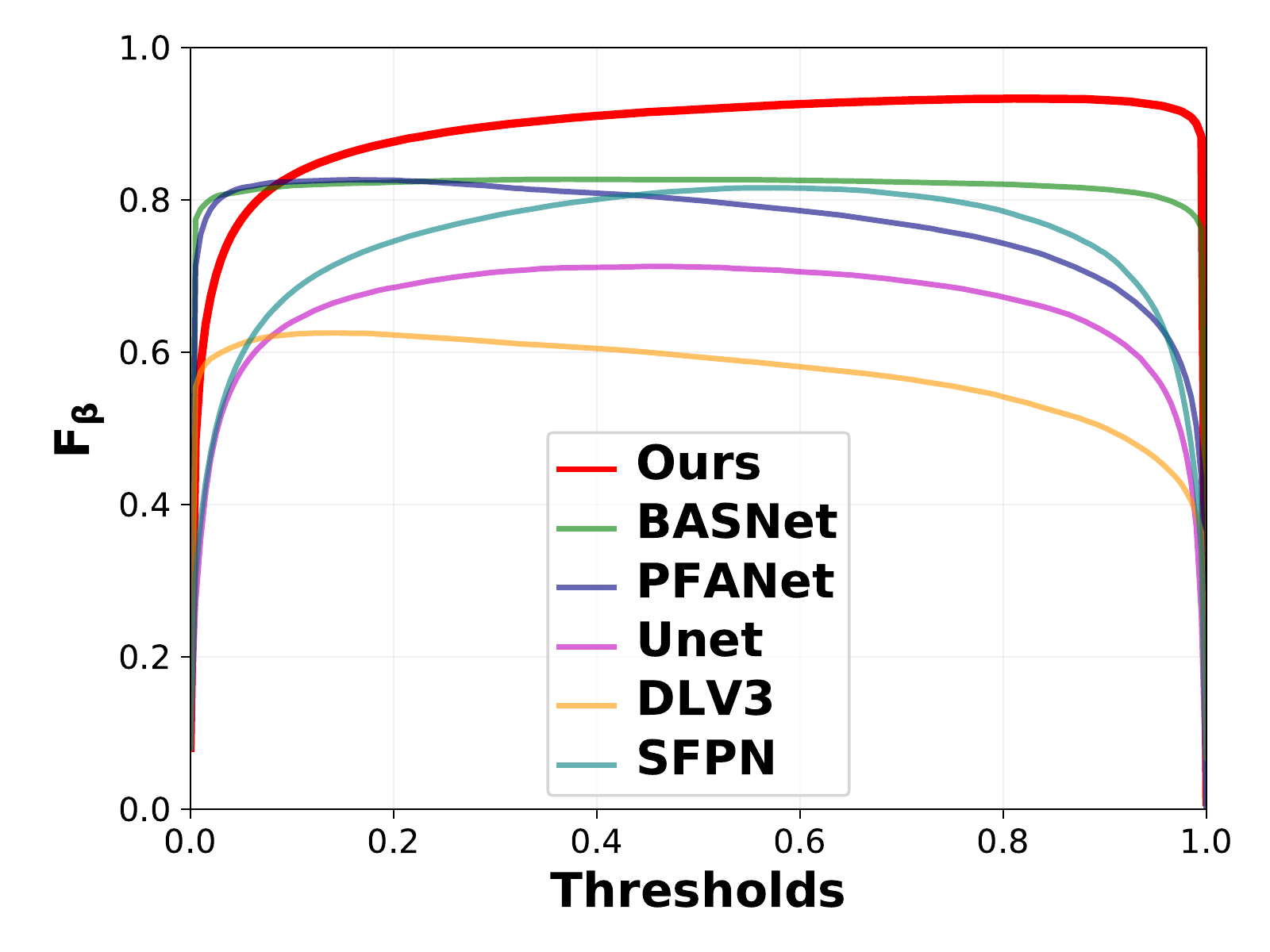} &
\includegraphics[width=0.25\linewidth]{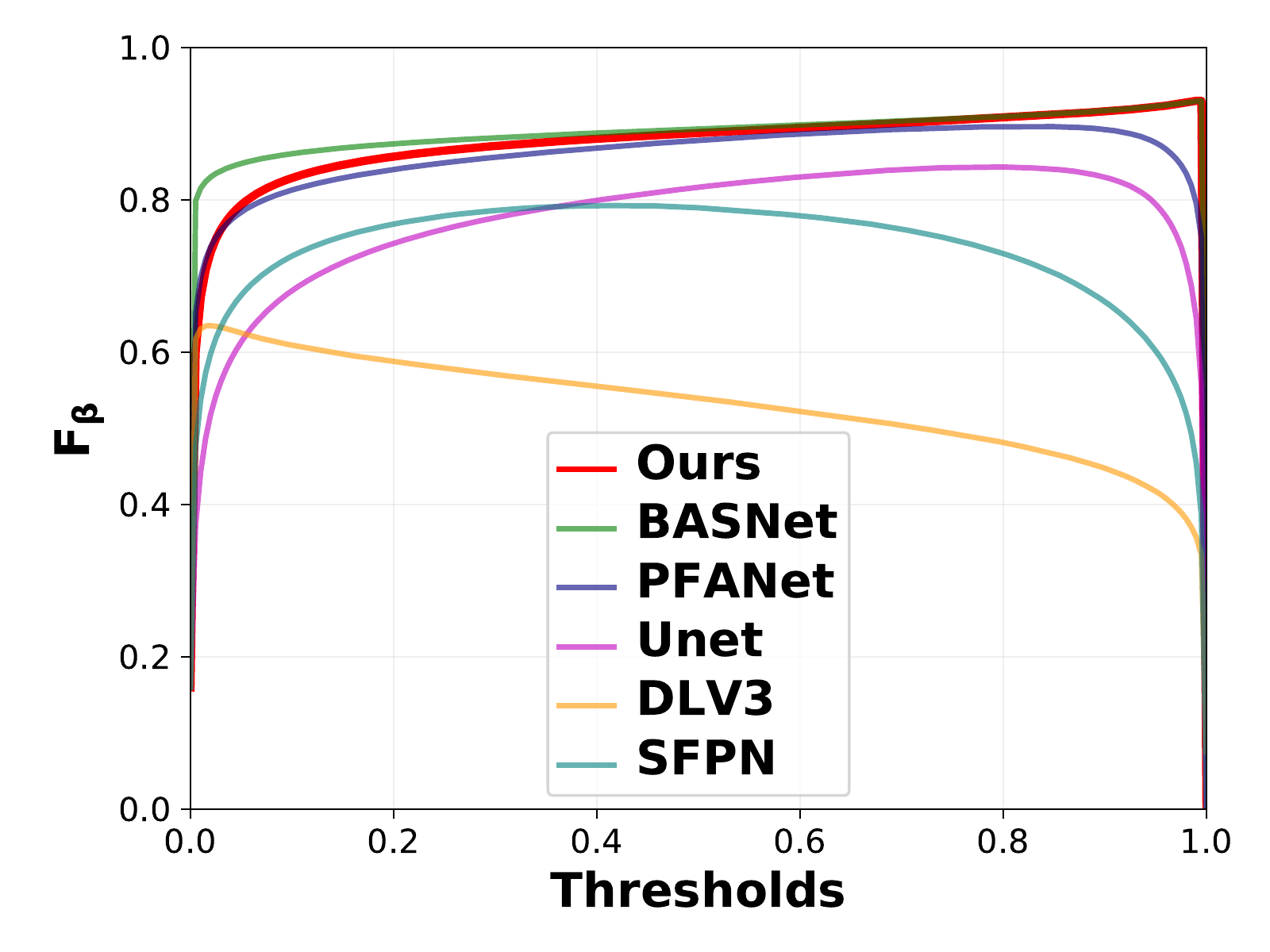} &
\includegraphics[width=0.25\linewidth]{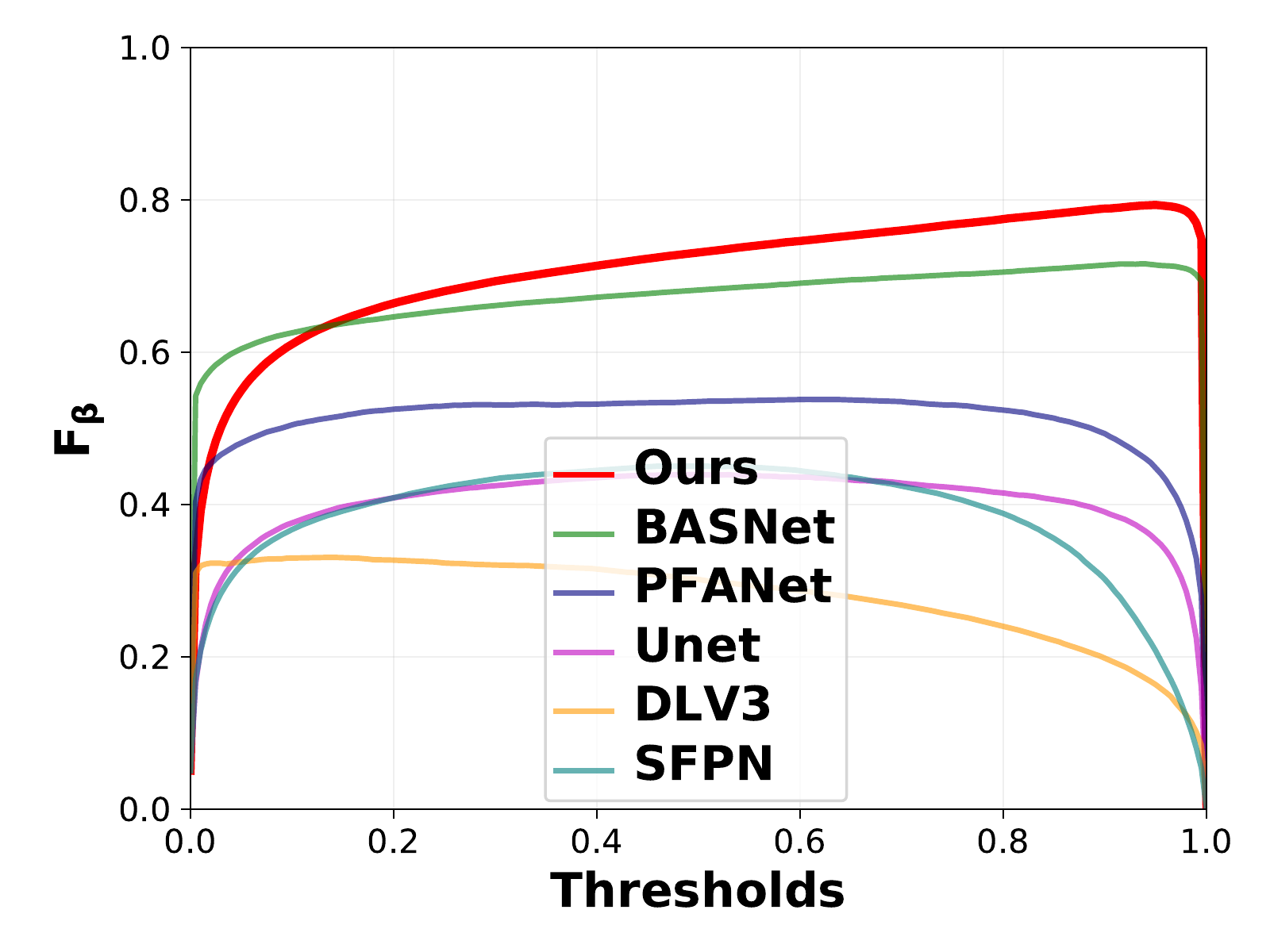} &
\includegraphics[width=0.25\linewidth]{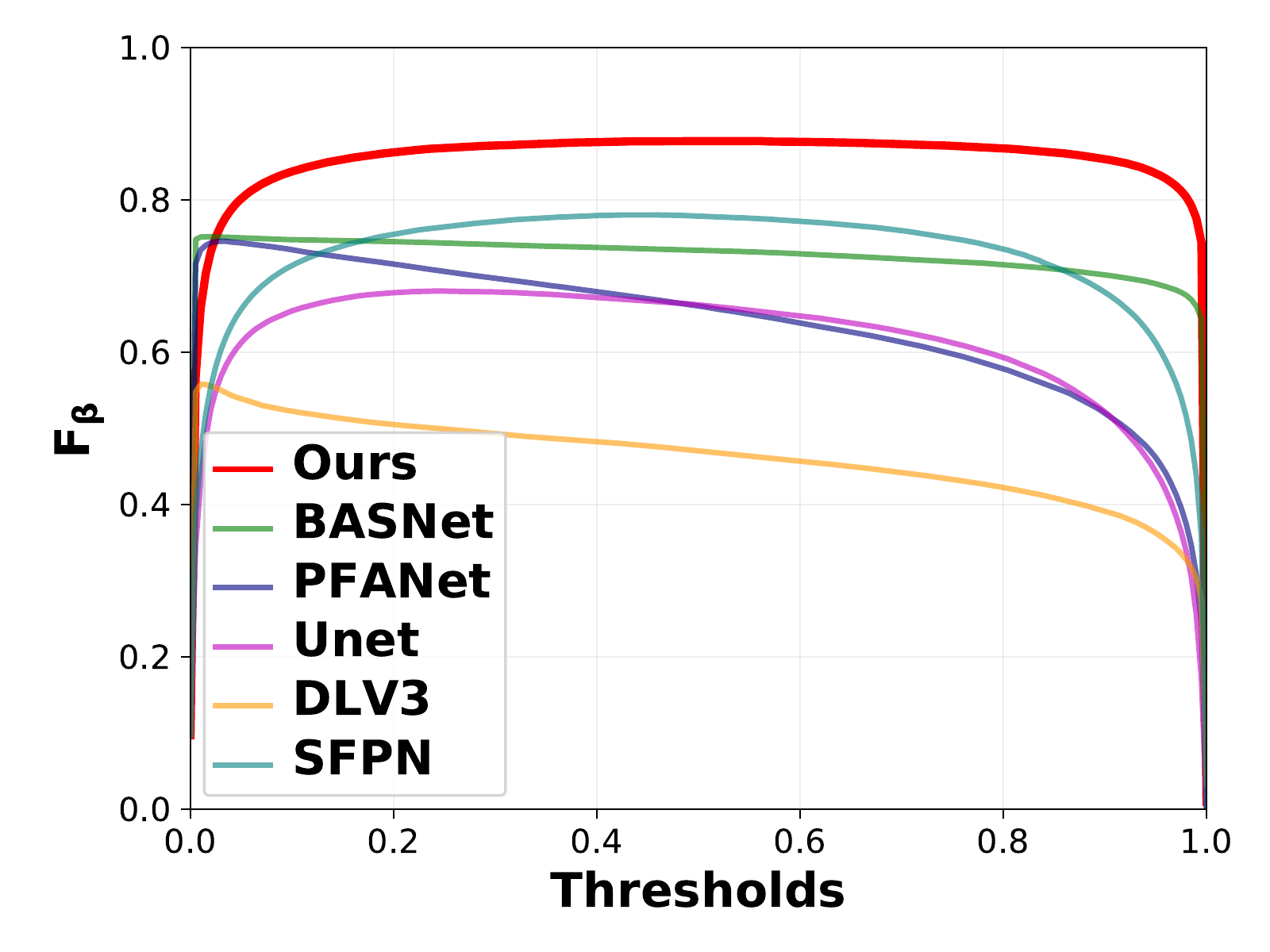} \\
\multicolumn{4}{c}{(b) $F$-score as a function of the threshold}
\end{tabular}
\caption{The precision-recall curves and the $F$-scores as a function of the thresholds. The four columns in this figure correspond to the four classes of testing patterns outlined in Table~\ref{tab:comparison}. }
\label{fig:quant_comparison}
\end{figure*}

\subsection{Evaluation Metrics}
Following the semantic segmentation and the saliency detection literature, we evaluate the performance of the baseline methods and our proposed methods on the test set with the following metrics: (i) The \textbf{mean intersection-over-union (mIoU)}. The threshold for mIoU was chosen to maximize the mIoU value over the validation set. (ii) The \textbf{precision-recall} curve. (iii)  The \textbf{mean absolute error (MAE)}. (iv) The \textbf{$F_{\beta}$-measure}, defined as 
\begin{equation}
    F_{\beta} = \frac{(1+\beta^2) \cdot \text{Precision} \cdot \text{Recall}}{\beta^2 \cdot \text{Precision} + \text{Recall}}
\end{equation}    
with $\beta$ set to 0.3 and 2 for emphasizing precision and recall, respectively. Saliency detection works usually report the $F_{\beta}$-measure value obtained by an adaptive threshold as suggested by \cite{5206596}. However, since our problem scope is different and so the adaptive threshold is less relevant, we report the maximum $F_{\beta}$-measure obtained on each category of the test set, as suggested by  \cite{BASNet:Boundary-AwareSalientObjectDetection}.

\subsection{Main Results}
\textbf{Baseline models}. We use DeepLabV3 (DLV3) \cite{DLV3}, Unet\cite{Unet}, BASNet\cite{BASNet:Boundary-AwareSalientObjectDetection}, PFANet\cite{PyramidFeatureAttentionNetworkforSaliencyDetection}, and panoptic FPN\cite{panopticfpn} as the baseline models. The baseline models are trained with the same data synthesis process as our proposed model. Each baseline model is initialized with either a pre-trained model (if available) or random weights. All models are trained until (1) training loss converged, or (2) validation loss and training loss diverges, whichever comes first. To demonstrate the raw detection capability, we do not include any post-processing steps such as conditional random field \cite{DeepLab, EfficientPiecewiseTrainingofDeepStructuredModelsforSemanticSegmentation} to refine the masks.

\begin{figure}[!]
\centering
\includegraphics[width=\linewidth]{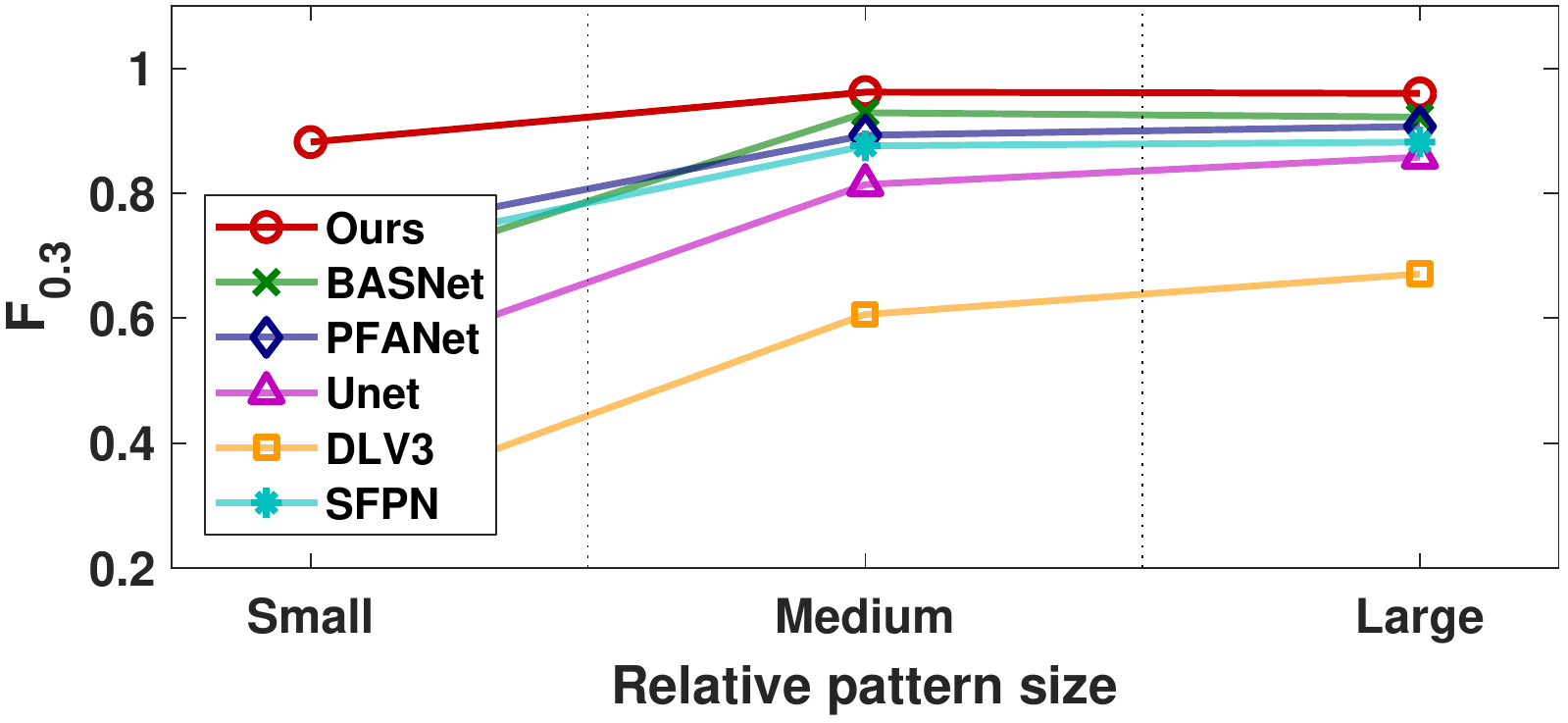}
\vspace{-2ex}
\caption{$F_{0.3}$ score as a function of the size of the added patterns. Despite all models are trained using the same training dataset, the proposed method is more resilient to the variation of the size.}
\label{fig: f-score-vs-pattern-size}
\vspace{-4ex}
\end{figure}

\textbf{Generalization to out-of-distribution patterns}. The mIoU, MAE, and the maximum $F$ scores of the competing methods are summarized in Table \ref{tab:comparison}, whereas the precision-recall curve, and the $F$-score as a function of threshold are shown in Figure \ref{fig:quant_comparison}. We divide our testing scenarios into four categories of data: stickers, lines, texts and logos. Since the models are all trained on the same Apple platform emoji, the performance analysis will indicate the generalization capability of each method.

\begin{figure*}[t]
\setlength\tabcolsep{1pt}
\renewcommand{\arraystretch}{0.5}
\centering
\begin{tabular}{cccccccc}
\includegraphics[width=0.12\linewidth]{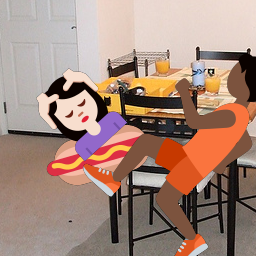} & \includegraphics[width=0.12\linewidth]{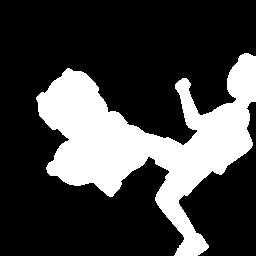} &
\includegraphics[width=0.12\linewidth]{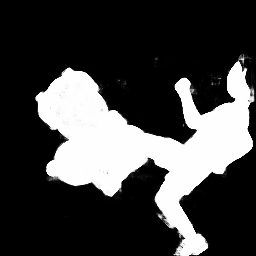} & \includegraphics[width=0.12\linewidth]{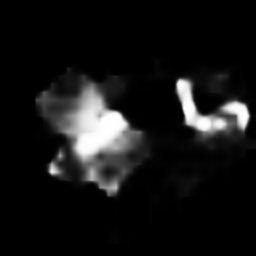} &
\includegraphics[width=0.12\linewidth]{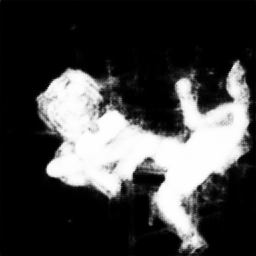} & \includegraphics[width=0.12\linewidth]{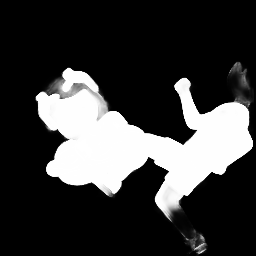} & \includegraphics[width=0.12\linewidth]{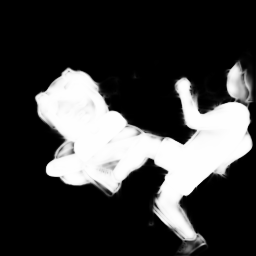} &
\includegraphics[width=0.12\linewidth]{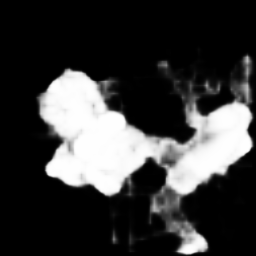} \\

\includegraphics[width=0.12\linewidth]{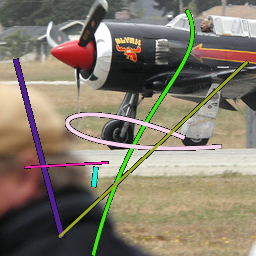} & \includegraphics[width=0.12\linewidth]{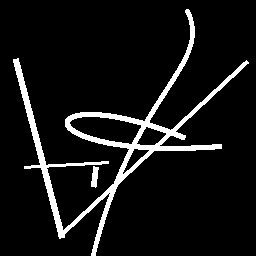} &
\includegraphics[width=0.12\linewidth]{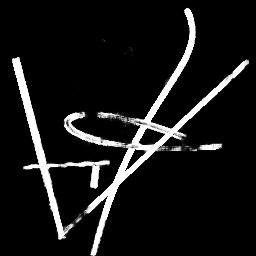} & \includegraphics[width=0.12\linewidth]{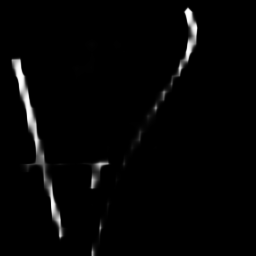} &
\includegraphics[width=0.12\linewidth]{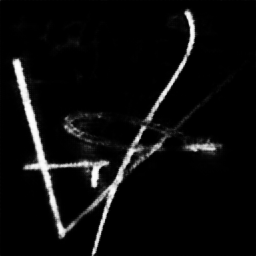} & \includegraphics[width=0.12\linewidth]{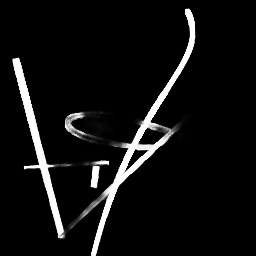} & \includegraphics[width=0.12\linewidth]{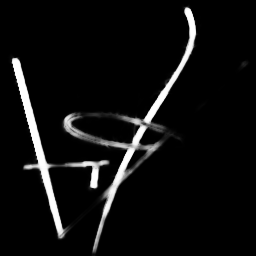} &
\includegraphics[width=0.12\linewidth]{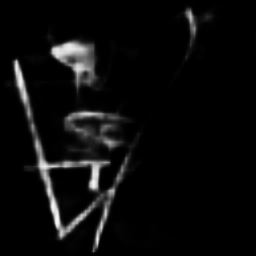} \\

\includegraphics[width=0.12\linewidth]{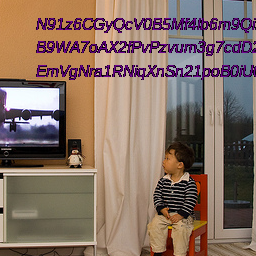} & \includegraphics[width=0.12\linewidth]{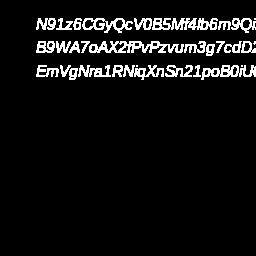} &
\includegraphics[width=0.12\linewidth]{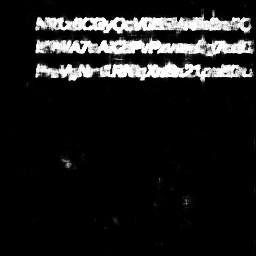} & \includegraphics[width=0.12\linewidth]{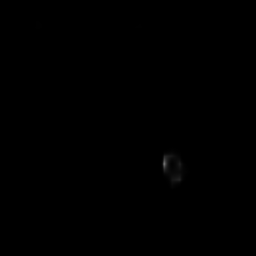} &
\includegraphics[width=0.12\linewidth]{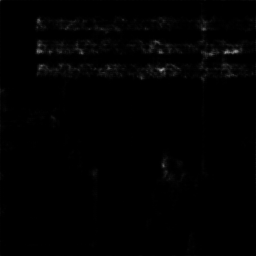} & \includegraphics[width=0.12\linewidth]{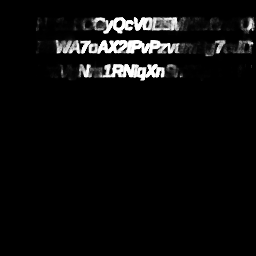} & \includegraphics[width=0.12\linewidth]{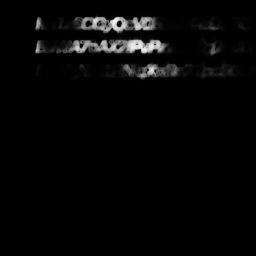} &
\includegraphics[width=0.12\linewidth]{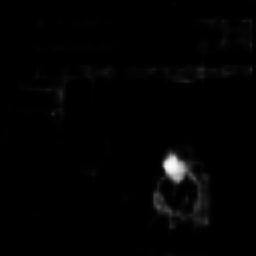} \\

\includegraphics[width=0.12\linewidth]{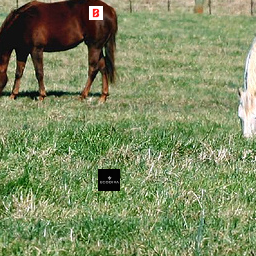} & \includegraphics[width=0.12\linewidth]{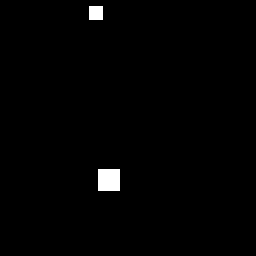} &
\includegraphics[width=0.12\linewidth]{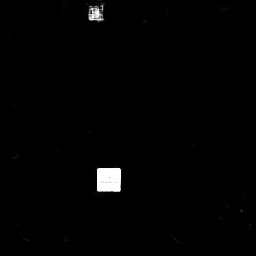} & \includegraphics[width=0.12\linewidth]{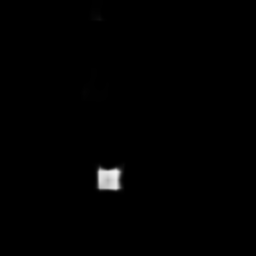} &
\includegraphics[width=0.12\linewidth]{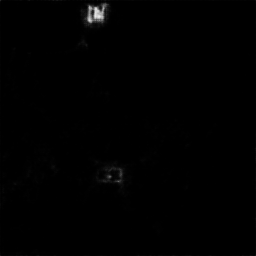} & \includegraphics[width=0.12\linewidth]{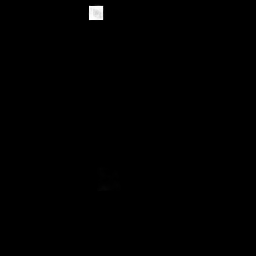} & \includegraphics[width=0.12\linewidth]{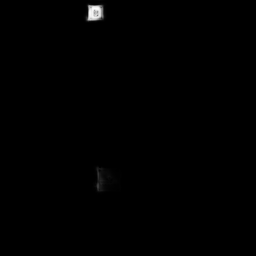} &
\includegraphics[width=0.12\linewidth]{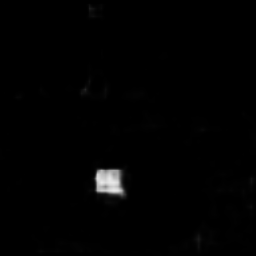} \\
& & & & & \\

(a) Image & (b) GT & (c) Ours & (d) DLV3 & (e) UNet & (f) BASNet & (g) PFANet & SFPN\\

\end{tabular}
\vspace{2ex}
\caption{Visual comparison of the proposed method and the competing methods. All models are trained using the same training dataset to ensure a fair comparison. Note that the proposed method can handle very large and very small perturbations, while some competing methods fail to do so.}
\label{fig: visualization}
\end{figure*}

According to Table \ref{tab:comparison} and Figure \ref{fig:quant_comparison}, the proposed method demonstrates better generalization than all baselines in all testing scenarios using all the evaluation metrics. Particularly, we make following observations 
\begin{itemize}
\setlength\itemsep{0ex}
    \item Regardless of the model capacity (shown in table \ref{table: complexity}), models that combine low level (shallower layer) features with high level context (Ours, Unet, BASNet, PFANet, SFPN) for making final prediction generalize better than those not (DLV3).
    \item Models that use residual unit as building block (Ours, BASNet) generate sharper boundaries
\end{itemize}

\textbf{Scale variations}. Another crucial aspect of solving graphics pattern segmentation problem is the ability to handle a wide range of scales. We show in Figure \ref{fig: f-score-vs-pattern-size} the maximum $F$-score as function of the relative sizes of the artificial patterns. 
We consider three scales of the artificial patterns relative to the image size: large, medium, and small. Detailed descriptions of these categories can be found in the supplementary materials. As shown in the figure, the proposed method has the most consistent performance among the competing methods.

\begin{table}[h]
\centering
\tabcolsep=0.15cm
\small
\begin{tabular}{ccccccc}
\hline
\hline
& Ours & DLV3 & UNet & BASNet & PFANet & SFPN \\
\# para. & 1.0M & 61M & 0.5M & 87M & 16.4M & 48M\\
\hline
\end{tabular}
\vspace{1ex}
\caption{Number of parameters used by competing methods.}
\label{table: complexity}
\end{table}

\textbf{Visual comparison}. In Figure~\ref{fig: visualization} we show several visual comparisons for very large and very small perturbations. We can see that the proposed method produces the best masks across the scales. Additional example can be found in the supplementary material.

\textbf{Wild data and ablation study}. Additional results such as the performance on wild test data collected from popular social media websites and extensive ablation studies are presented in the supplementary material due to space limit.

\section{Conclusion}
We formulate the defense against adversarially added artificial graphics patterns as an image segmentation problem and propose a promising solution. The benchmark for such a task is that the method has to handle a wide range of scales and a variety of types, shapes, colors, contrast, brightness, etc. This paper shows that the proposed cascading scheme and explicit multi-scale modeling are effective method for the problem. To train the model, we propose a multi-stage training scheme where we can control the desired precision and recall levels per-stage.

The generalization capability of the proposed method has been tested over the set of graphics patterns we reported. One interesting observation is that majority of the artificial graphics patterns are rendered through a limited set of primitive 2D graphics directives, although the fused images can be stored in any image format. Due to the simplicity and limited options of the available primitive directives, it is recommended to train on the more complex examples (stickers and text). Skipping the simpler patterns such as lines and logos will likely not hurt the performance.

Current system is optimized for opaque graphics patterns under mild corruption (e.g. noise, blur, blocking artifacts introduced by JPEG compression) and is well-suited for majority of the threat so far. However, it is expected that hackers will improve their techniques and exploit vulnerability of current systems. In particular, future research along this direction may focus on improving the performance on translucent patterns (patterns alpha blended with the original content) as well as the robustness against heavy corruption.

\noindent \textbf{Acknowledgement} The work is funded in part by the Army Research Office under the contract W911NF-20-1-0179 and the National Science Foundation under grants CCF-1763896 and CCF-1718007.

\clearpage
\pagebreak
\onecolumn

\begin{center}
\textbf{\large Supplementary material}
\end{center}
\setcounter{equation}{0}
\setcounter{figure}{0}
\setcounter{table}{0}
\setcounter{section}{0}
\makeatletter
\renewcommand{\theequation}{S\arabic{equation}}
\renewcommand{\thefigure}{S\arabic{figure}}
\renewcommand{\thetable}{S\arabic{table}}
\renewcommand{\thesection}{S\arabic{section}}
This supplementary document summarizes the following experimental results:

\begin{itemize}
\setlength\itemsep{0ex}
    \item Detailed description of the synthetic test set (section \ref{sec:test set details}).
    \item Ablation study of the proposed system (section \ref{sec: ablation study}).
    \item Evaluation on real testing downloaded from the internet (section \ref{sec: wild data test}).
    \item Additional visual comparisons (section \ref{sec: more visual results}).
\end{itemize}

\section{Test Set Details}
\label{sec:test set details}
\begin{figure}[ht]
\setlength\tabcolsep{1pt}
\renewcommand{\arraystretch}{0.5}
\centering
\begin{tabular}{ccccc}
Small &
\includegraphics[width=0.2\linewidth]{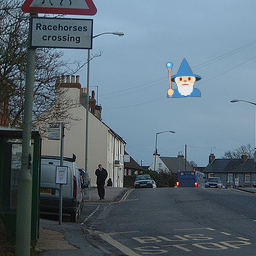}&
\includegraphics[width=0.2\linewidth]{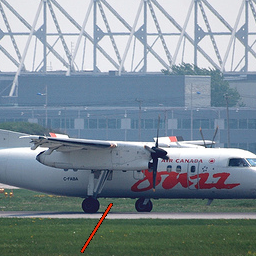}&
\includegraphics[width=0.2\linewidth]{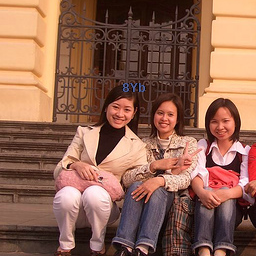}&
\includegraphics[width=0.2\linewidth]{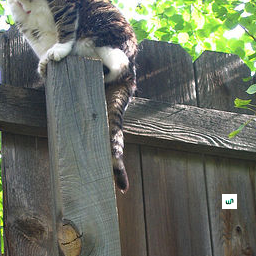}\\
& & \\
Medium &
\includegraphics[width=0.2\linewidth]{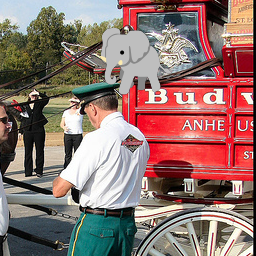}&
\includegraphics[width=0.2\linewidth]{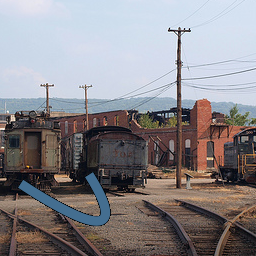}&
\includegraphics[width=0.2\linewidth]{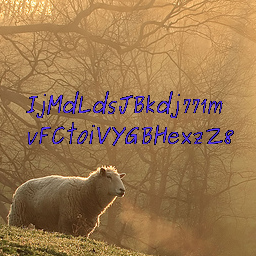}&
\includegraphics[width=0.2\linewidth]{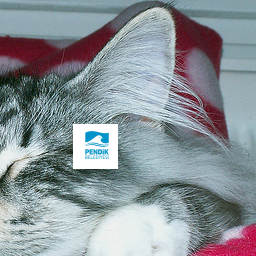}\\
& & \\
Large &
\includegraphics[width=0.2\linewidth]{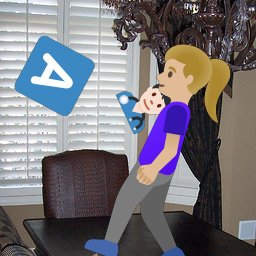}&
\includegraphics[width=0.2\linewidth]{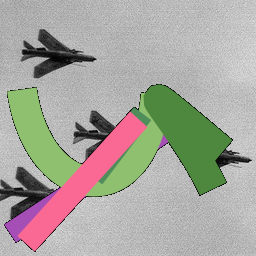}&
\includegraphics[width=0.2\linewidth]{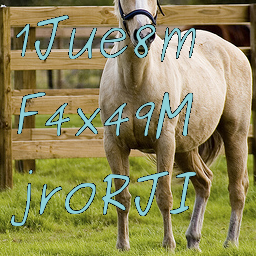}&
\includegraphics[width=0.2\linewidth]{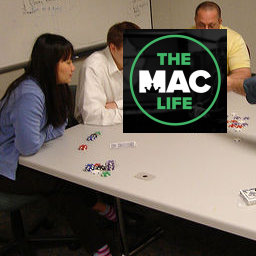}\\
& Stickers & Lines & Text & Logos \\
\end{tabular}
\vspace{2ex}
\caption{Example test images from each category and size level}
\label{fig: test set demo}
\end{figure}

Our synthetic test set is composed of 12,000 images synthesized with clean natural images and graphics patterns drastically different from training set. We used 4 types of patterns: stickers, lines, text, and logos. For each type of pattern, we synthesize test images at 3 different size levels: large, medium, and small. Sizes are defined slightly different across different patterns as it is very difficult to stipulate a common meaningful definition of effective size taken by a pattern for all 4 patterns (e.g. number of pixels overtaken by the rendered pattern in an image is a good measure of size for stickers and logos but not for text, since to achieve same amount of area as stickers/logos, a text box would occupy much larger part of image). During test set synthesis, we do not perform random attribute alignment between graphics patterns and image. We list the exact definition of size for each category and statistics used during test set synthesis for each category and size level below. Example images are as shown in figure \ref{fig: test set demo}

\textbf{Stickers/logos} The \emph{effective size} of stickers/logos pattern is defined as the ratio between number of pixels overwritten by patterns in an image and the total number of pixels of that image. During synthesis, we render a random number of patterns onto image such that the total effective area is within range for that size level. 
\begin{itemize}
    \item \textbf{Small}: size \~{} Uniform(0.001, 0.016), number of patterns $\in [1, 2]$
    \item \textbf{Medium}: size \~{} Uniform(0.016, 0.064), number of patterns $\in [1, 4]$
    \item \textbf{Large}: size \~{} Uniform(0.064, 0.4), number of patterns $\in [1, 12]$
\end{itemize}

\textbf{Lines} The \emph{effective size} of lines is defined as the ratio between width of a line and the length of shorter side of an image. Note although we use term line, the pattern rendered includes both line segment as well as free-form curves.
\begin{itemize}
    \item \textbf{Small}: size \~{} Uniform(0.008, 0.02), number of patterns $\in [1, 10]$
    \item \textbf{Medium}: size \~{} Uniform(0.02, 0.06), number of patterns $\in [1, 10]$
    \item \textbf{Large}: size \~{} Uniform(0.06, 0.15), number of patterns $\in [1, 6]$
\end{itemize}

\textbf{Text} The \emph{effective size} of text is dictated by both the size of a glyph (roughly, width of a single character relative to image width) and the total area of bounding box of text in the image since we could have a very long string of small font text that occupies entire image or a very large single character. We used following number during synthesis
\begin{itemize}
    \item \textbf{Small}: glyph size \~{} Uniform(0.05, 0.1), bounding box size $\in [0.002, 0.016]$
    \item \textbf{Medium}: glyph size \~{} Uniform(0.1, 0.2), bounding box size $\in [0.016, 0.25]$
    \item \textbf{Large}: glyph size \~{} Uniform(0.15, 0.4), bounding box size $\in [0.25, 0.6]$
\end{itemize}

\section{Ablation Study of Proposed Solution}
\label{sec: ablation study}
We perform ablation study on each element of our proposed solution and show the effectiveness of each proposed element in boosting the performance.
\subsection{Proposed Attribute Randomization}
For data synthesis, we compare our random graphics pattern attribute blending scheme with 1) a simple and straightforward attribute random perturbation, 2) no attribute perturbation. In the simple perturbation scheme, we randomly adjust pattern attributes (e.g. make pattern brighter/darker, less/more saturated, etc.) irrespective of the corresponding local/global attribute at the target location in the image

\begin{table}[ht]
\centering
\begin{tabular}{ccccc}
\hline
\hline
 & \multicolumn{4}{c}{\textbf{Overall Performance on Test Set}}\\ 
 & mIoU & MAE & $F_{0.3}$ & $F_{2}$ \\ \hline
\textbf{Proposed} & \textbf{0.745} & \textbf{0.022} & \textbf{0.873} & \textbf{0.863} \\
\textbf{Simple} & 0.667 & 0.031 & 0.843 & 0.793 \\
\textbf{None} & 0.606 & 0.042 & 0.799 & 0.743 \\
\hline\hline
\end{tabular}
\caption{Quantitative comparison of our proposed attribute randomization versus a simple randomization scheme and no randomization}
\label{tab:data synthesis ablation}
\end{table}

\subsection{Impact of JPEG Compression}
\begin{figure}[!]
\centering
\includegraphics[width=0.75\linewidth]{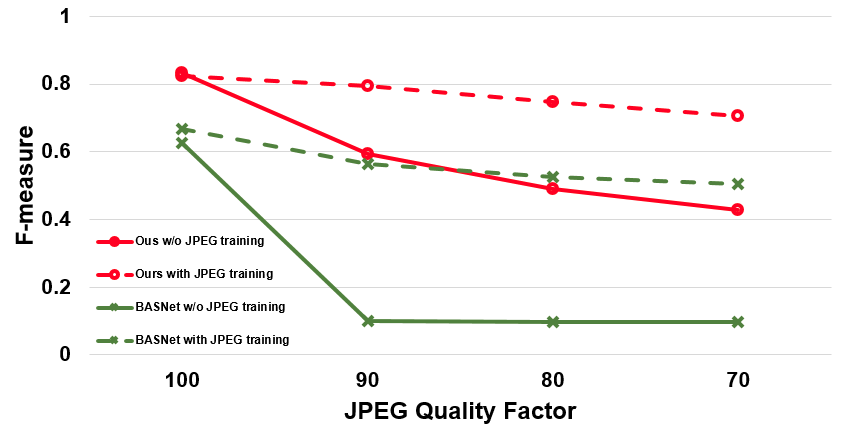}
\caption{$F_{0.3}$ score as a function of the JPEG compression quality factor. A lower quality factor means less quantization levels on DCT coefficient and therefore stronger blocking artifacts and poorer image quality. Only performance degradation of our proposed model and BASNet is shown for figure clarity, but all models are heavily influenced.}
\label{fig: f-measure vs jpeg compression}
\vspace{-2ex}
\end{figure}

Albeit already discussed in other scenarios (e.g. Photoshop face manipulation detection by Wang et al. \cite{wang2019detecting}), models trained for our problem heavily relies on low-level image features, and therefore we emphasize the necessity of applying JPEG compression to training data for achieving reasonable generalization to real online images. To this end, we show in figure \ref{fig: f-measure vs jpeg compression} the performance (F-measure) of different models, trained with/without JPEG compression, on test set under different JPEG compression settings. As can be seen, without JPEG compression during training, model performance is severely impaired by blocking artifacts introduced by JPEG compression even at quality level 90.

\subsection{Network Architecture}
We investigate the effectiveness of cascading scheme by training variants of our multi-scale cascade network with 1-level (i.e. no cascade), multi-scale input 2/3/4-level cascade, and single scale input 3-level cascade (SS 3-level). For fair comparison, we keep total size (number of parameters) of feature extractors of each network roughly the same. Specifically, we used 12-resblocks for 1-level network feature extractor, 6-resblocks for each of two feature extractors of 2-level network, 4-resblocks for each of three feature extractors of 3-level network, and 3-resblocks for each of four feature extractors of 4-level network. As shown in table \ref{tab:architecture ablation}, although 1-level network demonstrates on-par/slightly better performance on large patterns, our usage of cascading scheme improves consistency of performance across different pattern sizes. The 3-level cascade network that only uses single scale input (input at original resolution of 256x256) achieves similar performance as compared to its multi-scale input counterpart. but it should be noted that: 1) multi-scale version still has better consistency across patterns sizes, 2) single scale version generates very large feature maps at all hidden layers and the computation budget required is much higher at both training and inference time. The extra computation budget limits the possibility of using a larger backbone at each cascade level or cascading more levels.

\begin{table}[ht]
\centering
\begin{tabular}{c|cc|cc|cc|cc}
\hline
\hline
 & \multicolumn{8}{c}{\textbf{Performance on Test Set}}\\ 
 \hline
 & \multicolumn{2}{c|}{Small} & \multicolumn{2}{c|}{Medium} & \multicolumn{2}{c|}{Large} & \multicolumn{2}{c}{Gap between Large and Small}\\
 & mIoU & $F_{0.3}$ & mIoU & $F_{0.3}$ & mIoU & $F_{0.3}$ & $|\Delta mIoU|$ & $|\Delta F_{0.3}|$ \\ \hline
\textbf{1-level} & 0.545 & 0.695 & 0.735 & 0.856 & 0.809 & 0.923 & 0.262 & 0.228\\ 
\textbf{2-level} & 0.582 & 0.732 & 0.790 & 0.910 & 0.801 & 0.920 & 0.219 & 0.188\\
\textbf{3-level} & \textbf{0.664} & \textbf{0.809} & \textbf{0.796} & \textbf{0.913} & 0.782 & 0.902 & 0.118 & 0.093\\
\textbf{4-level} & 0.616 & 0.782 & 0.757 & 0.893 & 0.724 & 0.869 & \textbf{0.108} & \textbf{0.087}\\
\textbf{SS 3-level} & 0.645 & 0.788 & 0.781 & 0.906 & \textbf{0.820} & \textbf{0.926} & 0.175 & 0.138 \\
\hline\hline
\end{tabular}
\vspace{2ex}
\caption{Quantitative comparison of models with different number of cascade levels. Note that 3-level model performs best on small and medium size patterns and has smallest performance gap between performance on large and small pattern}
\label{tab:architecture ablation}
\end{table}

\subsection{Multi-stage training}
Deep supervision has been demonstrated to be effective in training segmentation networks and is widely adopted by numerous works \cite{DeeplySupervisedSalientObjectDetectionwithShortConnections, BASNet:Boundary-AwareSalientObjectDetection, SalientObjectDetectionwithRecurrentFullyConvolutionalNetworks, AStagewiseRefinementModelforDetectingSalientObjectsinImages, He_2019_CVPR}. Our multi-stage training can be perceived as a special form of deep supervision. Therefore, we compare our training scheme with training entire multi-scale cascade network jointly with supervision on each sub-network. Table \ref{tab:multi-stage training} shows that our multi-stage training scheme enables our cascade network to achieve better performance.

\begin{table}[h]
\centering
\begin{tabular}{c|cccc}
\hline
\hline
 & \multicolumn{4}{c}{\textbf{Overall Performance on Test Set}}\\ 
 & mIoU & MAE & $F_{0.3}$ & $F_{2}$ \\ \hline
\textbf{Multi-stage} & \textbf{0.745} & \textbf{0.022} & \textbf{0.873} & \textbf{0.863} \\ 
\textbf{Joint} & 0.565 & 0.047 & 0.755 & 0.722 \\
\hline\hline
\end{tabular}
\vspace{2ex}
\caption{Quantitative comparison of multi-stage training versus simultaneous joint training}
\label{tab:multi-stage training}
\end{table}

\section{Wild Data Test}
\label{sec: wild data test}
\begin{figure}[ht]
\setlength\tabcolsep{1pt}
\renewcommand{\arraystretch}{0.5}
\centering
\begin{tabular}{cc}
\includegraphics[width=0.4\linewidth]{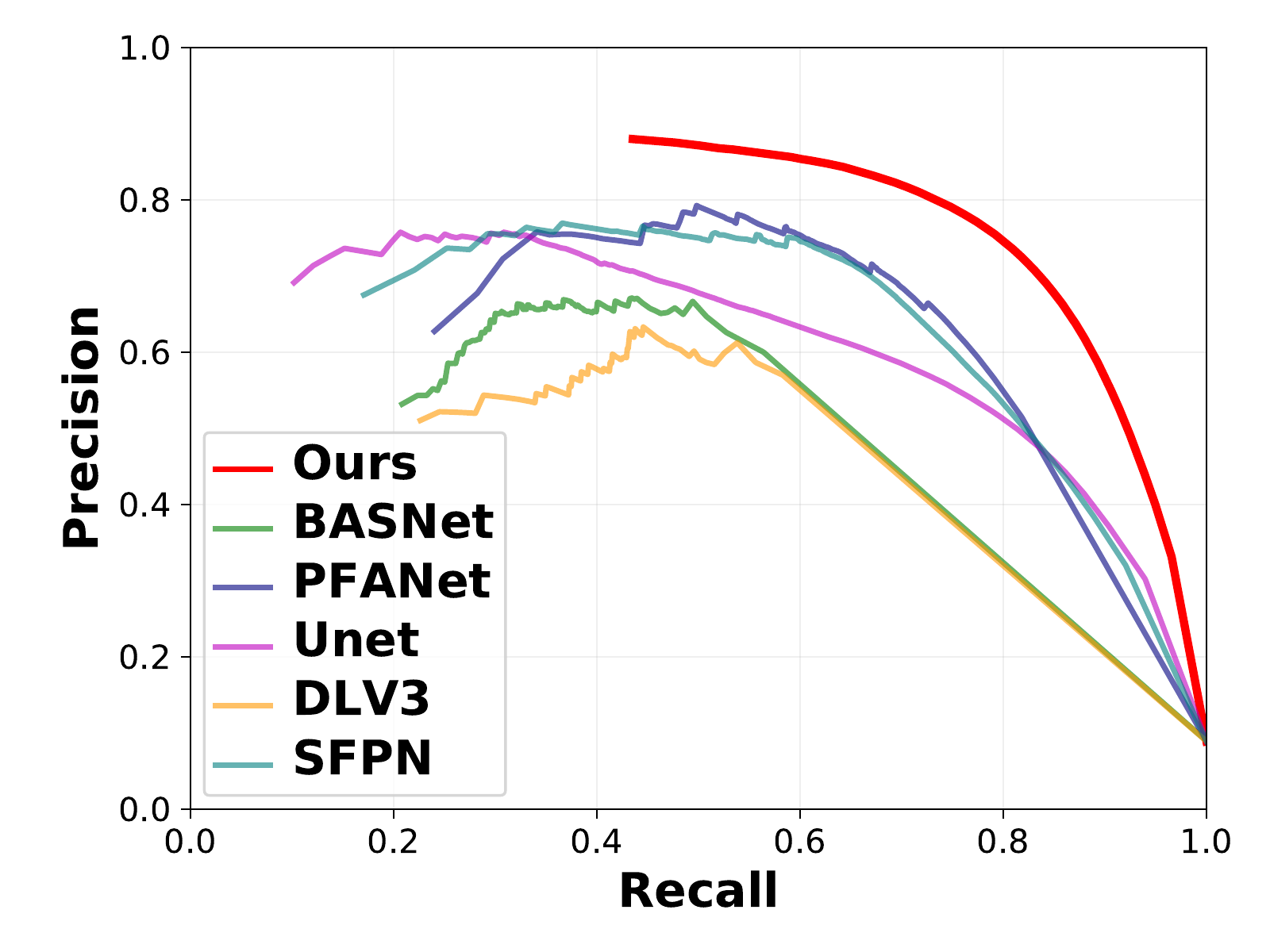}&
\includegraphics[width=0.4\linewidth]{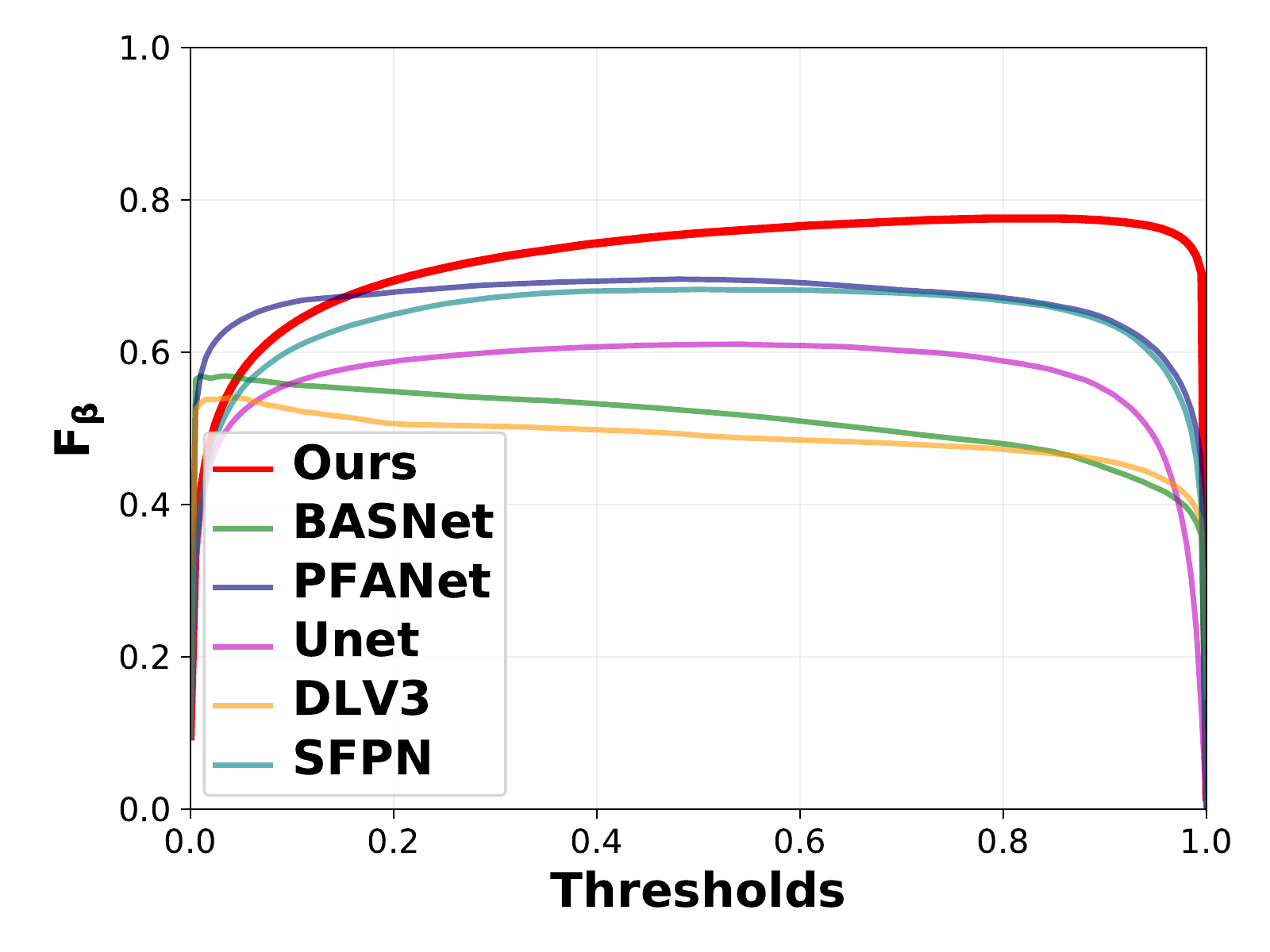}\\
(a) Precision-Recall Curve & (b) F-measure vs. Thresholds \\
\end{tabular}
\vspace{2ex}
\caption{Precision-recall and F-measure curve of each competing methods evaluated on images collected from popular social media websites. The models here are the same as the ones in the main result section of the paper (i.e. no extra training or additional data used)}
\label{fig: wild test plots}
\end{figure}

\begin{figure*}[bh]
\setlength\tabcolsep{1pt}
\renewcommand{\arraystretch}{0.5}
\centering
\begin{tabular}{cccccccc}
\includegraphics[width=0.12\linewidth]{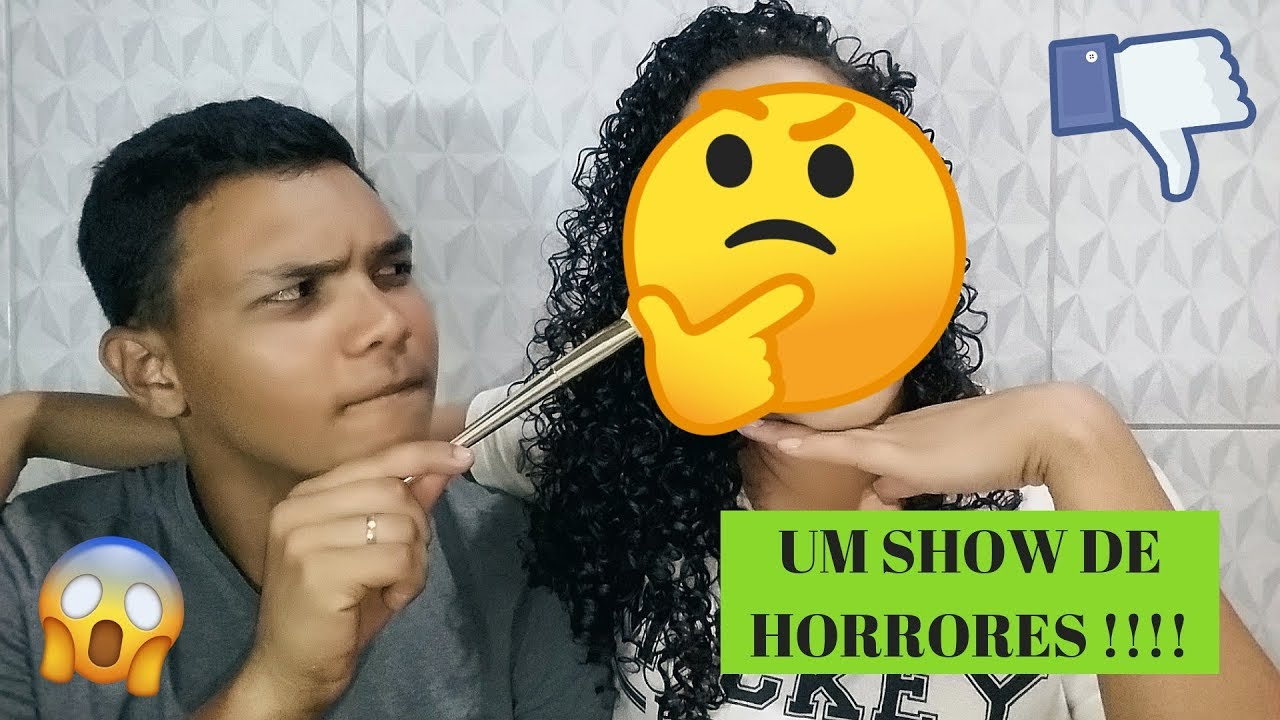} &
\includegraphics[width=0.12\linewidth]{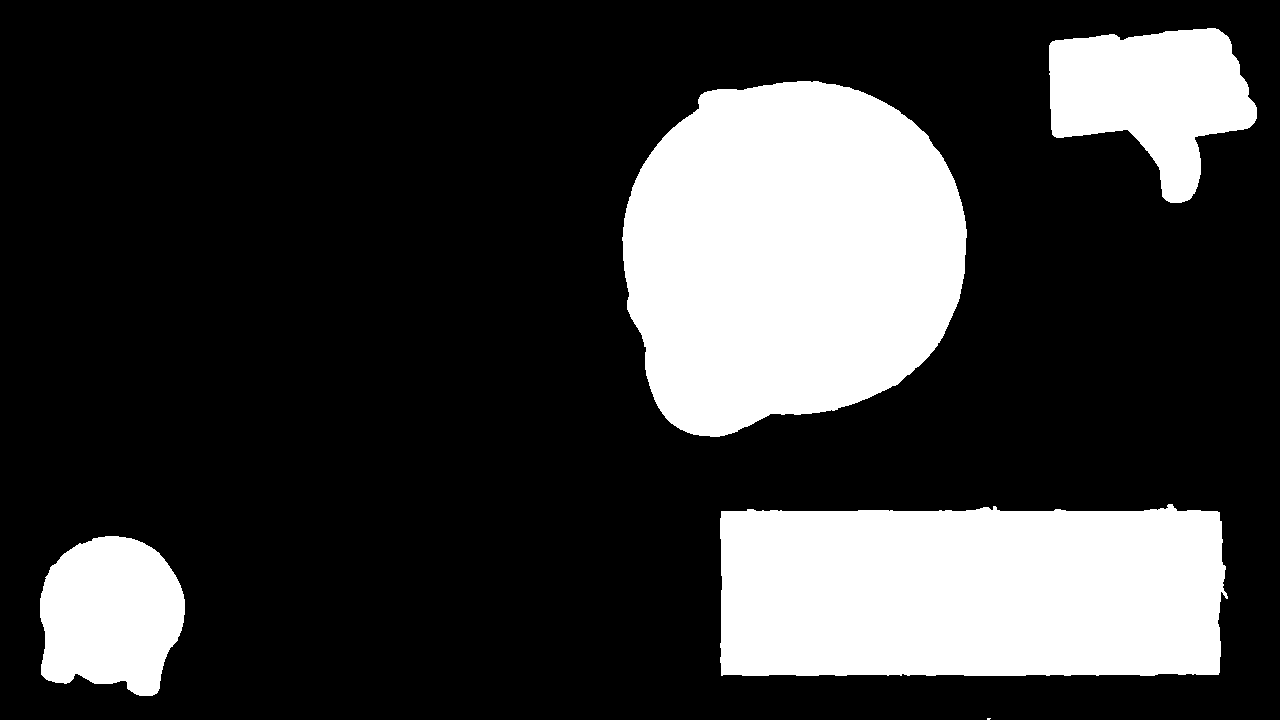} &
\includegraphics[width=0.12\linewidth]{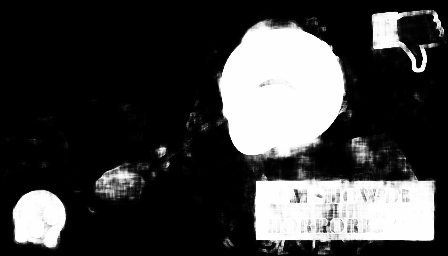} &
\includegraphics[width=0.12\linewidth]{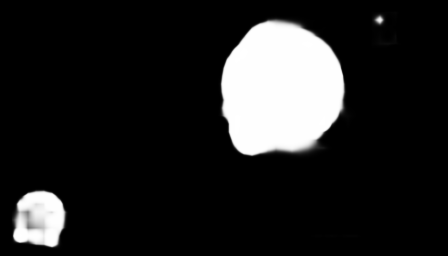} &
\includegraphics[width=0.12\linewidth]{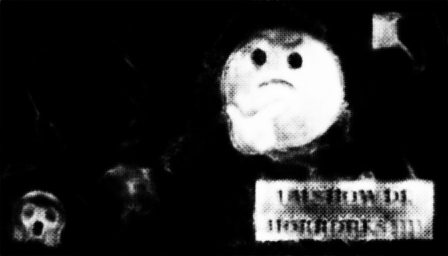} &
\includegraphics[width=0.12\linewidth]{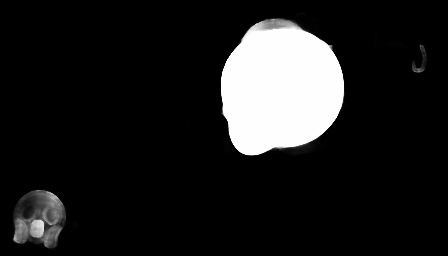} &
\includegraphics[width=0.12\linewidth]{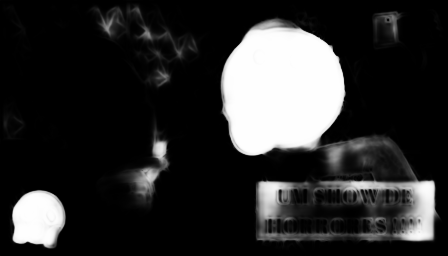} &
\includegraphics[width=0.12\linewidth]{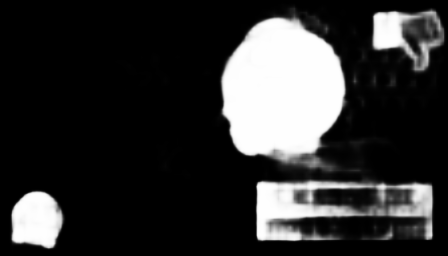} \\

\includegraphics[width=0.12\linewidth,height=0.14\linewidth]{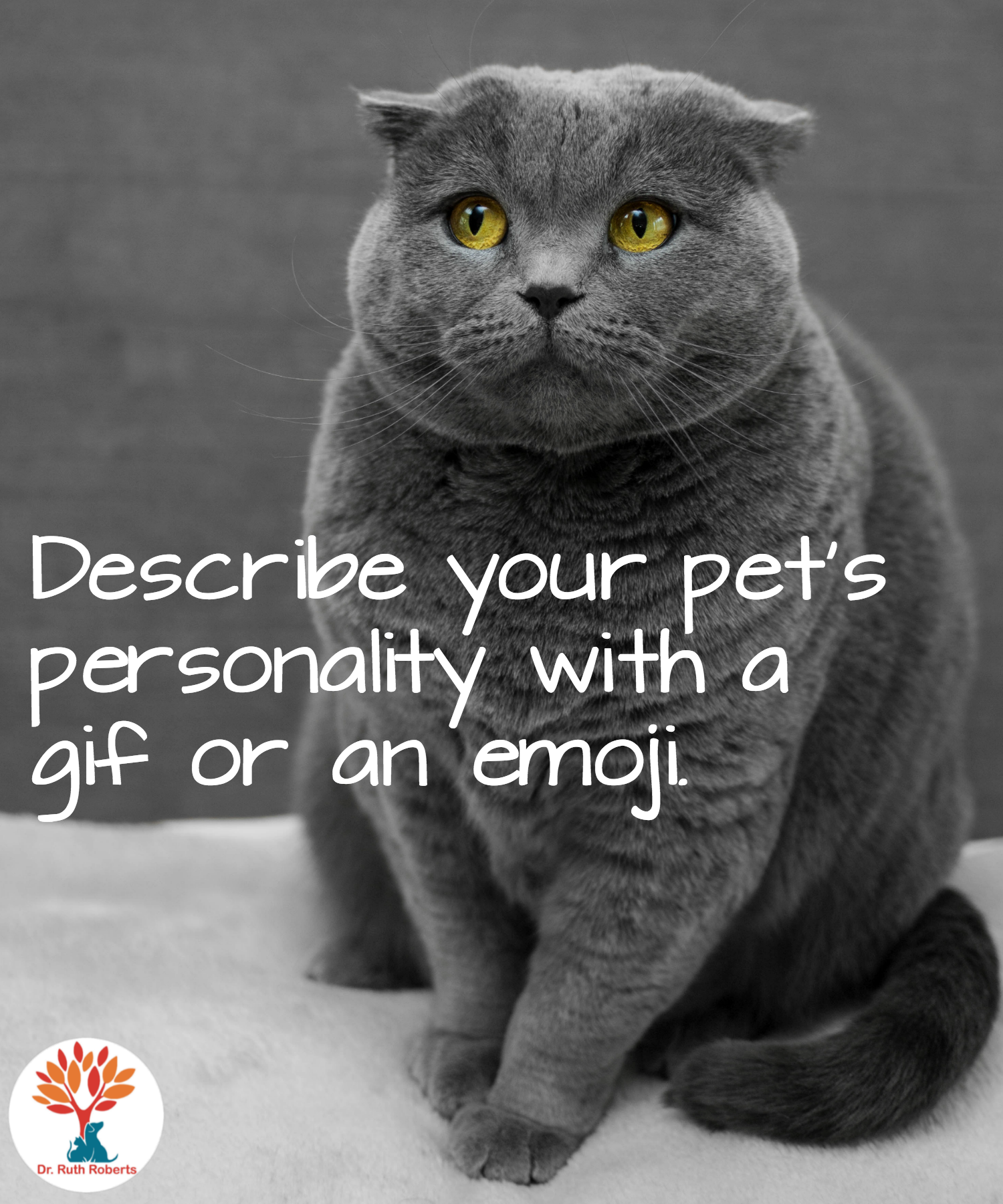} &
\includegraphics[width=0.12\linewidth,height=0.14\linewidth]{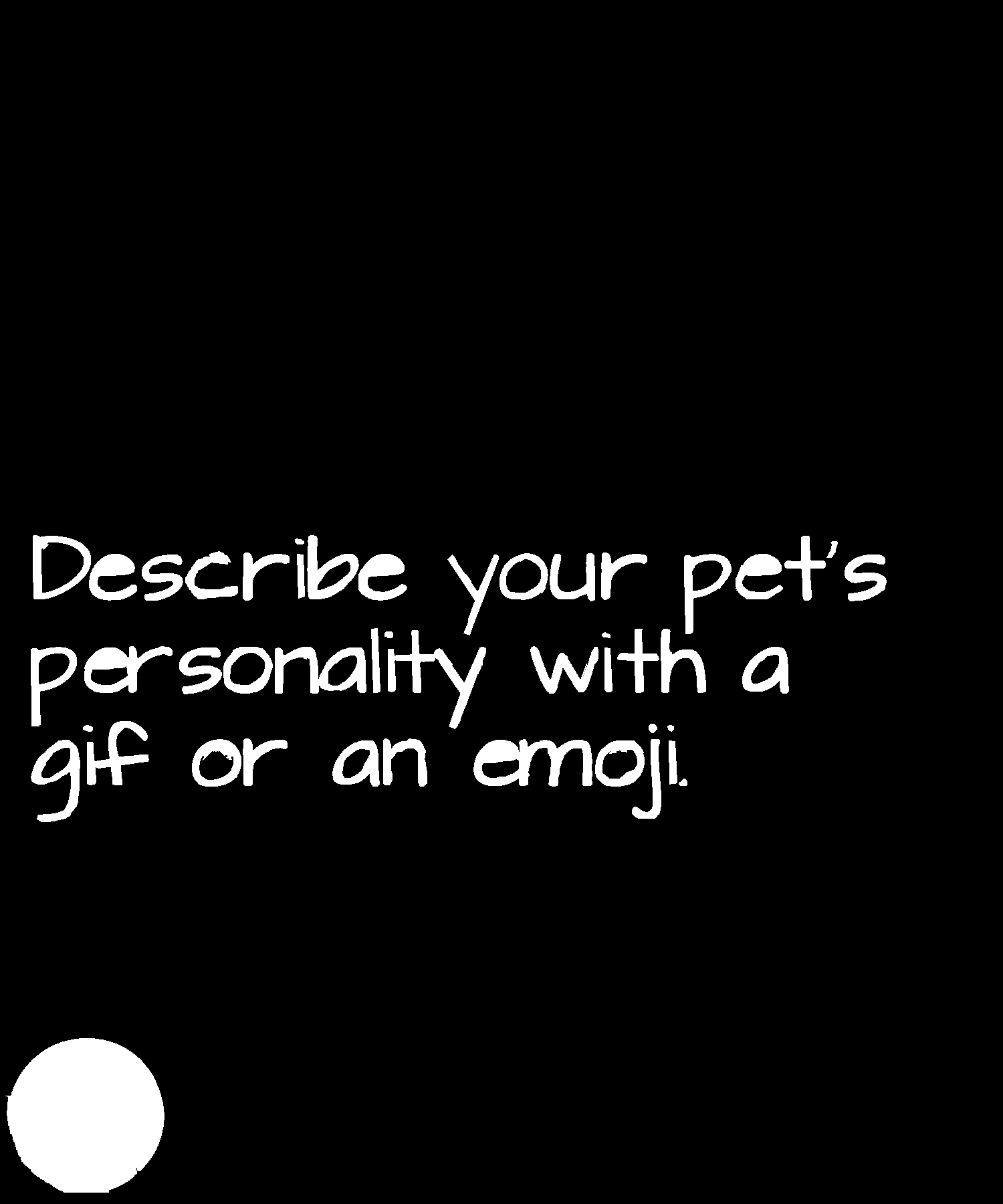} &
\includegraphics[width=0.12\linewidth,height=0.14\linewidth]{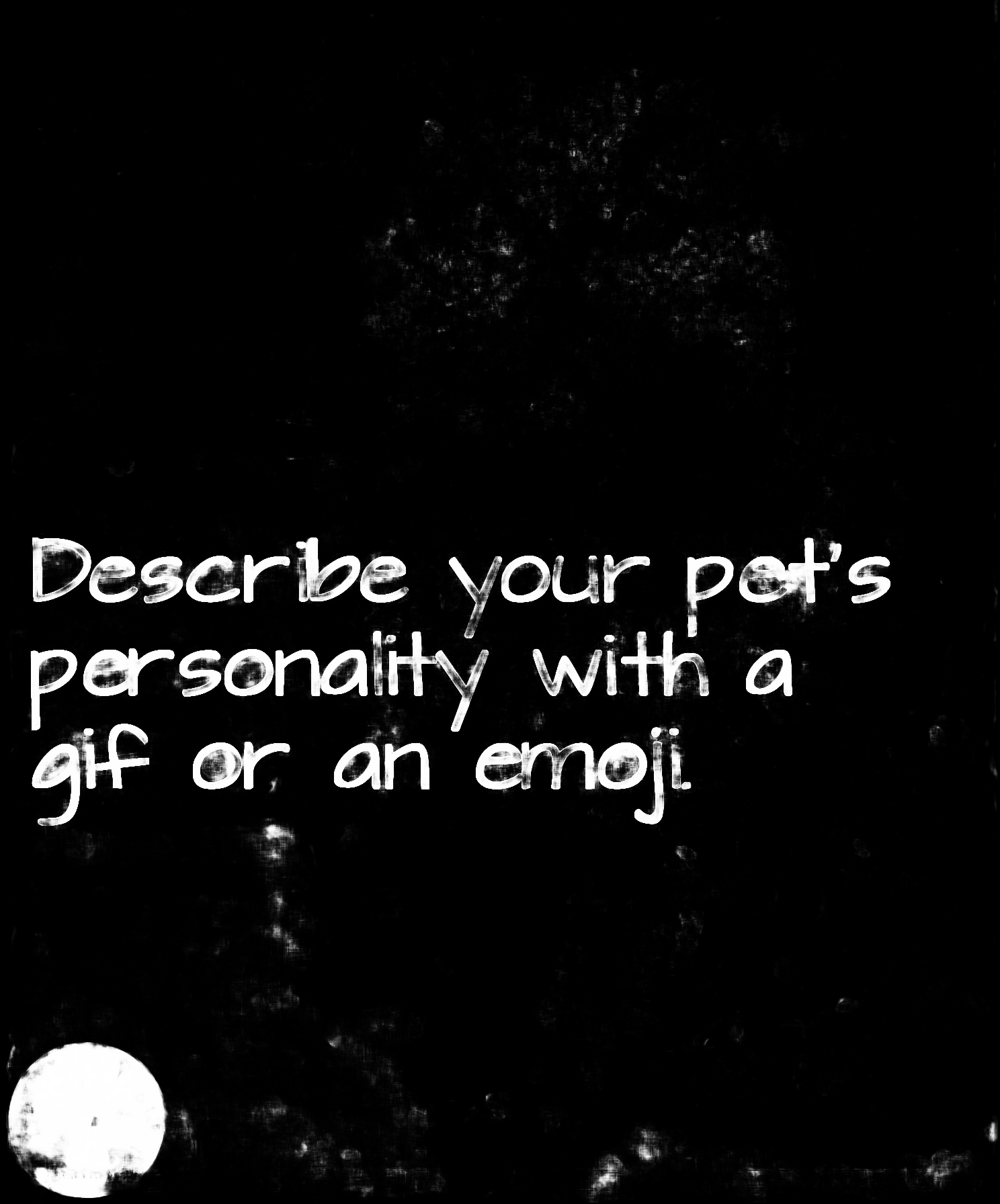} &
\includegraphics[width=0.12\linewidth,height=0.14\linewidth]{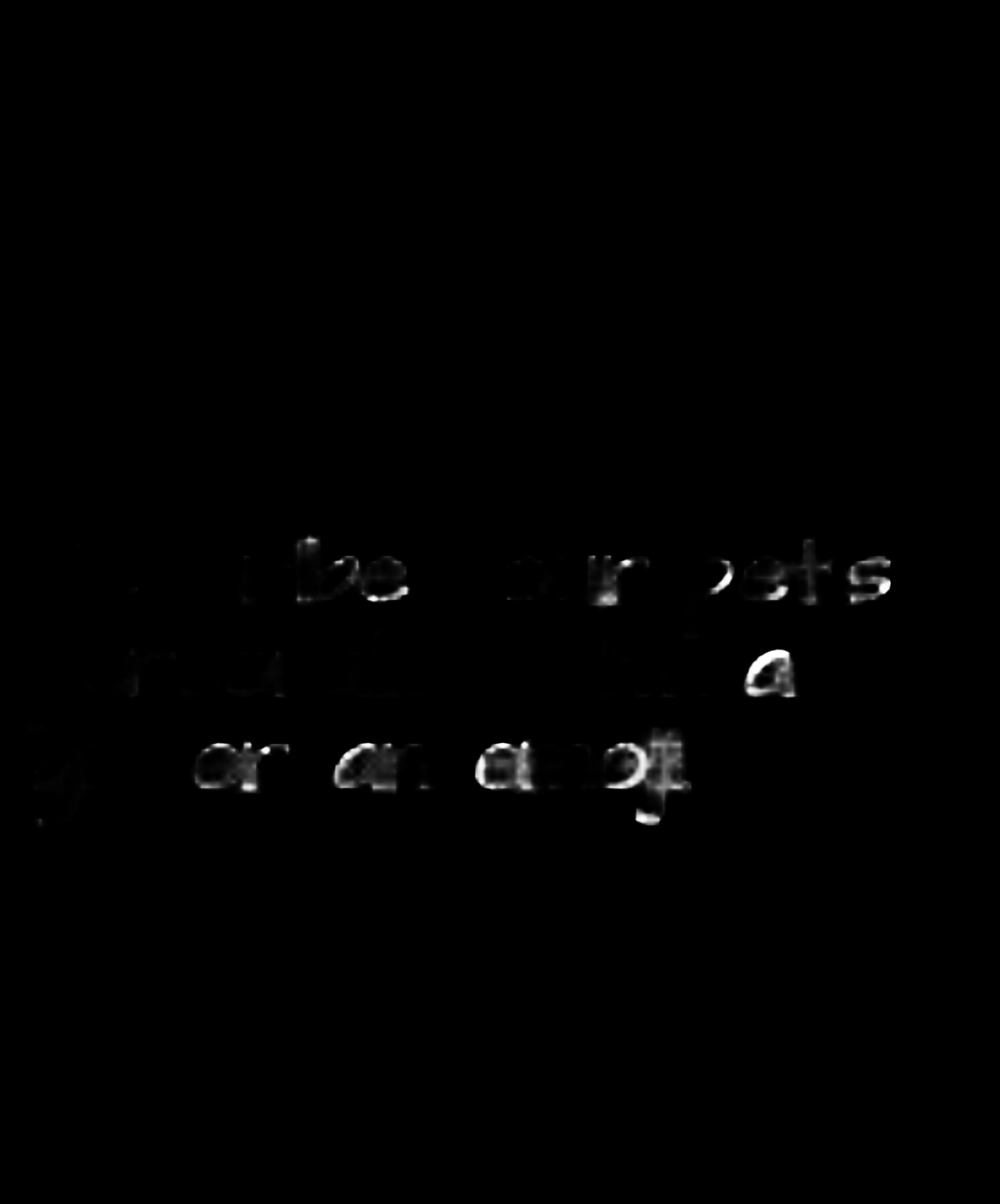} &
\includegraphics[width=0.12\linewidth,height=0.14\linewidth]{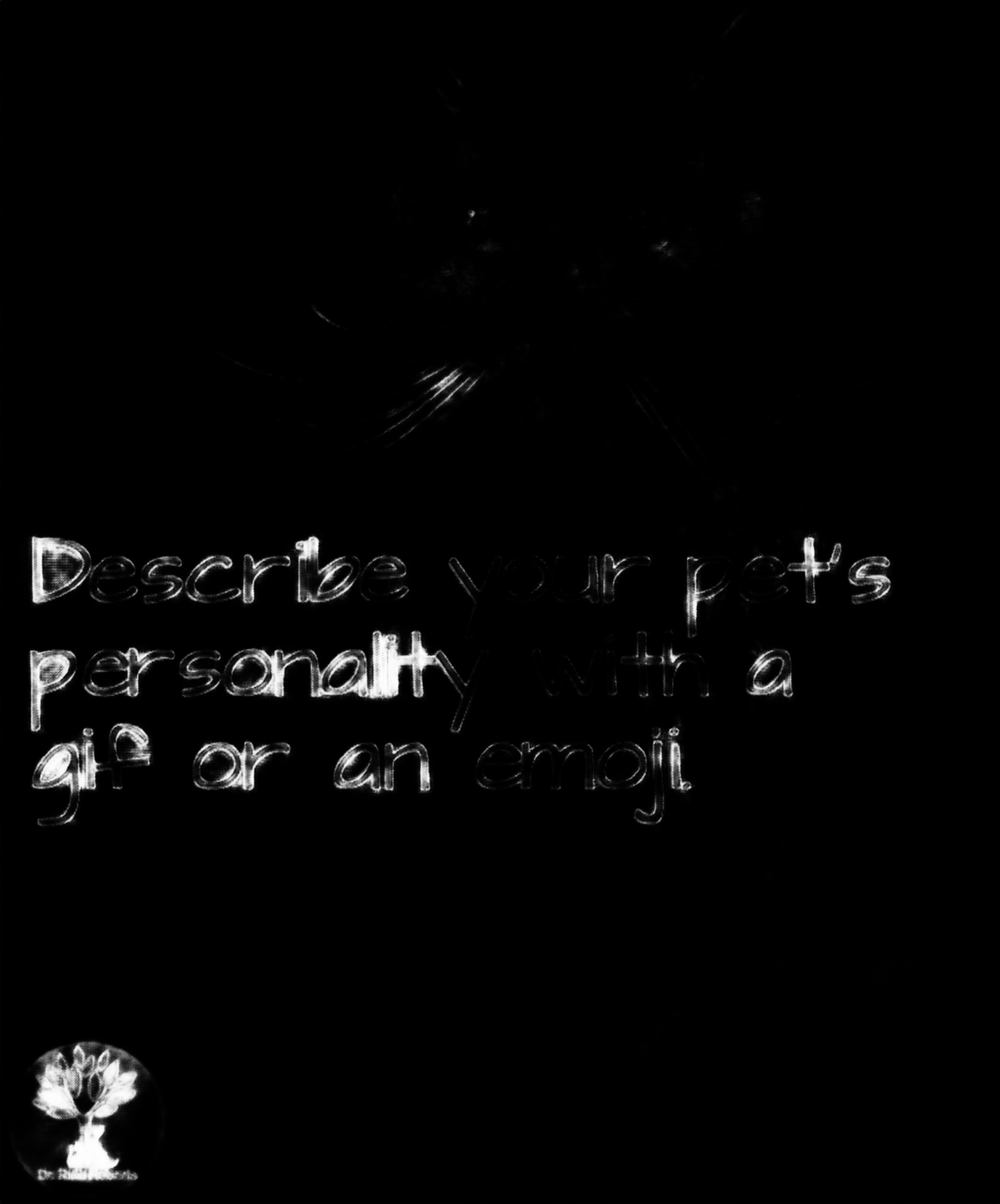} &
\includegraphics[width=0.12\linewidth,height=0.14\linewidth]{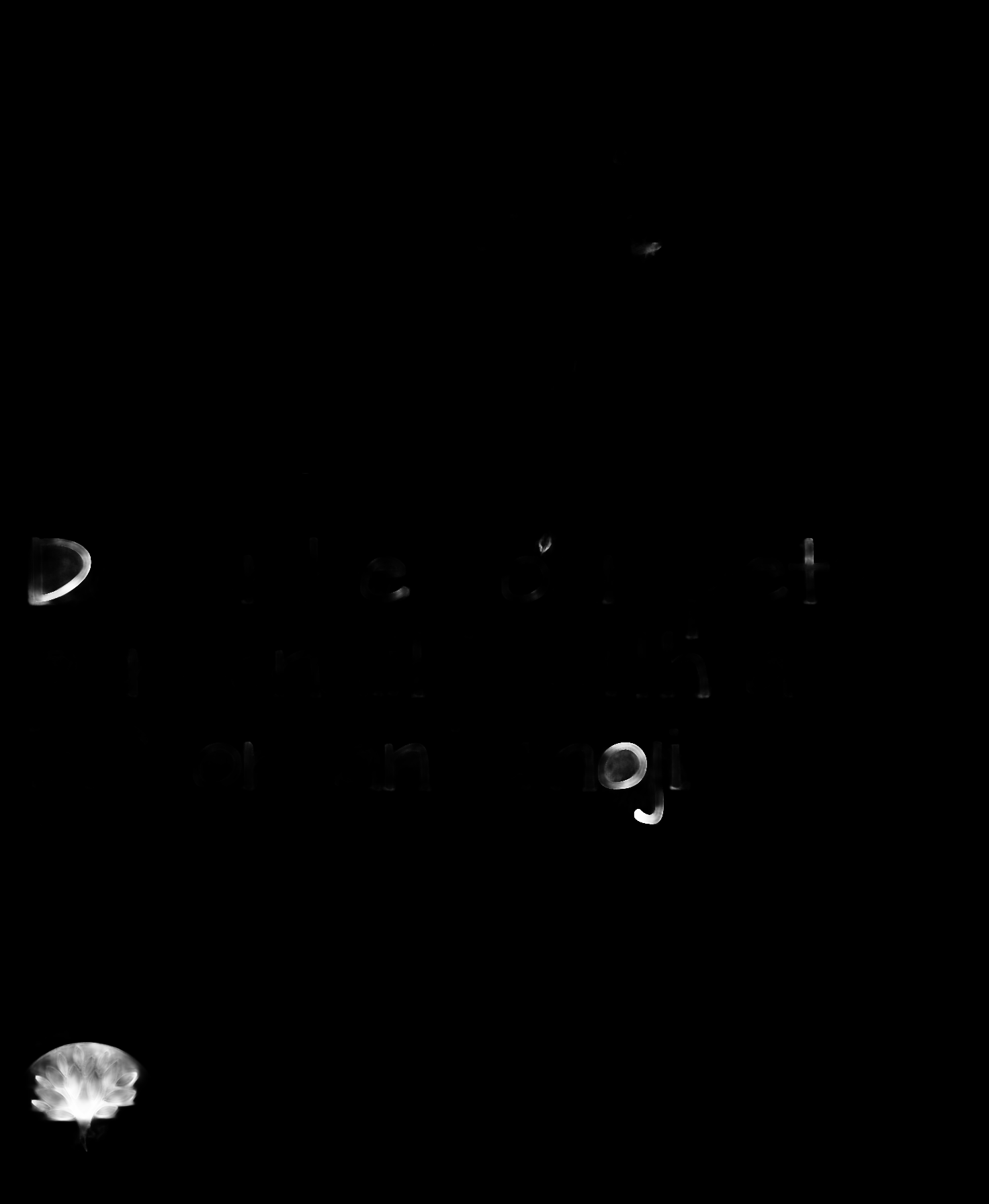} &
\includegraphics[width=0.12\linewidth,height=0.14\linewidth]{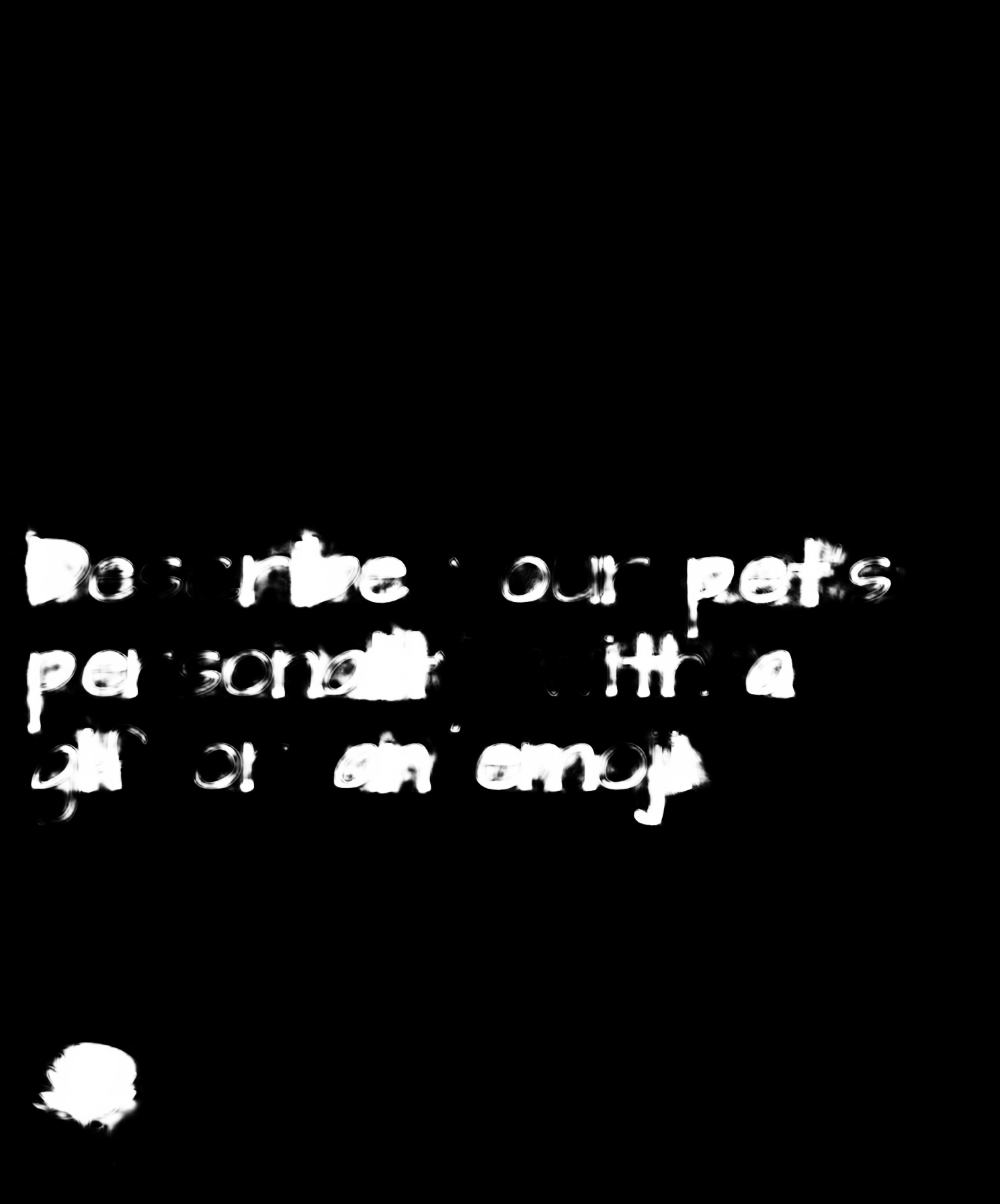} &
\includegraphics[width=0.12\linewidth,height=0.14\linewidth]{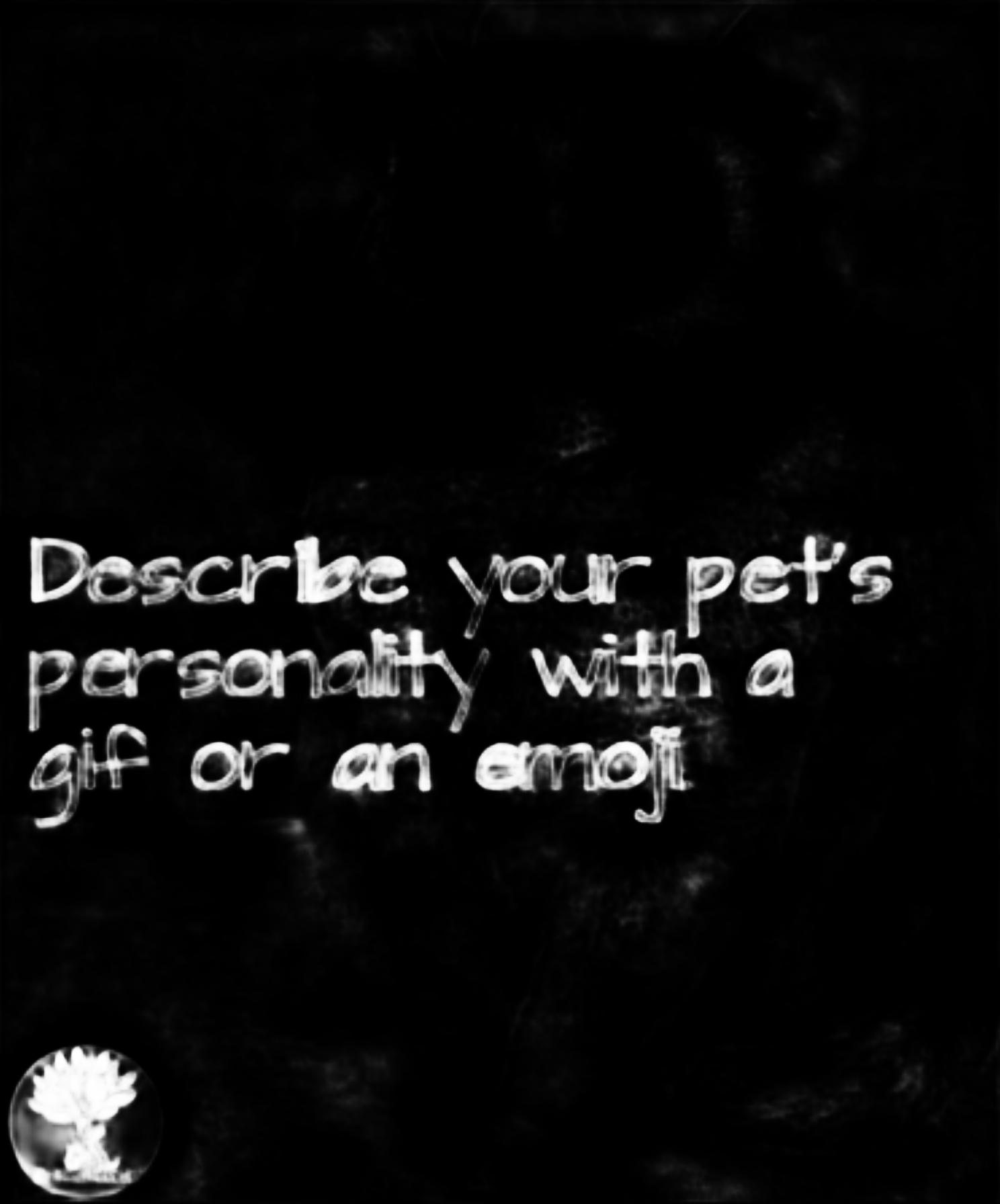} \\

\includegraphics[width=0.12\linewidth]{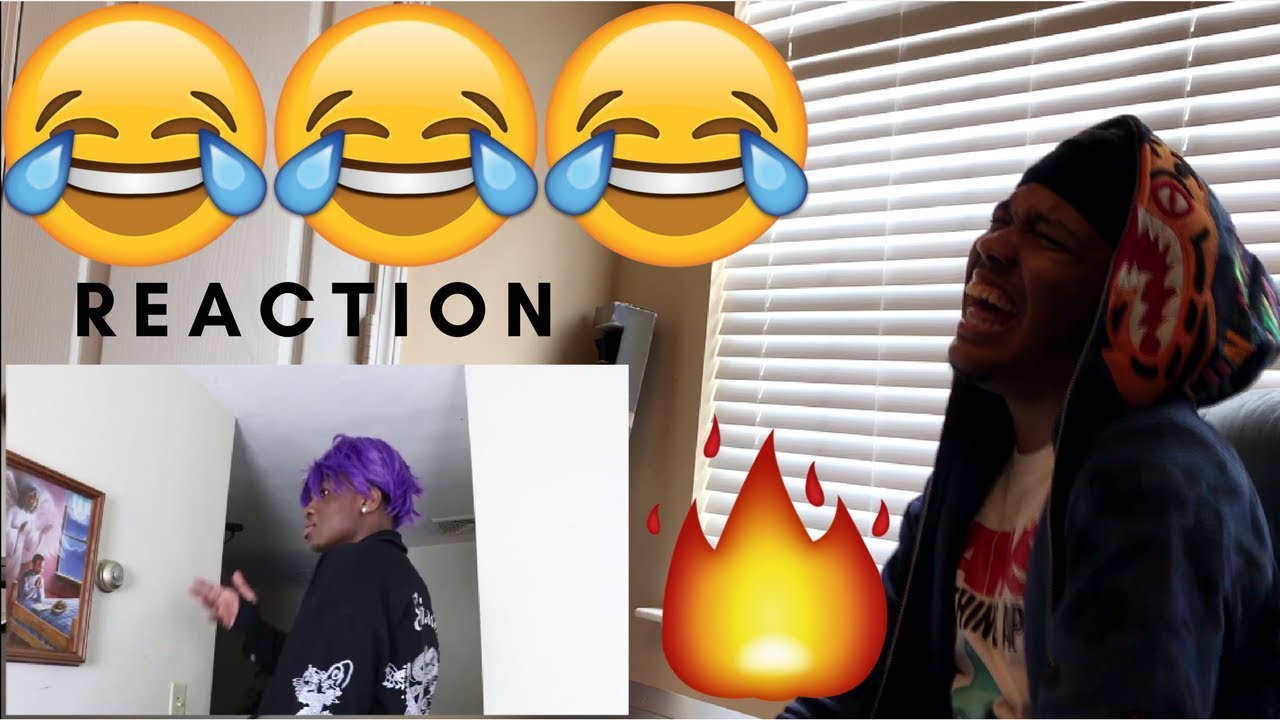} &
\includegraphics[width=0.12\linewidth]{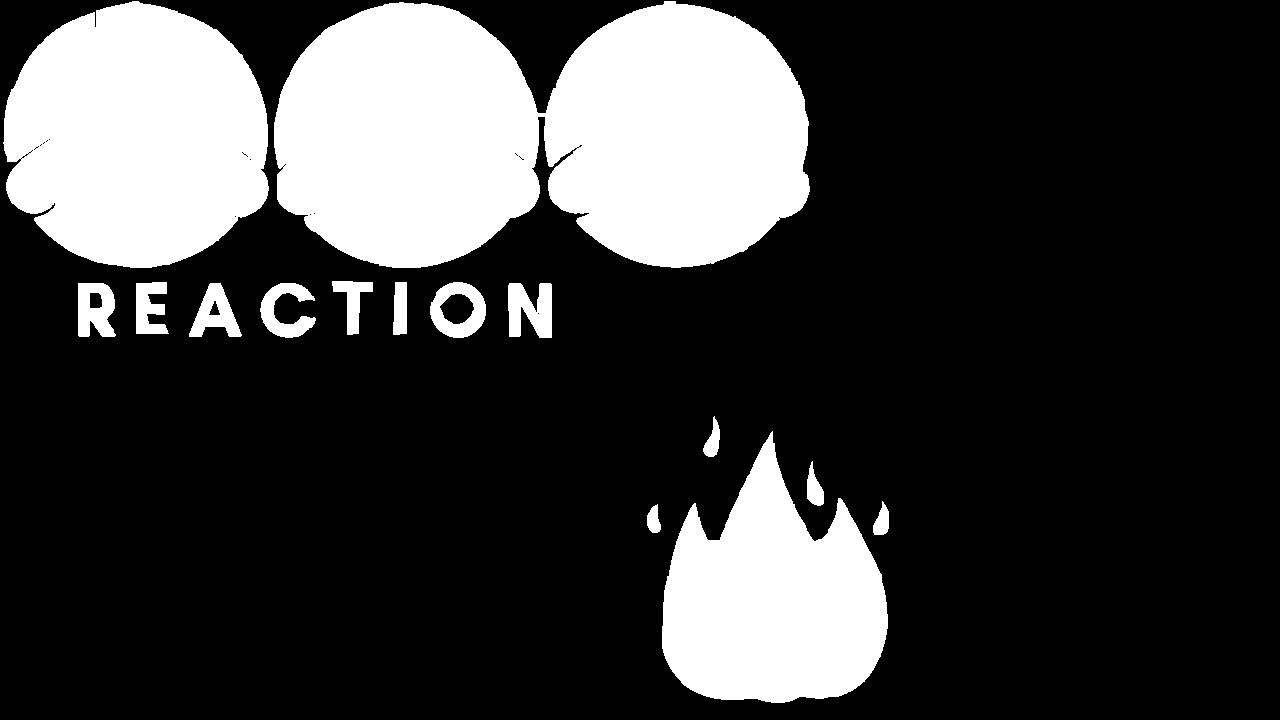} &
\includegraphics[width=0.12\linewidth]{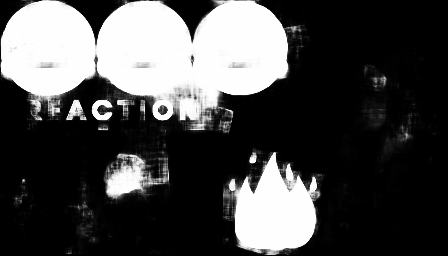} &
\includegraphics[width=0.12\linewidth]{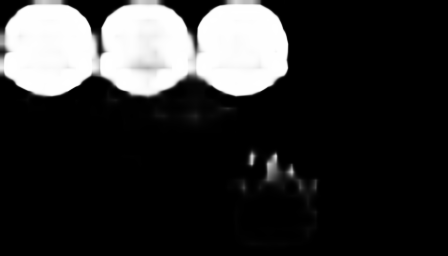} &
\includegraphics[width=0.12\linewidth]{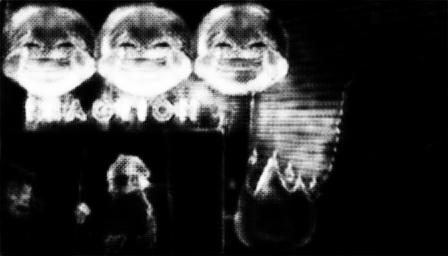} &
\includegraphics[width=0.12\linewidth]{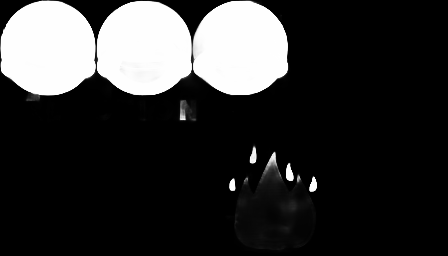} &
\includegraphics[width=0.12\linewidth]{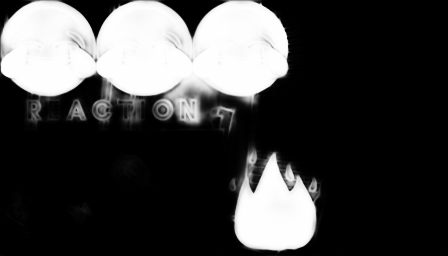} &
\includegraphics[width=0.12\linewidth]{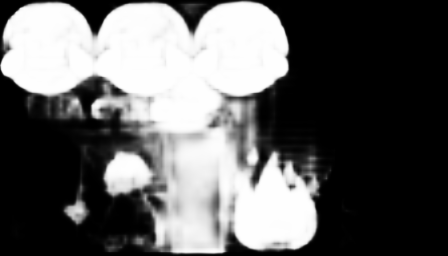} \\

\includegraphics[width=0.12\linewidth]{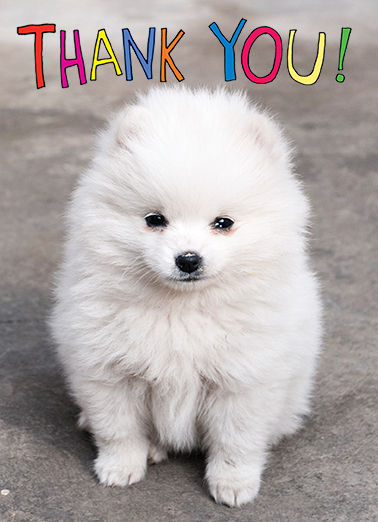} &
\includegraphics[width=0.12\linewidth]{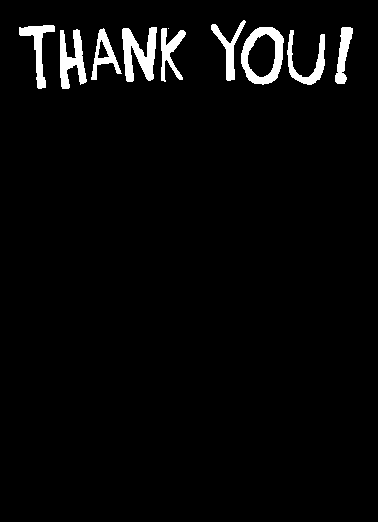} &
\includegraphics[width=0.12\linewidth]{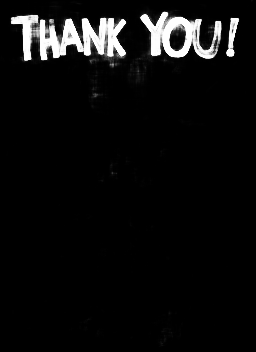} &
\includegraphics[width=0.12\linewidth]{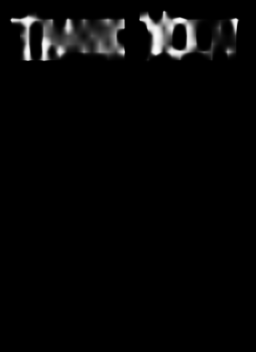} &
\includegraphics[width=0.12\linewidth]{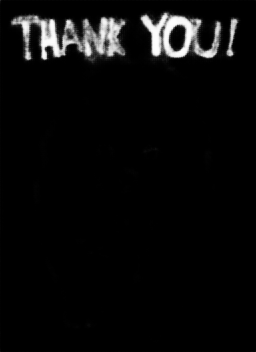} &
\includegraphics[width=0.12\linewidth]{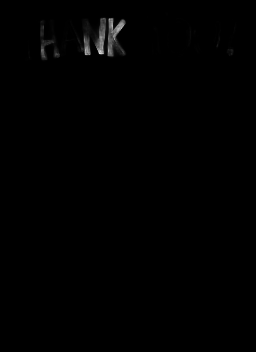} &
\includegraphics[width=0.12\linewidth]{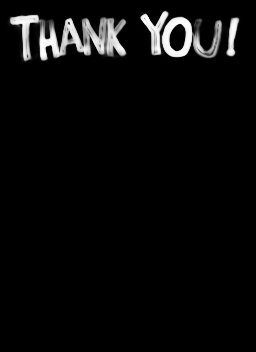} &
\includegraphics[width=0.12\linewidth]{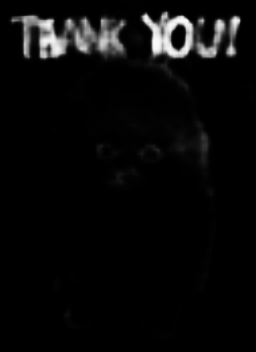} \\

\includegraphics[width=0.12\linewidth]{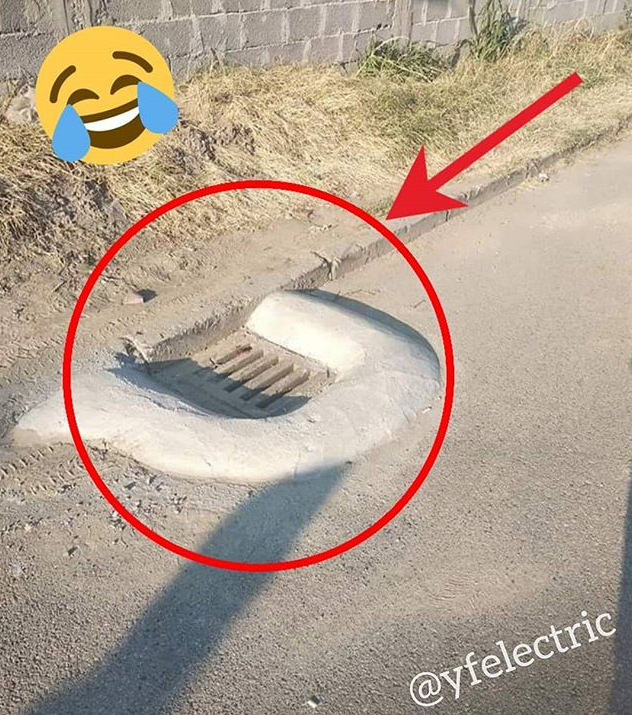} &
\includegraphics[width=0.12\linewidth]{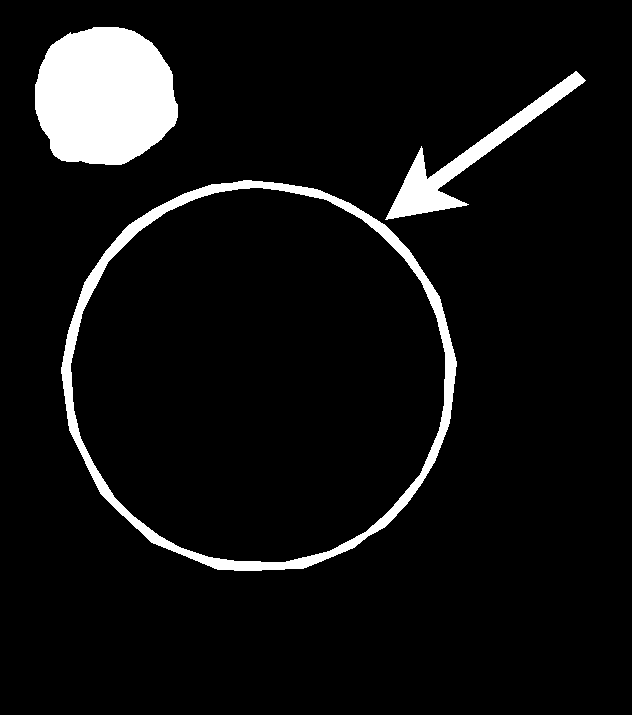} &
\includegraphics[width=0.12\linewidth]{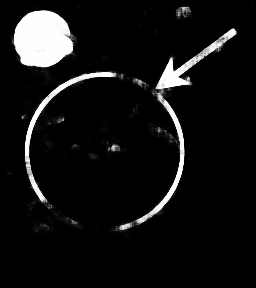} &
\includegraphics[width=0.12\linewidth]{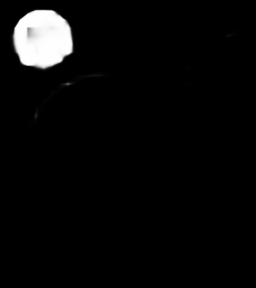} &
\includegraphics[width=0.12\linewidth]{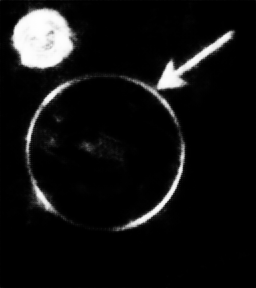} &
\includegraphics[width=0.12\linewidth]{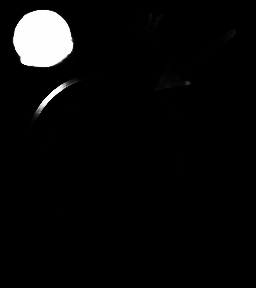} &
\includegraphics[width=0.12\linewidth]{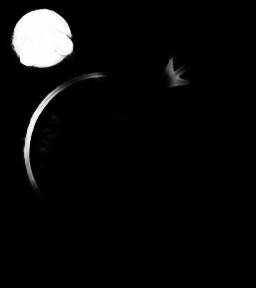} &
\includegraphics[width=0.12\linewidth]{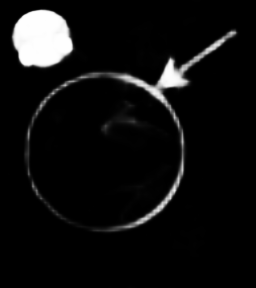} \\

\includegraphics[width=0.12\linewidth]{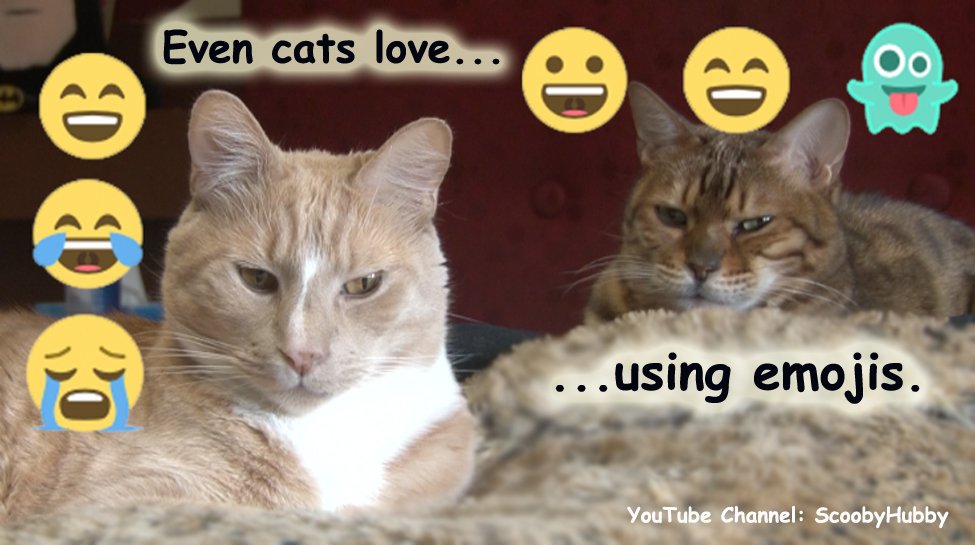} &
\includegraphics[width=0.12\linewidth]{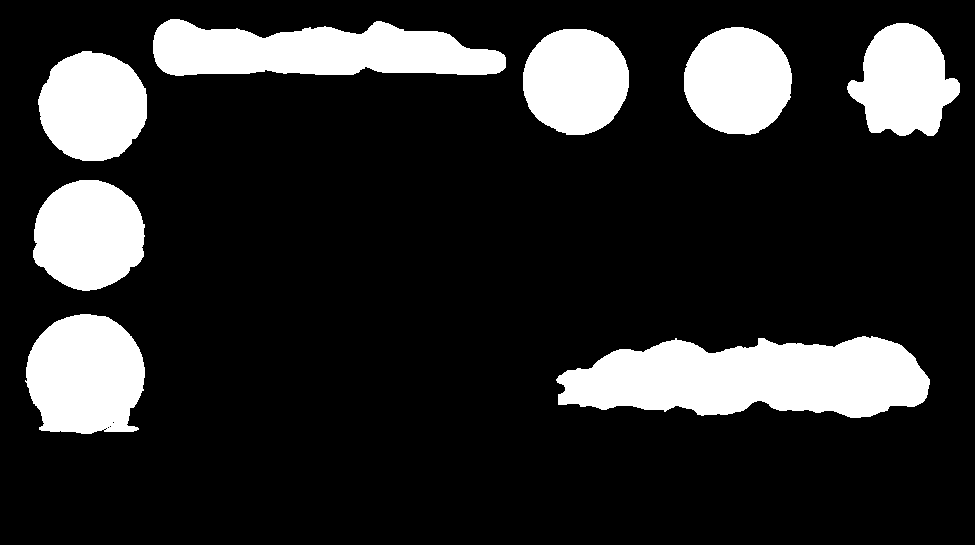} &
\includegraphics[width=0.12\linewidth]{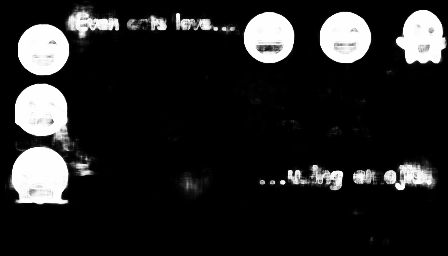} &
\includegraphics[width=0.12\linewidth]{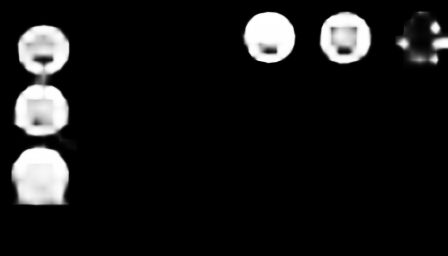} &
\includegraphics[width=0.12\linewidth]{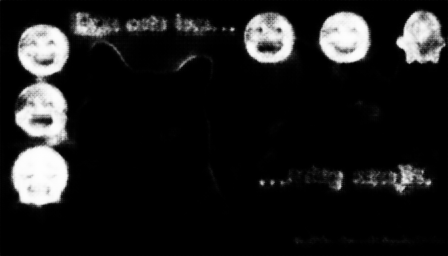} &
\includegraphics[width=0.12\linewidth]{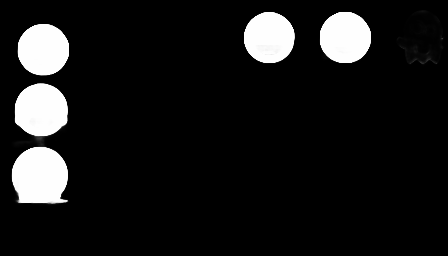} &
\includegraphics[width=0.12\linewidth]{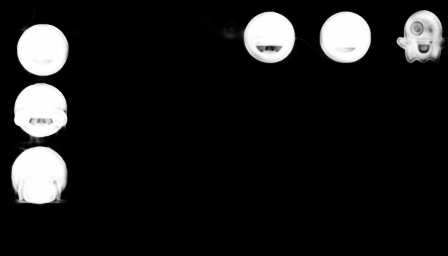} &
\includegraphics[width=0.12\linewidth]{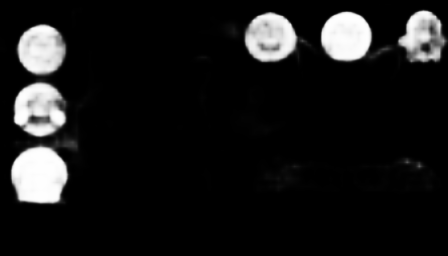} \\

\includegraphics[width=0.12\linewidth]{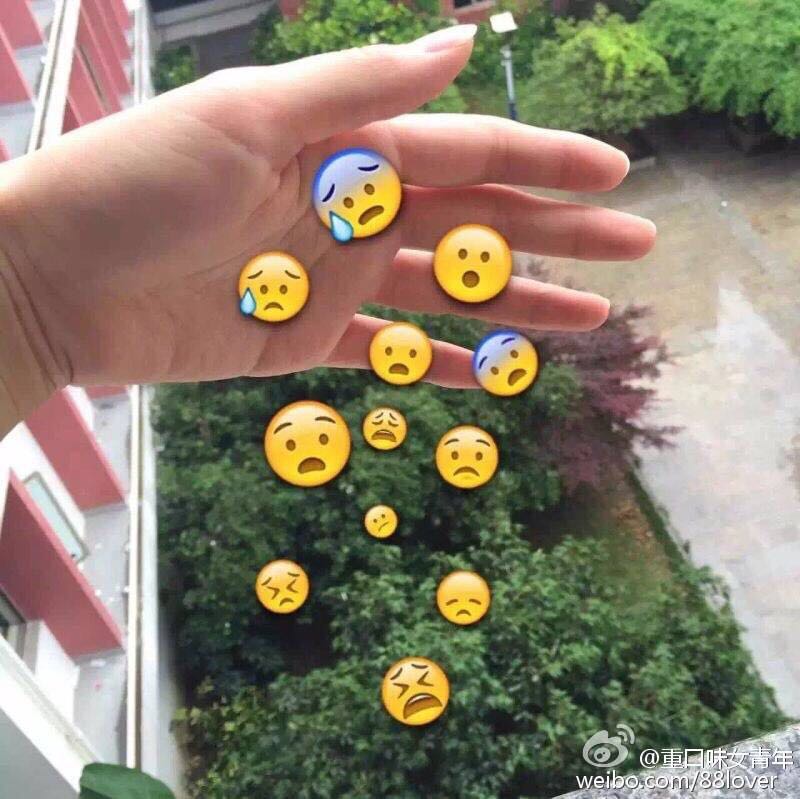} &
\includegraphics[width=0.12\linewidth]{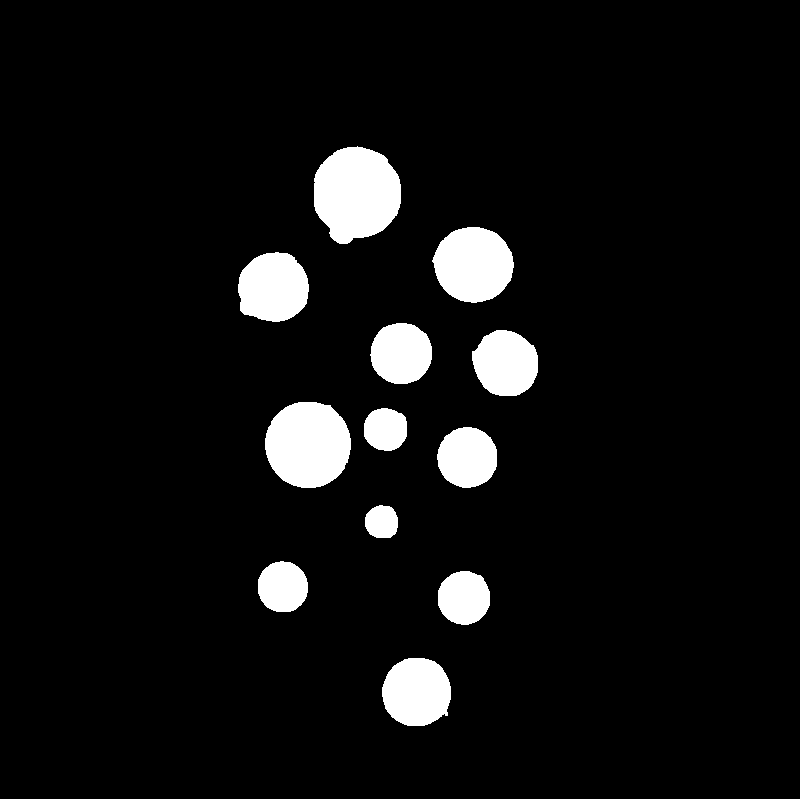} &
\includegraphics[width=0.12\linewidth]{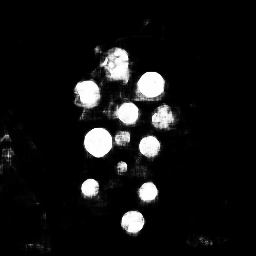} &
\includegraphics[width=0.12\linewidth]{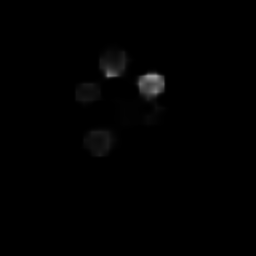} &
\includegraphics[width=0.12\linewidth]{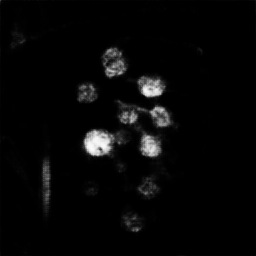} &
\includegraphics[width=0.12\linewidth]{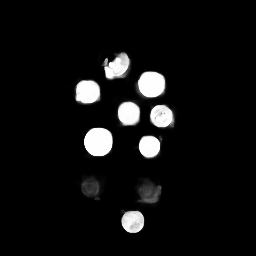} &
\includegraphics[width=0.12\linewidth]{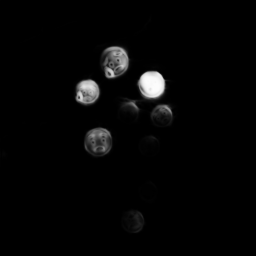} &
\includegraphics[width=0.12\linewidth]{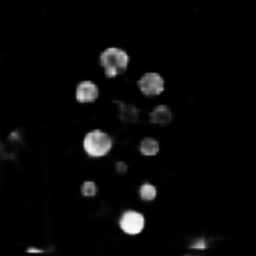} \\

\includegraphics[width=0.12\linewidth]{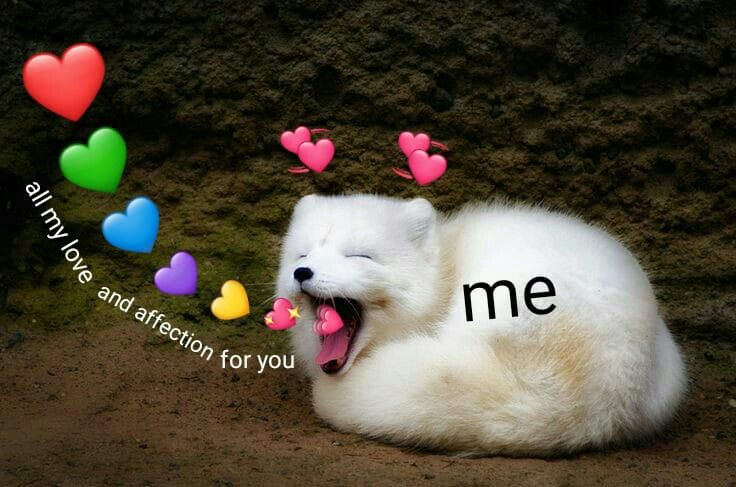} &
\includegraphics[width=0.12\linewidth]{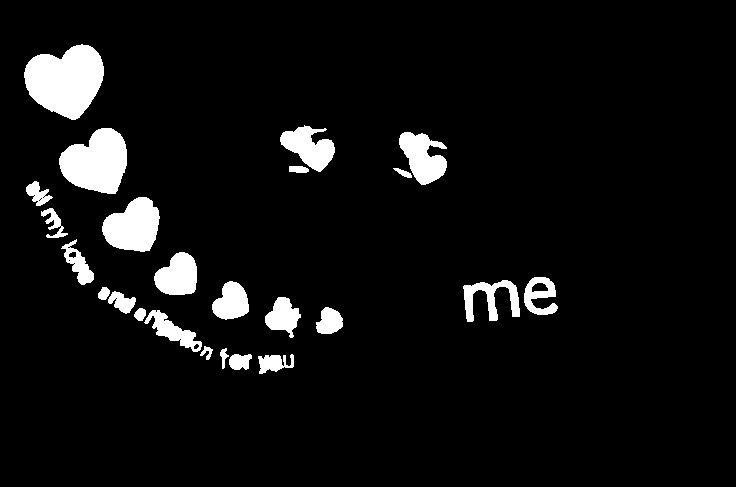} &
\includegraphics[width=0.12\linewidth]{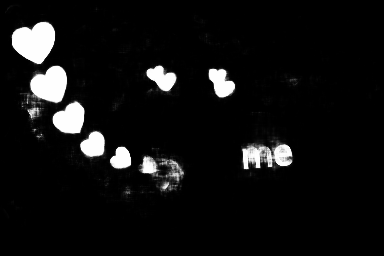} &
\includegraphics[width=0.12\linewidth]{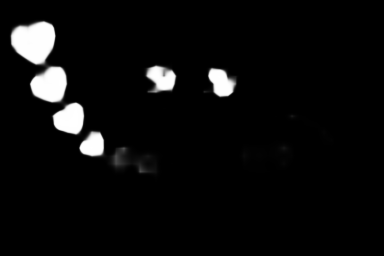} &
\includegraphics[width=0.12\linewidth]{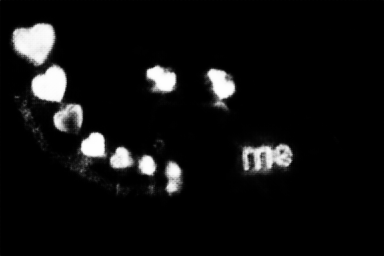} &
\includegraphics[width=0.12\linewidth]{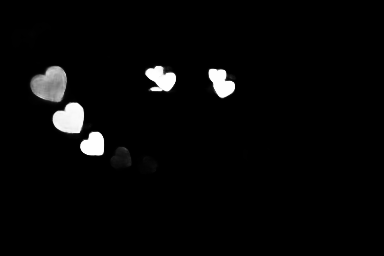} &
\includegraphics[width=0.12\linewidth]{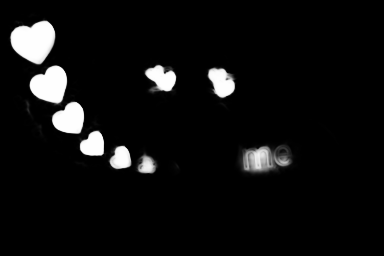} &
\includegraphics[width=0.12\linewidth]{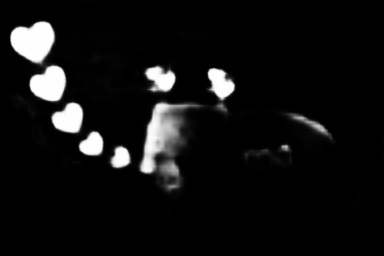} \\

\includegraphics[width=0.12\linewidth]{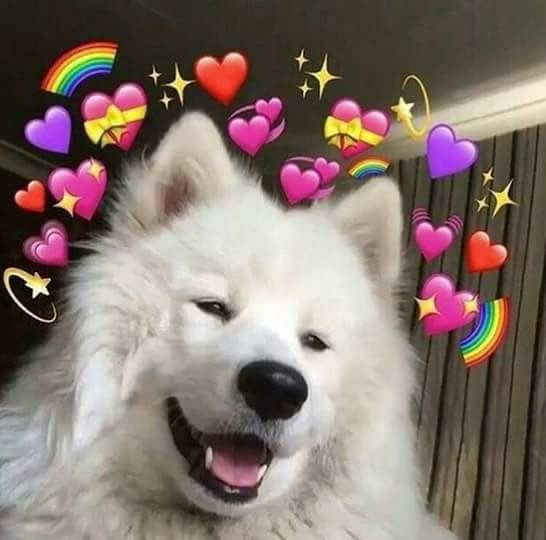} &
\includegraphics[width=0.12\linewidth]{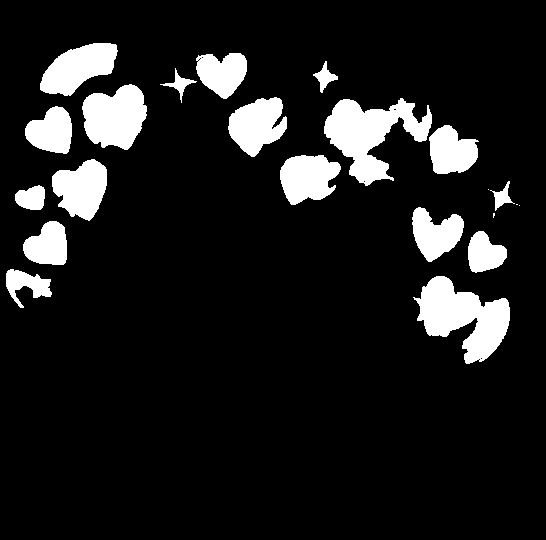} &
\includegraphics[width=0.12\linewidth]{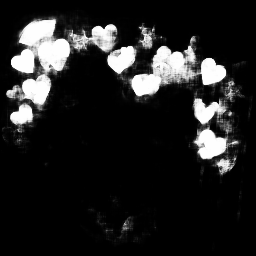} &
\includegraphics[width=0.12\linewidth]{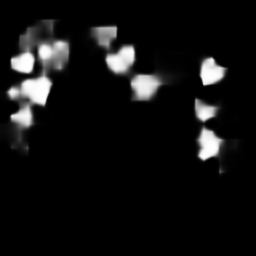} &
\includegraphics[width=0.12\linewidth]{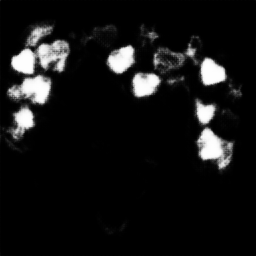} &
\includegraphics[width=0.12\linewidth]{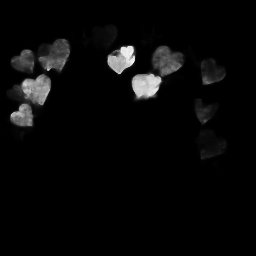} &
\includegraphics[width=0.12\linewidth]{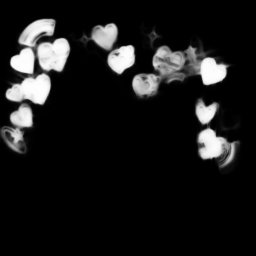} &
\includegraphics[width=0.12\linewidth]{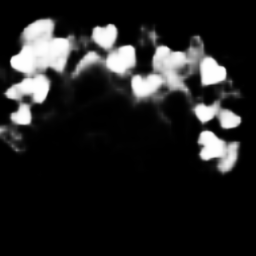} \\

\includegraphics[width=0.12\linewidth]{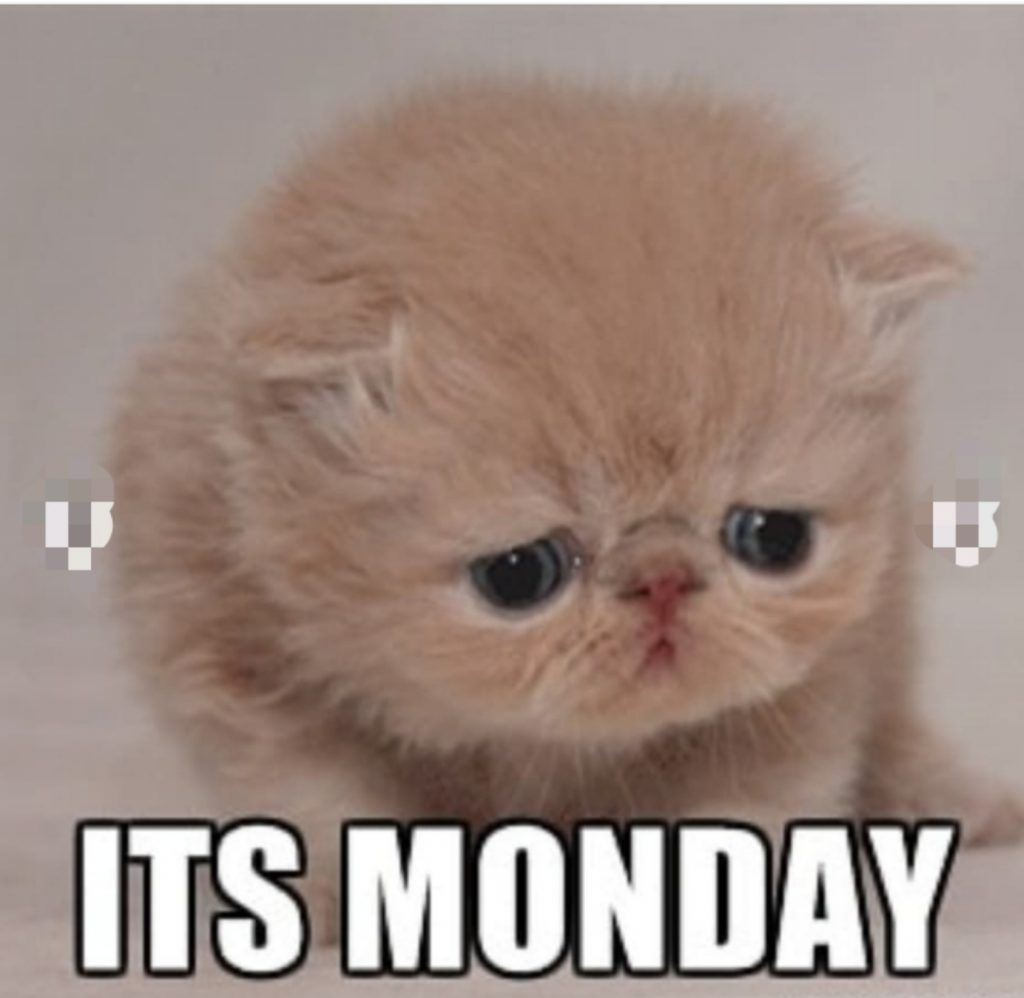} &
\includegraphics[width=0.12\linewidth]{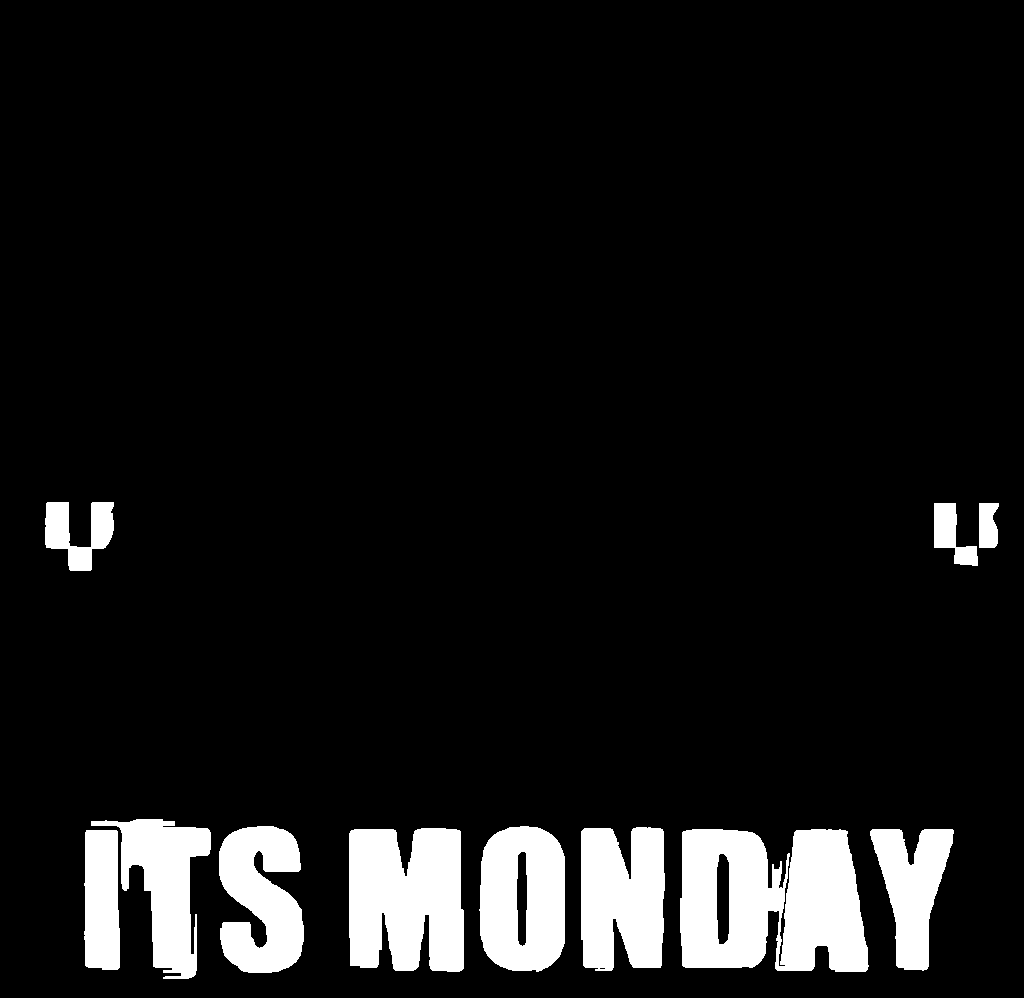} &
\includegraphics[width=0.12\linewidth]{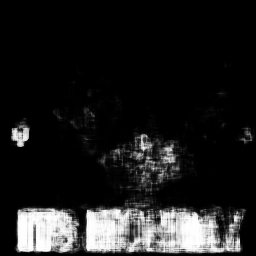} &
\includegraphics[width=0.12\linewidth]{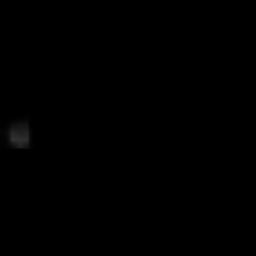} &
\includegraphics[width=0.12\linewidth]{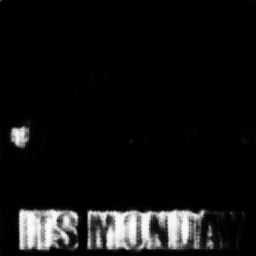} &
\includegraphics[width=0.12\linewidth]{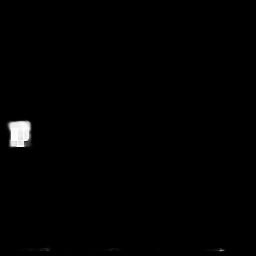} &
\includegraphics[width=0.12\linewidth]{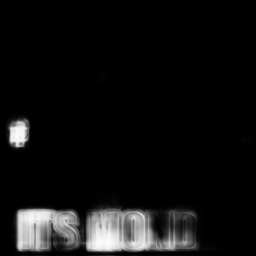} &
\includegraphics[width=0.12\linewidth]{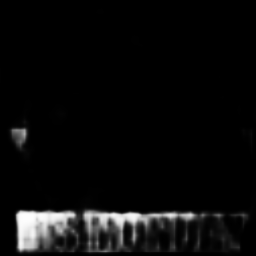} \\

& & & & & \\
(a) Image & (b) GT & (c) Ours & (d) DLV3 & (e) UNet & (f) BASNet & (g) PFANet & (h) SFPN \\
\end{tabular}
\vspace{2ex}
\caption{Visual comparison of the proposed method and the competing methods on wild data collected from popular social media websites}
\label{fig: wild test visualization}
\end{figure*}

We compare performance of our proposed model and competing models on 100 images collected from popular social media website using same metric as in the main text. Collected images are manually labeled. As shown in figure \ref{fig: wild test plots} (a, b) and table \ref{tab:wild test stats}, our proposed model displays best performance on these wild images. Some visual comparison examples are shown in figure \ref{fig: wild test visualization}. Note that the overall performance is poorer as compared with synthetic test set because some images are heavily corrupted by blur and noise.

\begin{table}[ht]
\centering
\begin{tabular}{ccccc}
\hline
\hline
 & \multicolumn{4}{c}{\textbf{Overall Performance on Test Set}}\\ 
 & mIoU & MAE & $F_{0.3}$ & $F_{2}$ \\ \hline
\textbf{Ours} & \textbf{0.631} & \textbf{0.044} & \textbf{0.776} & \textbf{0.793} \\ 
\textbf{Unet} & 0.445 & 0.071 & 0.610 & 0.685 \\
\textbf{DLV3} & 0.410 & 0.052 & 0.541 & 0.546 \\
\textbf{BASNet} & 0.410 & 0.063 & 0.569 & 0.541 \\
\textbf{PFANet} & 0.541 & \textbf{0.044} & 0.696 & 0.707 \\
\textbf{SFPN} & 0.540 & 0.050 & 0.683 & 0.700 \\
\hline\hline
\end{tabular}
\vspace{2ex}
\caption{Quantitative comparison of competing methods on data from wild}
\label{tab:wild test stats}
\end{table}

\section{Additional Visual Comparisons}
\label{sec: more visual results}
\begin{figure*}[bh]
\vspace{-2ex}
\setlength\tabcolsep{1pt}
\renewcommand{\arraystretch}{0.5}
\centering
\begin{tabular}{cccccccc}
\includegraphics[width=0.12\linewidth]{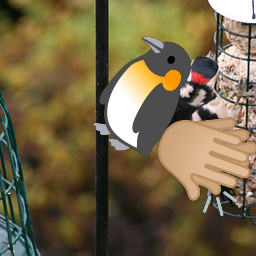} &
\includegraphics[width=0.12\linewidth]{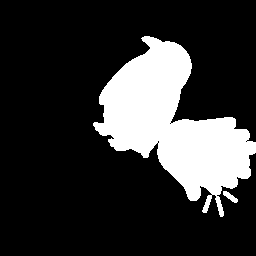} &
\includegraphics[width=0.12\linewidth]{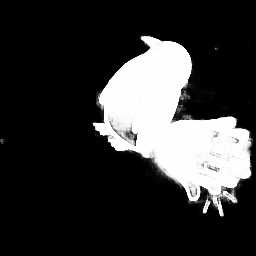} &
\includegraphics[width=0.12\linewidth]{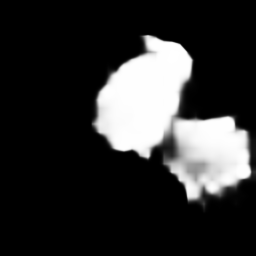} &
\includegraphics[width=0.12\linewidth]{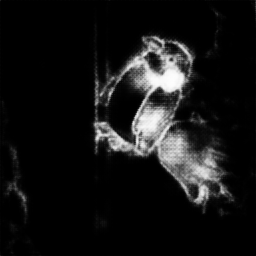} &
\includegraphics[width=0.12\linewidth]{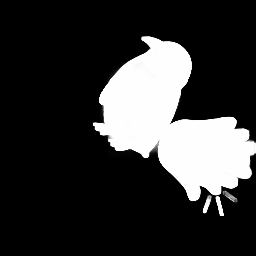} &
\includegraphics[width=0.12\linewidth]{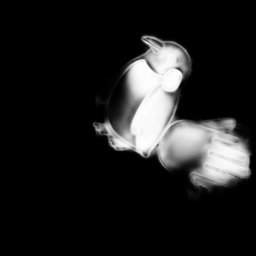} &
\includegraphics[width=0.12\linewidth]{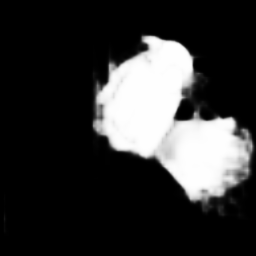} \\
\includegraphics[width=0.12\linewidth]{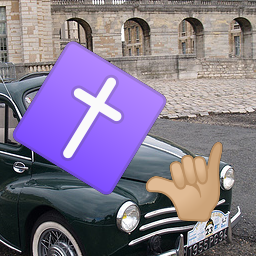} &
\includegraphics[width=0.12\linewidth]{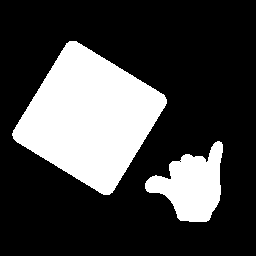} &
\includegraphics[width=0.12\linewidth]{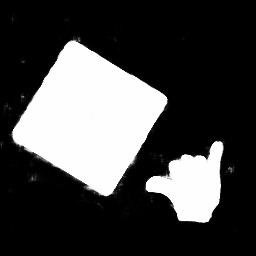} &
\includegraphics[width=0.12\linewidth]{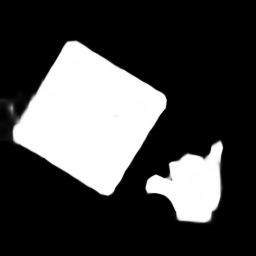} &
\includegraphics[width=0.12\linewidth]{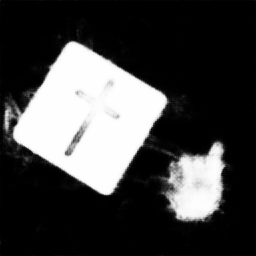} &
\includegraphics[width=0.12\linewidth]{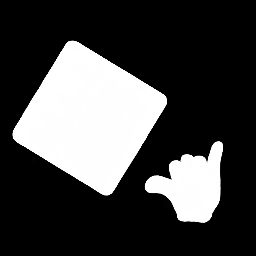} &
\includegraphics[width=0.12\linewidth]{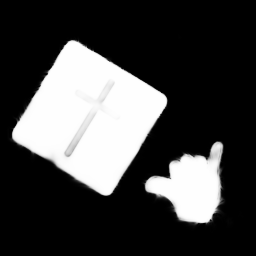} &
\includegraphics[width=0.12\linewidth]{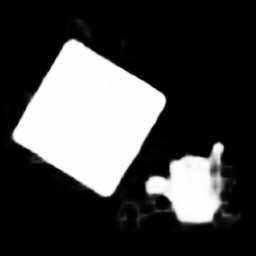} \\
\includegraphics[width=0.12\linewidth]{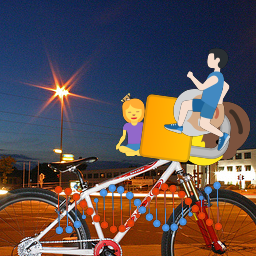} &
\includegraphics[width=0.12\linewidth]{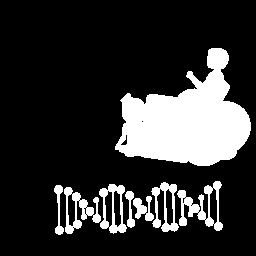} &
\includegraphics[width=0.12\linewidth]{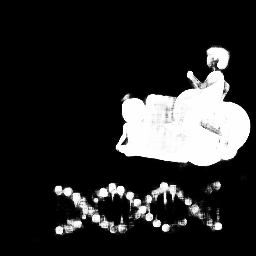} &
\includegraphics[width=0.12\linewidth]{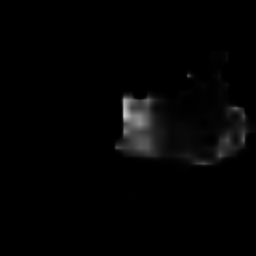} &
\includegraphics[width=0.12\linewidth]{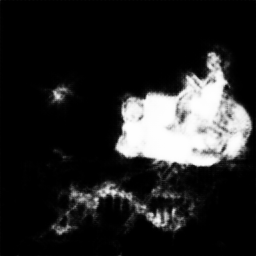} &
\includegraphics[width=0.12\linewidth]{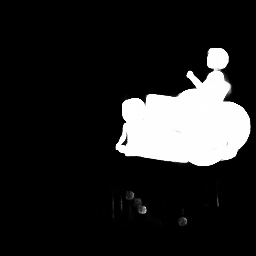} &
\includegraphics[width=0.12\linewidth]{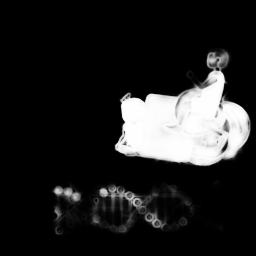} &
\includegraphics[width=0.12\linewidth]{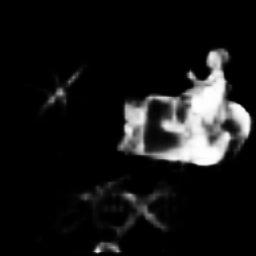} \\
\includegraphics[width=0.12\linewidth]{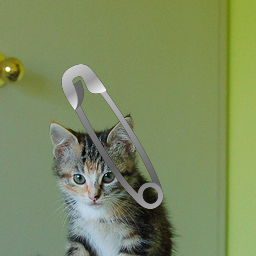} &
\includegraphics[width=0.12\linewidth]{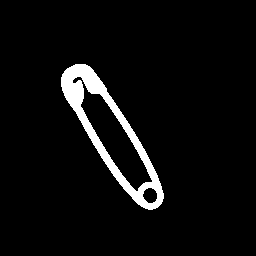} &
\includegraphics[width=0.12\linewidth]{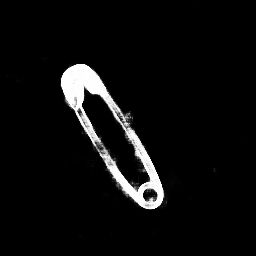} &
\includegraphics[width=0.12\linewidth]{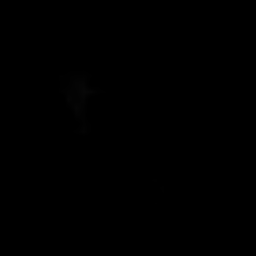} &
\includegraphics[width=0.12\linewidth]{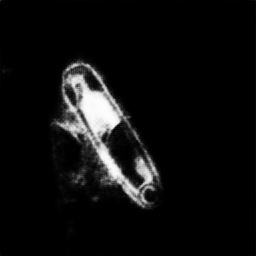} &
\includegraphics[width=0.12\linewidth]{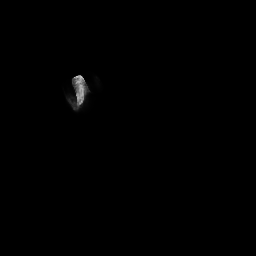} &
\includegraphics[width=0.12\linewidth]{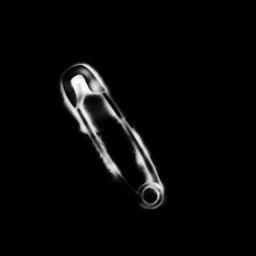} &
\includegraphics[width=0.12\linewidth]{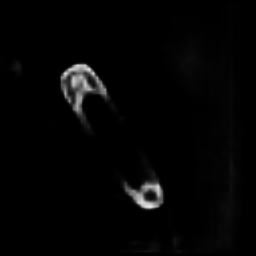} \\
\includegraphics[width=0.12\linewidth]{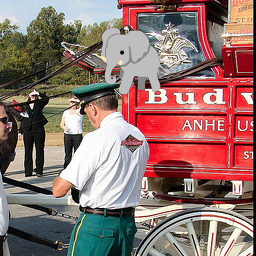} &
\includegraphics[width=0.12\linewidth]{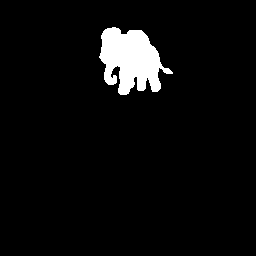} &
\includegraphics[width=0.12\linewidth]{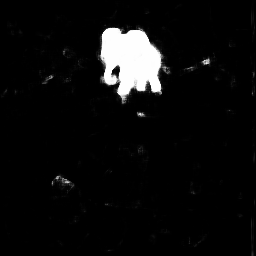} &
\includegraphics[width=0.12\linewidth]{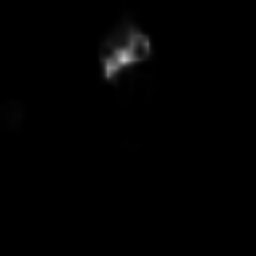} &
\includegraphics[width=0.12\linewidth]{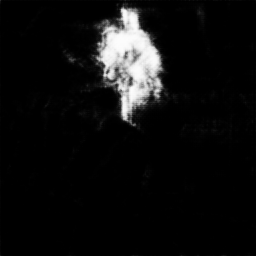} &
\includegraphics[width=0.12\linewidth]{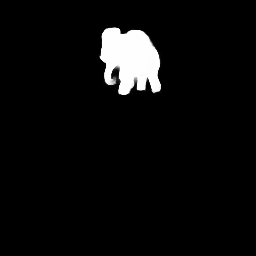} &
\includegraphics[width=0.12\linewidth]{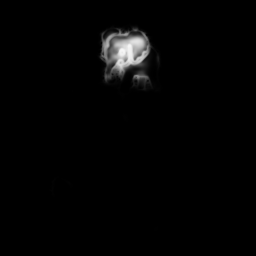} &
\includegraphics[width=0.12\linewidth]{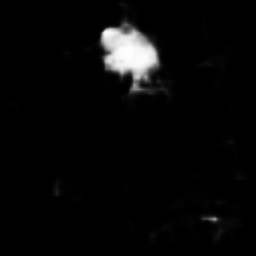} \\
\includegraphics[width=0.12\linewidth]{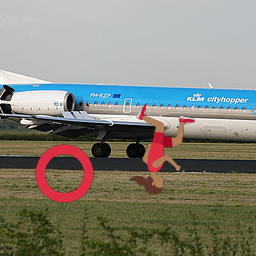} &
\includegraphics[width=0.12\linewidth]{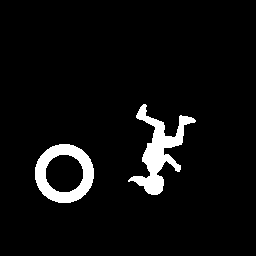} &
\includegraphics[width=0.12\linewidth]{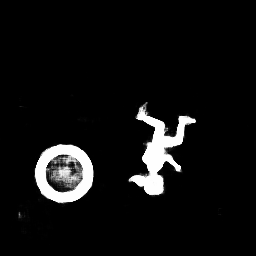} &
\includegraphics[width=0.12\linewidth]{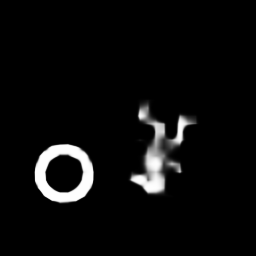} &
\includegraphics[width=0.12\linewidth]{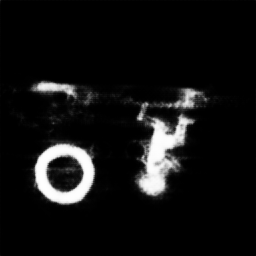} &
\includegraphics[width=0.12\linewidth]{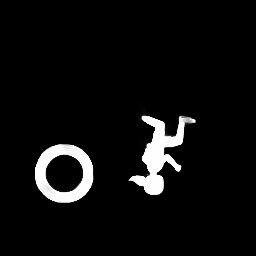} &
\includegraphics[width=0.12\linewidth]{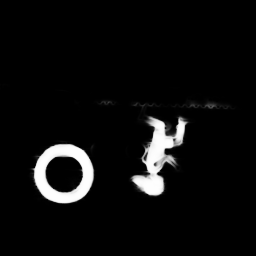} &
\includegraphics[width=0.12\linewidth]{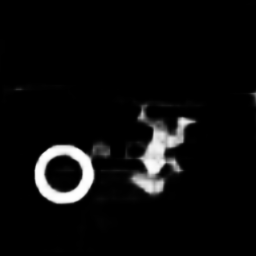} \\
\includegraphics[width=0.12\linewidth]{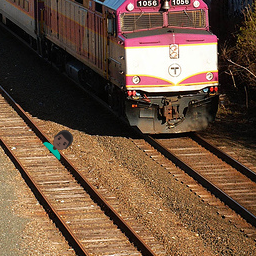} &
\includegraphics[width=0.12\linewidth]{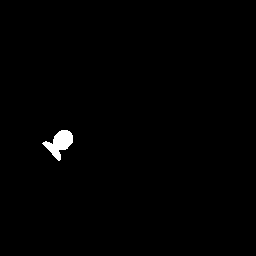} &
\includegraphics[width=0.12\linewidth]{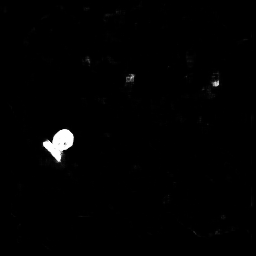} &
\includegraphics[width=0.12\linewidth]{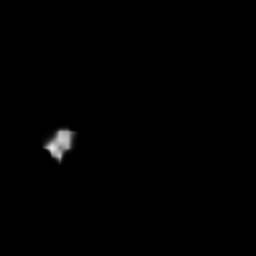} &
\includegraphics[width=0.12\linewidth]{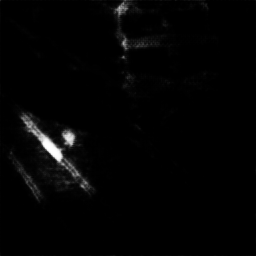} &
\includegraphics[width=0.12\linewidth]{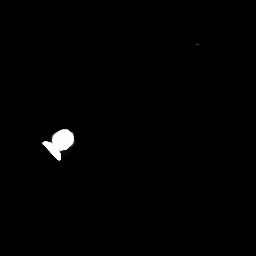} &
\includegraphics[width=0.12\linewidth]{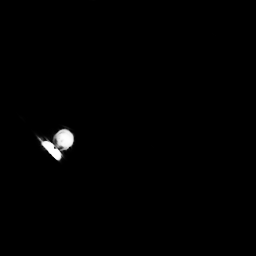} &
\includegraphics[width=0.12\linewidth]{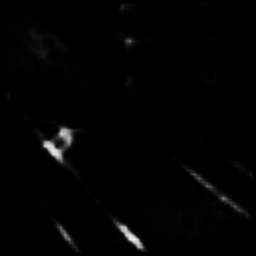} \\
\includegraphics[width=0.12\linewidth]{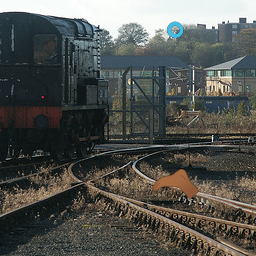} &
\includegraphics[width=0.12\linewidth]{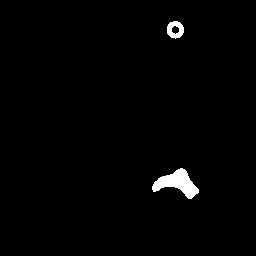} &
\includegraphics[width=0.12\linewidth]{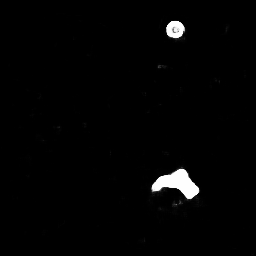} &
\includegraphics[width=0.12\linewidth]{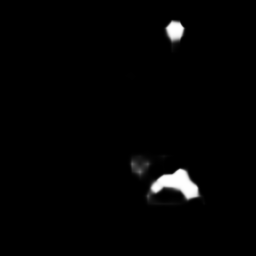} &
\includegraphics[width=0.12\linewidth]{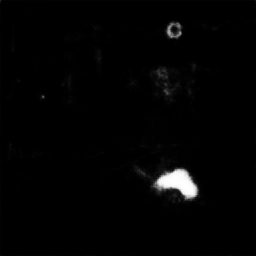} &
\includegraphics[width=0.12\linewidth]{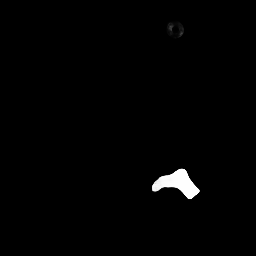} &
\includegraphics[width=0.12\linewidth]{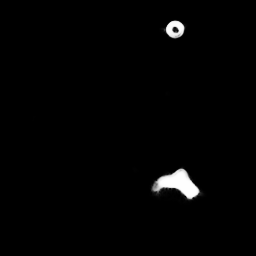} &
\includegraphics[width=0.12\linewidth]{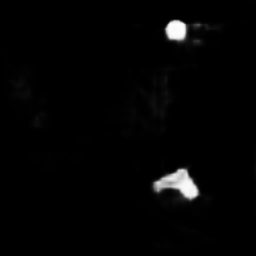} \\
\includegraphics[width=0.12\linewidth]{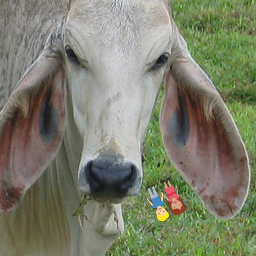} &
\includegraphics[width=0.12\linewidth]{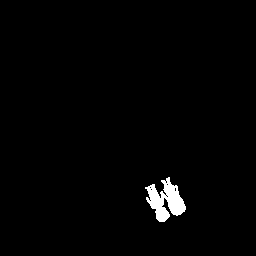} &
\includegraphics[width=0.12\linewidth]{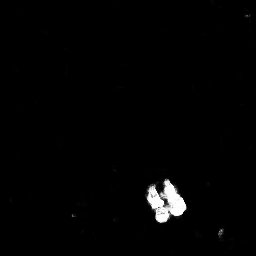} &
\includegraphics[width=0.12\linewidth]{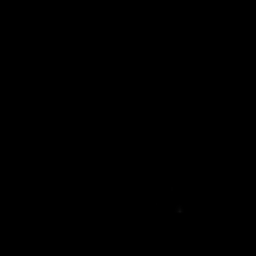} &
\includegraphics[width=0.12\linewidth]{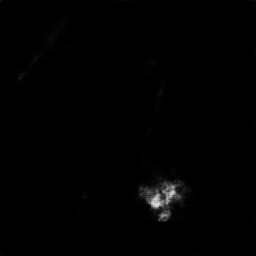} &
\includegraphics[width=0.12\linewidth]{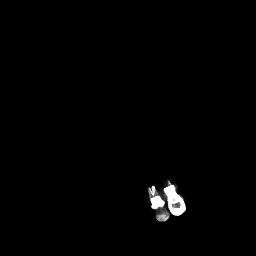} &
\includegraphics[width=0.12\linewidth]{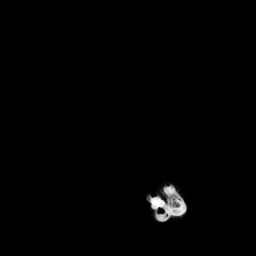} &
\includegraphics[width=0.12\linewidth]{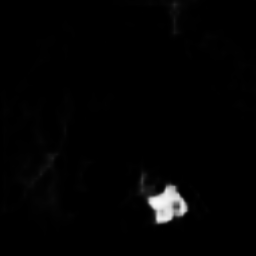} \\

& & & & & \\
(a) Image & (b) GT & (c) Ours & (d) DLV3 & (e) UNet & (f) BASNet & (g) PFANet & (h) SFPN \\
\end{tabular}
\vspace{2ex}
\caption{Visual comparison of the proposed method and the competing methods on stickers}
\label{fig: test set visualization stickers}
\end{figure*}

\begin{figure*}[bh]
\setlength\tabcolsep{1pt}
\renewcommand{\arraystretch}{0.5}
\centering
\begin{tabular}{cccccccc}
\includegraphics[width=0.12\linewidth]{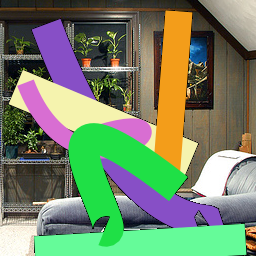} &
\includegraphics[width=0.12\linewidth]{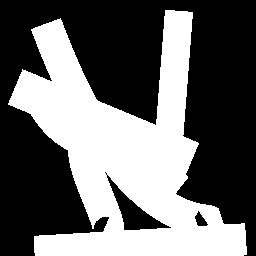} &
\includegraphics[width=0.12\linewidth]{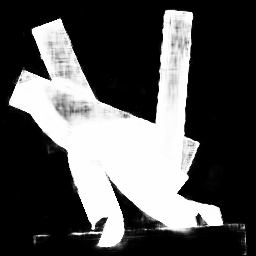} &
\includegraphics[width=0.12\linewidth]{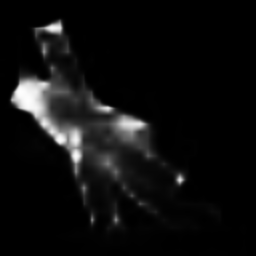} &
\includegraphics[width=0.12\linewidth]{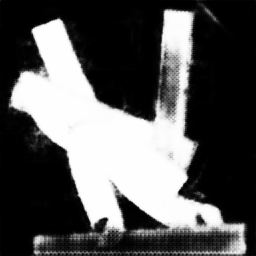} &
\includegraphics[width=0.12\linewidth]{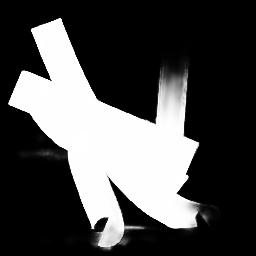} &
\includegraphics[width=0.12\linewidth]{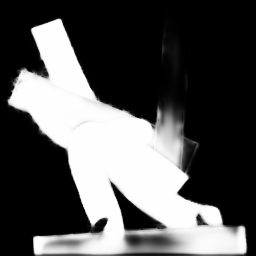} &
\includegraphics[width=0.12\linewidth]{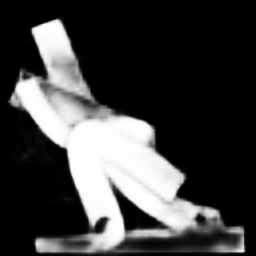} \\
\includegraphics[width=0.12\linewidth]{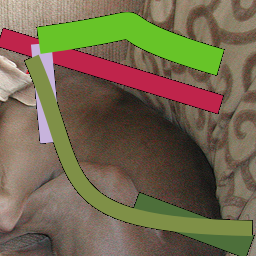} &
\includegraphics[width=0.12\linewidth]{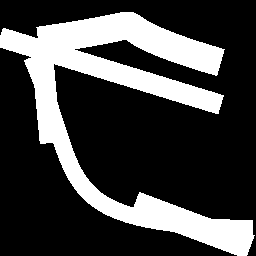} &
\includegraphics[width=0.12\linewidth]{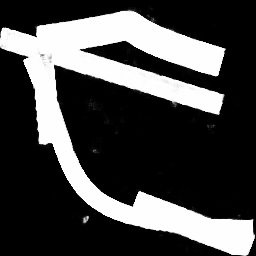} &
\includegraphics[width=0.12\linewidth]{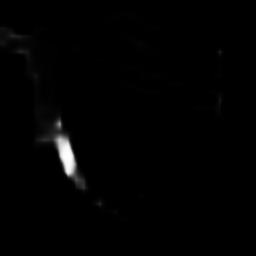} &
\includegraphics[width=0.12\linewidth]{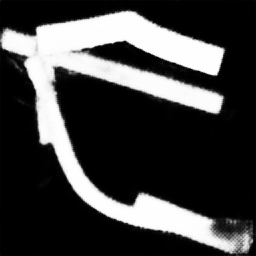} &
\includegraphics[width=0.12\linewidth]{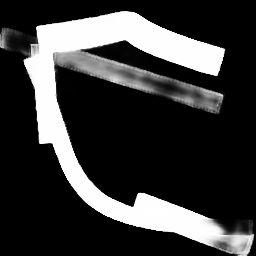} &
\includegraphics[width=0.12\linewidth]{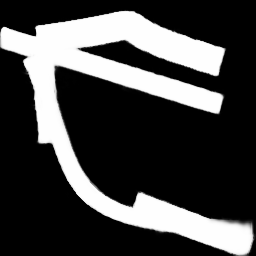} &
\includegraphics[width=0.12\linewidth]{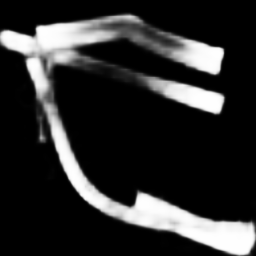} \\
\includegraphics[width=0.12\linewidth]{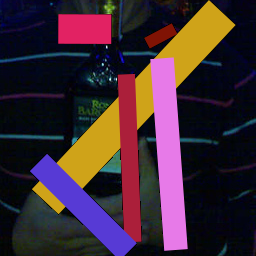} &
\includegraphics[width=0.12\linewidth]{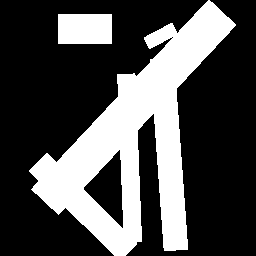} &
\includegraphics[width=0.12\linewidth]{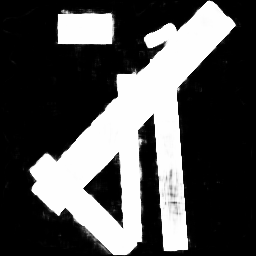} &
\includegraphics[width=0.12\linewidth]{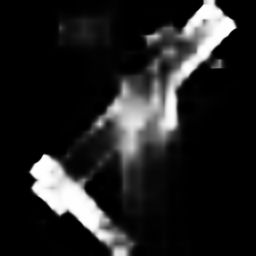} &
\includegraphics[width=0.12\linewidth]{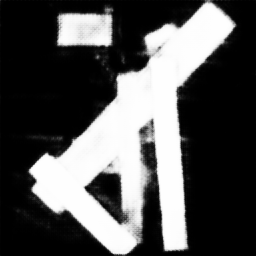} &
\includegraphics[width=0.12\linewidth]{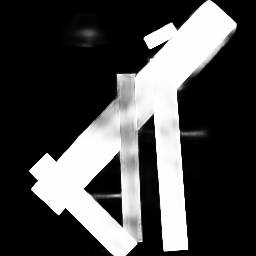} &
\includegraphics[width=0.12\linewidth]{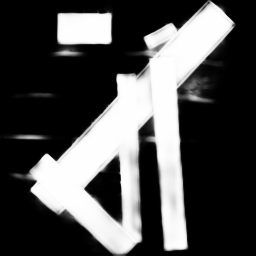} &
\includegraphics[width=0.12\linewidth]{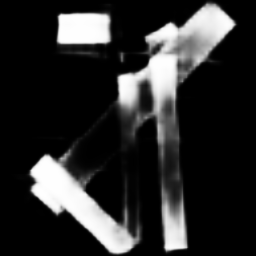} \\
\includegraphics[width=0.12\linewidth]{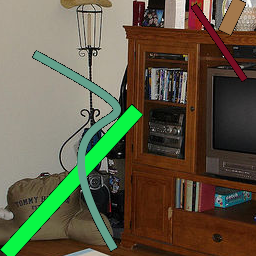} &
\includegraphics[width=0.12\linewidth]{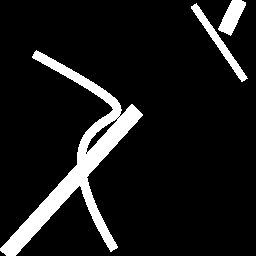} &
\includegraphics[width=0.12\linewidth]{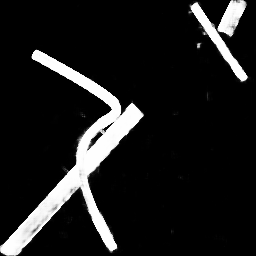} &
\includegraphics[width=0.12\linewidth]{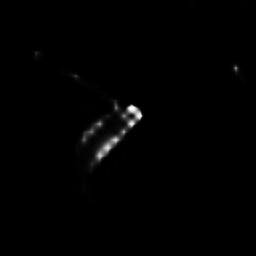} &
\includegraphics[width=0.12\linewidth]{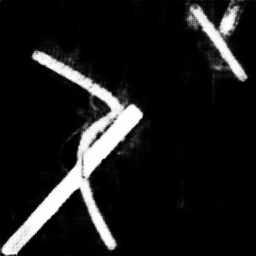} &
\includegraphics[width=0.12\linewidth]{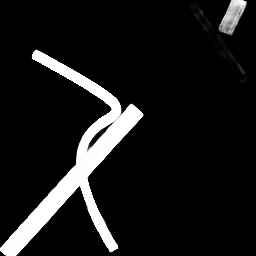} &
\includegraphics[width=0.12\linewidth]{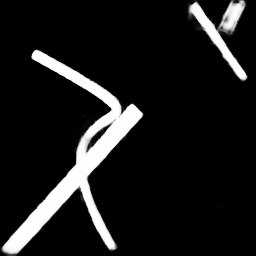} &
\includegraphics[width=0.12\linewidth]{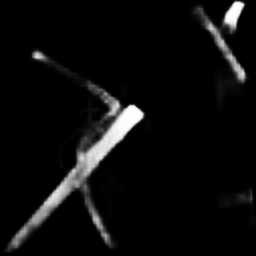} \\
\includegraphics[width=0.12\linewidth]{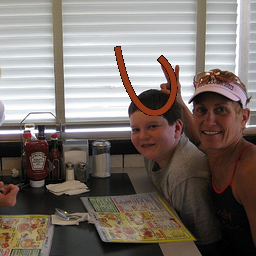} &
\includegraphics[width=0.12\linewidth]{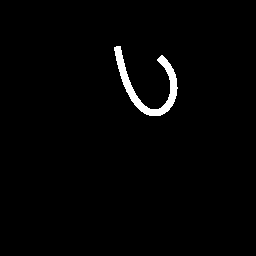} &
\includegraphics[width=0.12\linewidth]{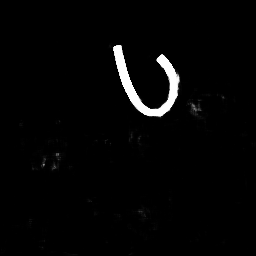} &
\includegraphics[width=0.12\linewidth]{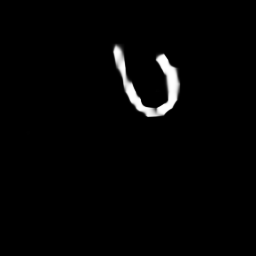} &
\includegraphics[width=0.12\linewidth]{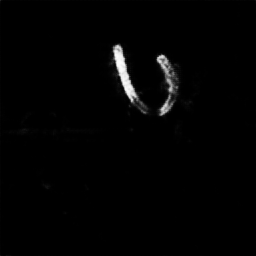} &
\includegraphics[width=0.12\linewidth]{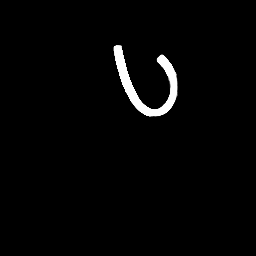} &
\includegraphics[width=0.12\linewidth]{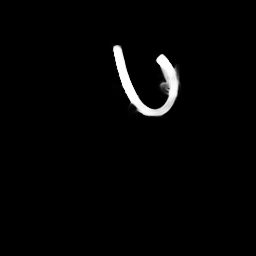} &
\includegraphics[width=0.12\linewidth]{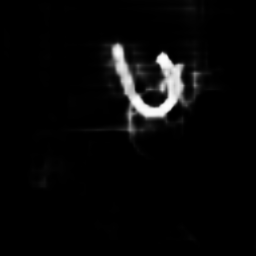} \\
\includegraphics[width=0.12\linewidth]{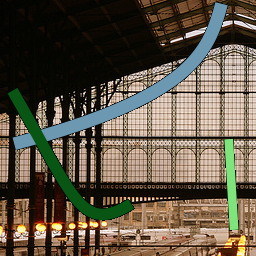} &
\includegraphics[width=0.12\linewidth]{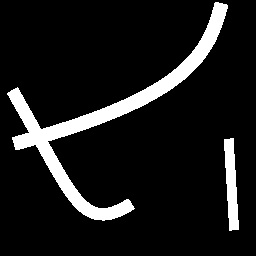} &
\includegraphics[width=0.12\linewidth]{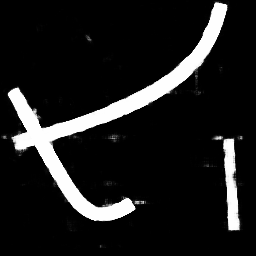} &
\includegraphics[width=0.12\linewidth]{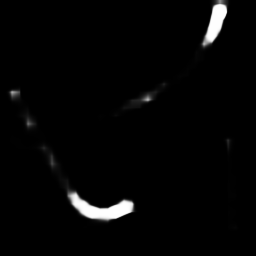} &
\includegraphics[width=0.12\linewidth]{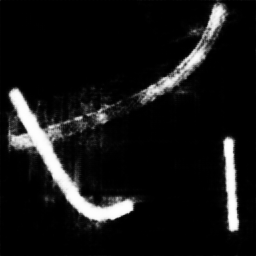} &
\includegraphics[width=0.12\linewidth]{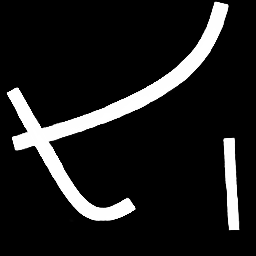} &
\includegraphics[width=0.12\linewidth]{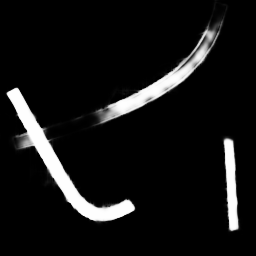} &
\includegraphics[width=0.12\linewidth]{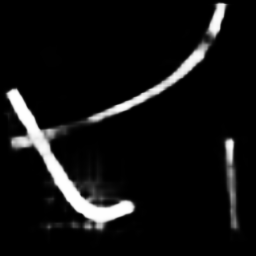} \\
\includegraphics[width=0.12\linewidth]{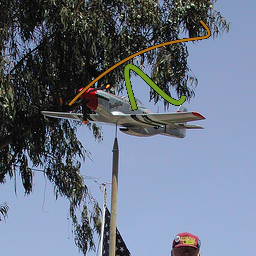} &
\includegraphics[width=0.12\linewidth]{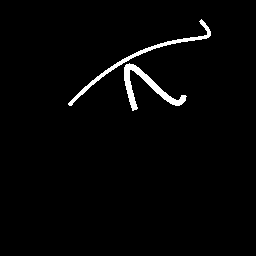} &
\includegraphics[width=0.12\linewidth]{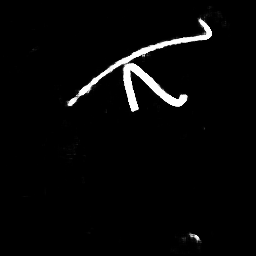} &
\includegraphics[width=0.12\linewidth]{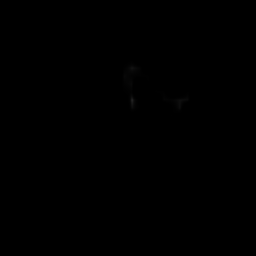} &
\includegraphics[width=0.12\linewidth]{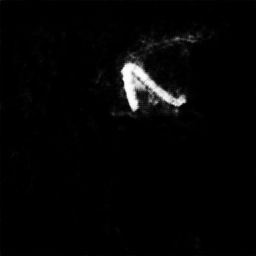} &
\includegraphics[width=0.12\linewidth]{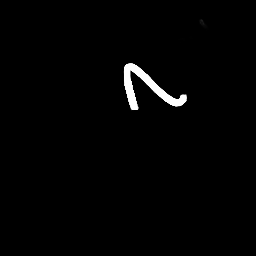} &
\includegraphics[width=0.12\linewidth]{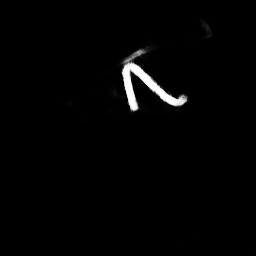} &
\includegraphics[width=0.12\linewidth]{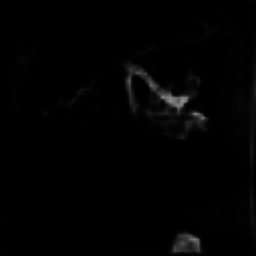} \\
\includegraphics[width=0.12\linewidth]{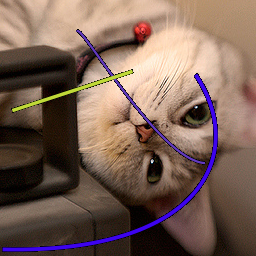} &
\includegraphics[width=0.12\linewidth]{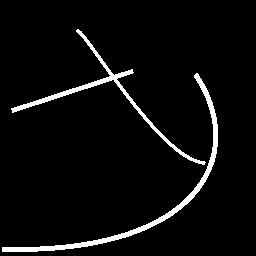} &
\includegraphics[width=0.12\linewidth]{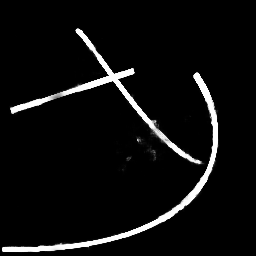} &
\includegraphics[width=0.12\linewidth]{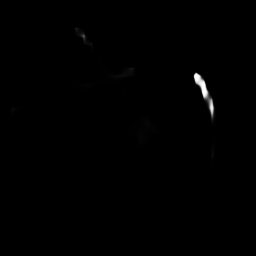} &
\includegraphics[width=0.12\linewidth]{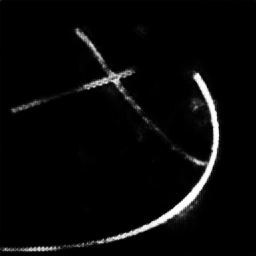} &
\includegraphics[width=0.12\linewidth]{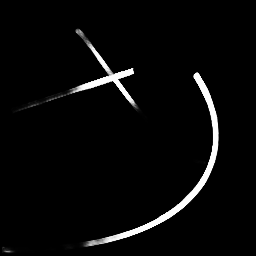} &
\includegraphics[width=0.12\linewidth]{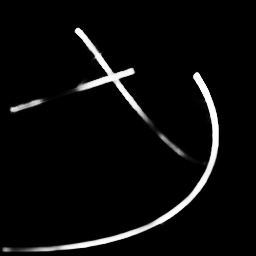} &
\includegraphics[width=0.12\linewidth]{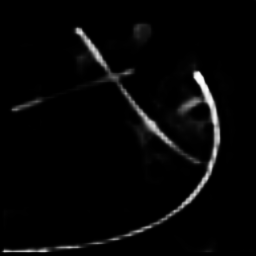} \\
\includegraphics[width=0.12\linewidth]{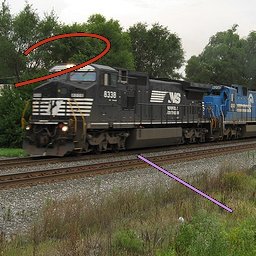} &
\includegraphics[width=0.12\linewidth]{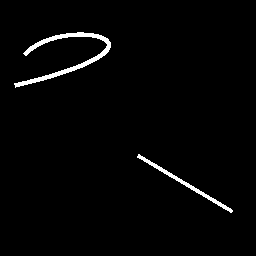} &
\includegraphics[width=0.12\linewidth]{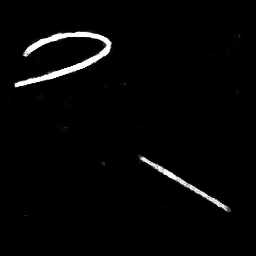} &
\includegraphics[width=0.12\linewidth]{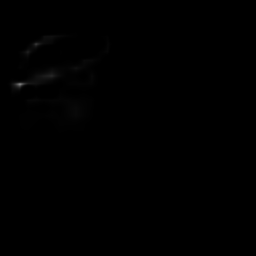} &
\includegraphics[width=0.12\linewidth]{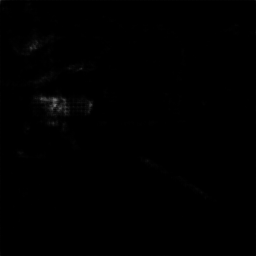} &
\includegraphics[width=0.12\linewidth]{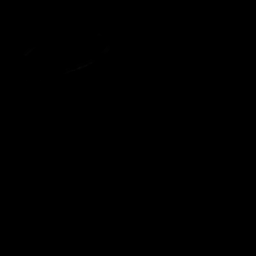} &
\includegraphics[width=0.12\linewidth]{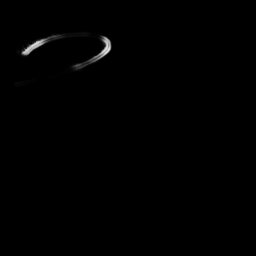} &
\includegraphics[width=0.12\linewidth]{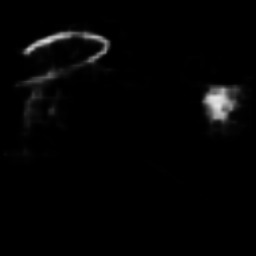} \\

& & & & & \\
(a) Image & (b) GT & (c) Ours & (d) DLV3 & (e) UNet & (f) BASNet & (g) PFANet & (h) SFPN \\
\end{tabular}
\vspace{2ex}
\caption{Visual comparison of the proposed method and the competing methods on lines}
\label{fig: test set visualization 1}
\end{figure*}

\begin{figure*}[bh]
\setlength\tabcolsep{1pt}
\renewcommand{\arraystretch}{0.5}
\centering
\begin{tabular}{cccccccc}
\includegraphics[width=0.12\linewidth]{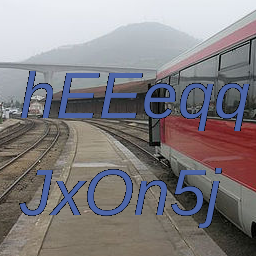} &
\includegraphics[width=0.12\linewidth]{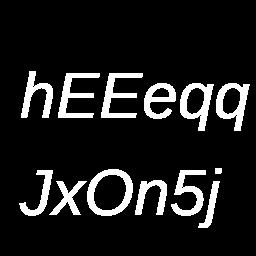} &
\includegraphics[width=0.12\linewidth]{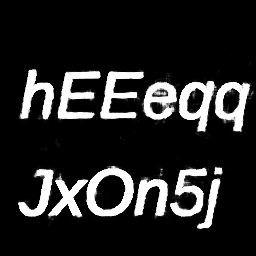} &
\includegraphics[width=0.12\linewidth]{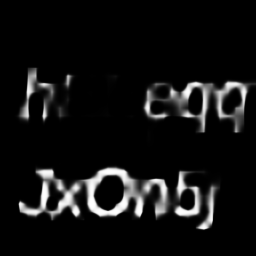} &
\includegraphics[width=0.12\linewidth]{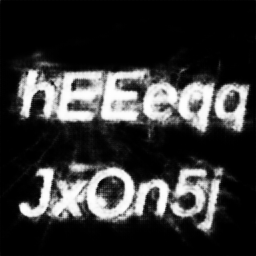} &
\includegraphics[width=0.12\linewidth]{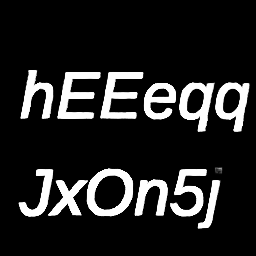} &
\includegraphics[width=0.12\linewidth]{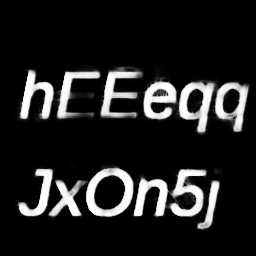} &
\includegraphics[width=0.12\linewidth]{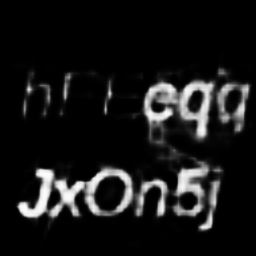} \\
\includegraphics[width=0.12\linewidth]{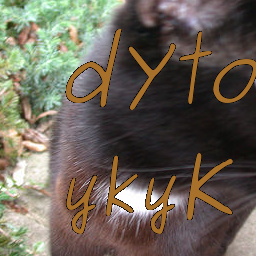} &
\includegraphics[width=0.12\linewidth]{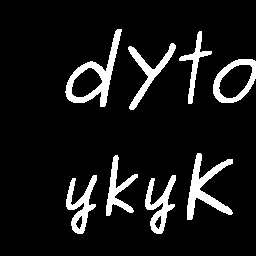} &
\includegraphics[width=0.12\linewidth]{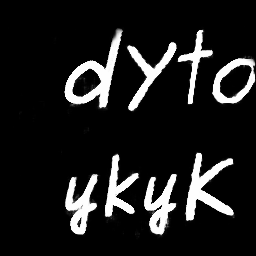} &
\includegraphics[width=0.12\linewidth]{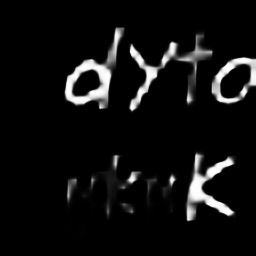} &
\includegraphics[width=0.12\linewidth]{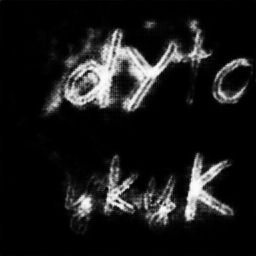} &
\includegraphics[width=0.12\linewidth]{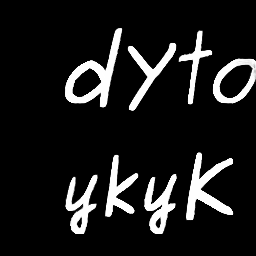} &
\includegraphics[width=0.12\linewidth]{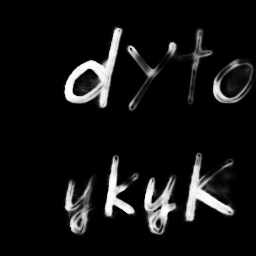} &
\includegraphics[width=0.12\linewidth]{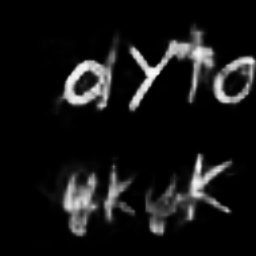} \\
\includegraphics[width=0.12\linewidth]{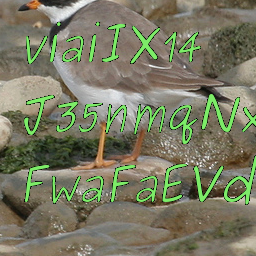} &
\includegraphics[width=0.12\linewidth]{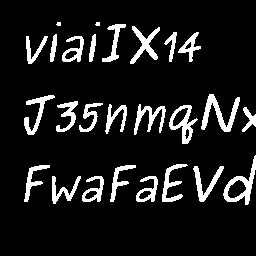} &
\includegraphics[width=0.12\linewidth]{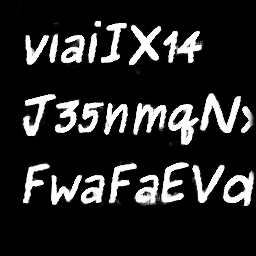} &
\includegraphics[width=0.12\linewidth]{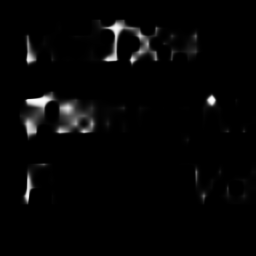} &
\includegraphics[width=0.12\linewidth]{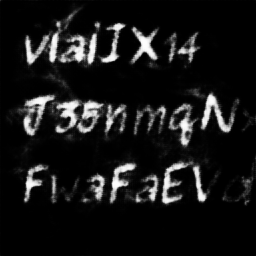} &
\includegraphics[width=0.12\linewidth]{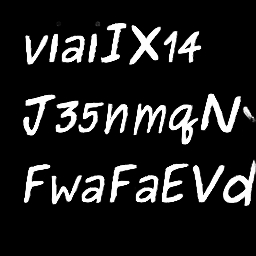} &
\includegraphics[width=0.12\linewidth]{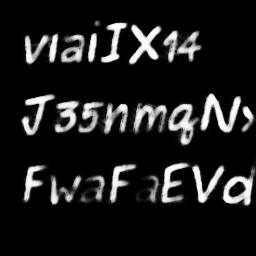} &
\includegraphics[width=0.12\linewidth]{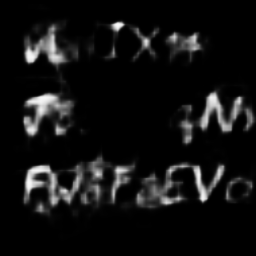} \\
\includegraphics[width=0.12\linewidth]{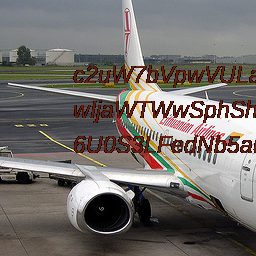} &
\includegraphics[width=0.12\linewidth]{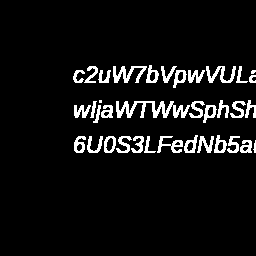} &
\includegraphics[width=0.12\linewidth]{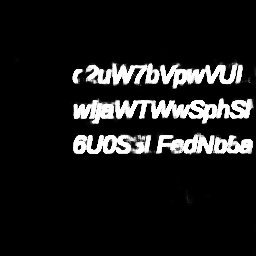} &
\includegraphics[width=0.12\linewidth]{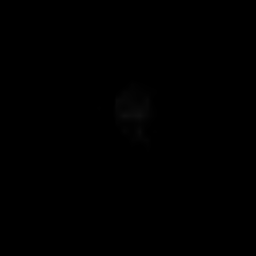} &
\includegraphics[width=0.12\linewidth]{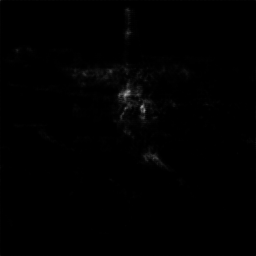} &
\includegraphics[width=0.12\linewidth]{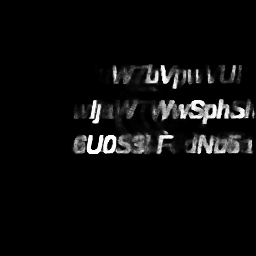} &
\includegraphics[width=0.12\linewidth]{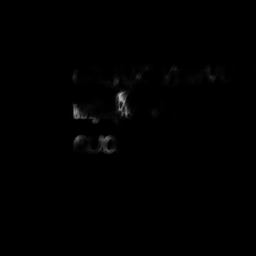} &
\includegraphics[width=0.12\linewidth]{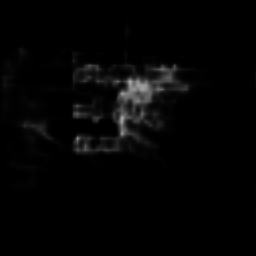} \\
\includegraphics[width=0.12\linewidth]{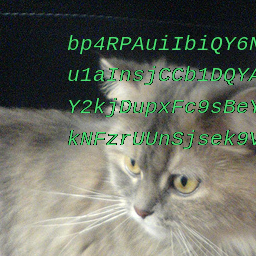} &
\includegraphics[width=0.12\linewidth]{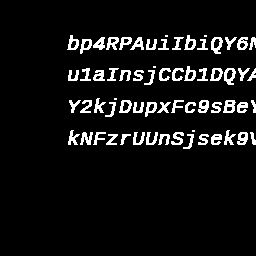} &
\includegraphics[width=0.12\linewidth]{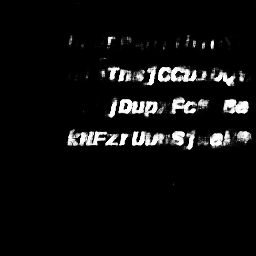} &
\includegraphics[width=0.12\linewidth]{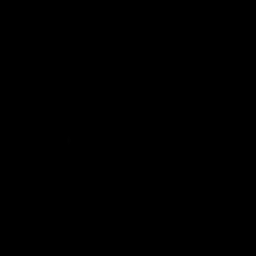} &
\includegraphics[width=0.12\linewidth]{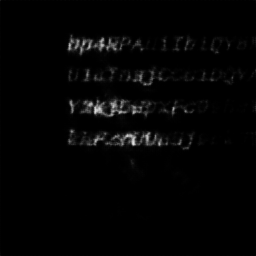} &
\includegraphics[width=0.12\linewidth]{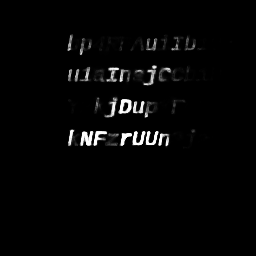} &
\includegraphics[width=0.12\linewidth]{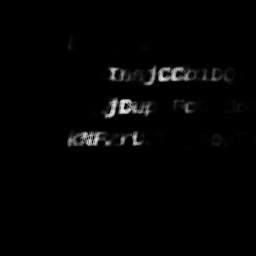} &
\includegraphics[width=0.12\linewidth]{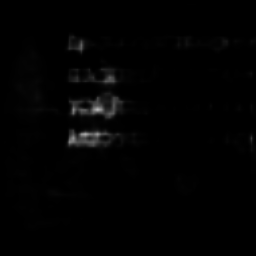} \\
\includegraphics[width=0.12\linewidth]{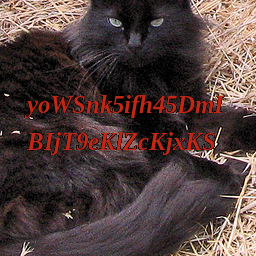} &
\includegraphics[width=0.12\linewidth]{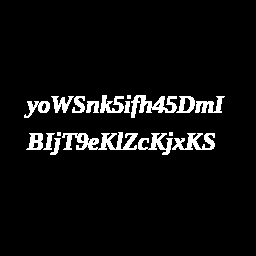} &
\includegraphics[width=0.12\linewidth]{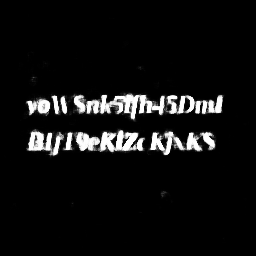} &
\includegraphics[width=0.12\linewidth]{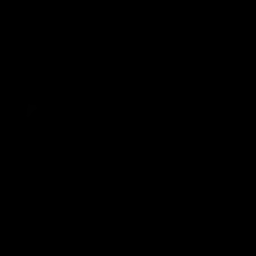} &
\includegraphics[width=0.12\linewidth]{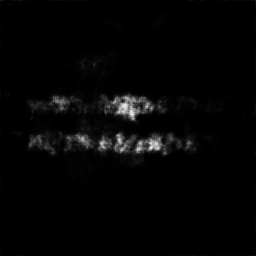} &
\includegraphics[width=0.12\linewidth]{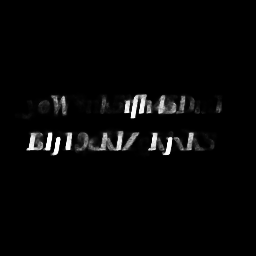} &
\includegraphics[width=0.12\linewidth]{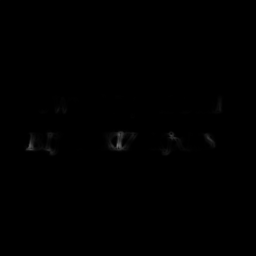} &
\includegraphics[width=0.12\linewidth]{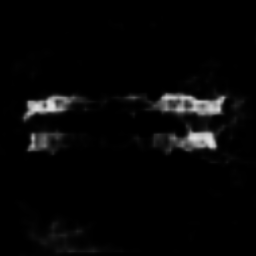} \\
\includegraphics[width=0.12\linewidth]{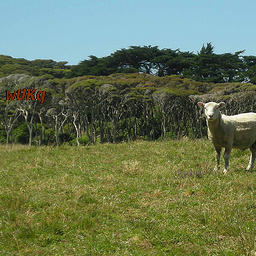} &
\includegraphics[width=0.12\linewidth]{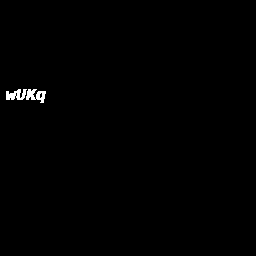} &
\includegraphics[width=0.12\linewidth]{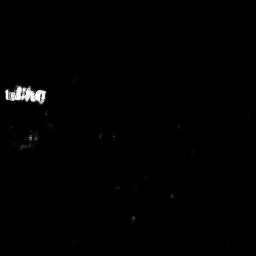} &
\includegraphics[width=0.12\linewidth]{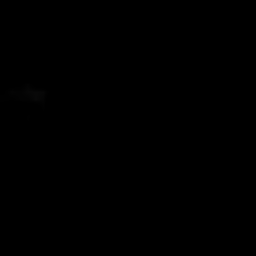} &
\includegraphics[width=0.12\linewidth]{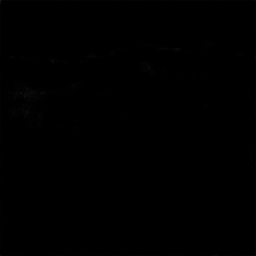} &
\includegraphics[width=0.12\linewidth]{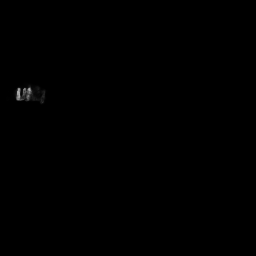} &
\includegraphics[width=0.12\linewidth]{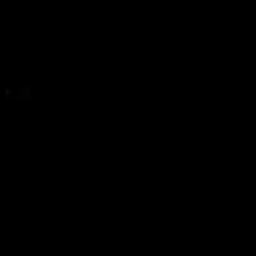} &
\includegraphics[width=0.12\linewidth]{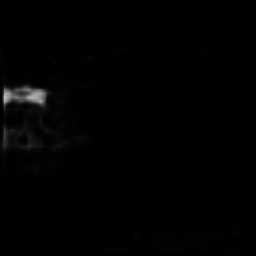} \\
\includegraphics[width=0.12\linewidth]{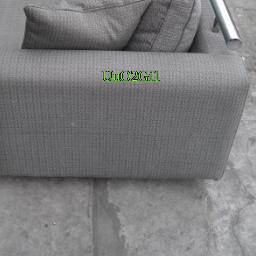} &
\includegraphics[width=0.12\linewidth]{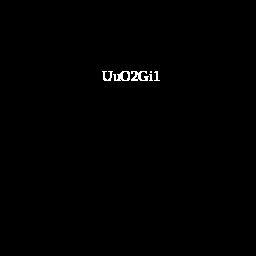} &
\includegraphics[width=0.12\linewidth]{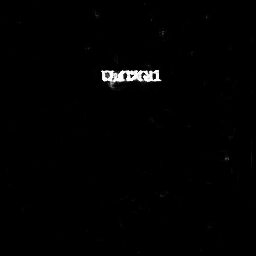} &
\includegraphics[width=0.12\linewidth]{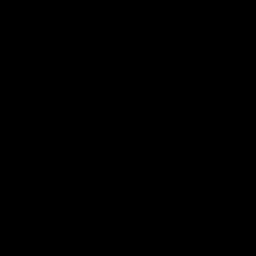} &
\includegraphics[width=0.12\linewidth]{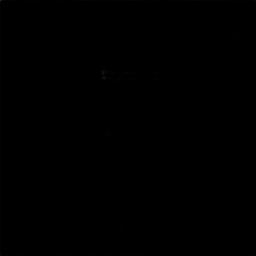} &
\includegraphics[width=0.12\linewidth]{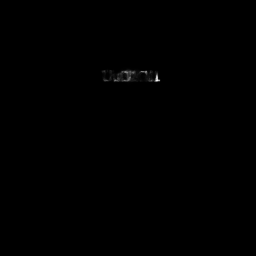} &
\includegraphics[width=0.12\linewidth]{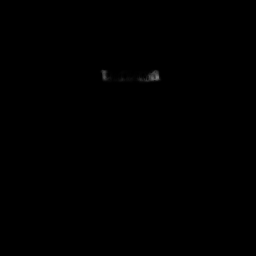} &
\includegraphics[width=0.12\linewidth]{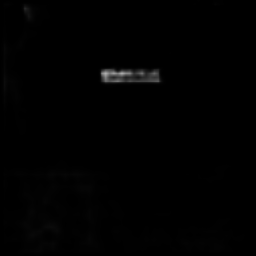} \\
\includegraphics[width=0.12\linewidth]{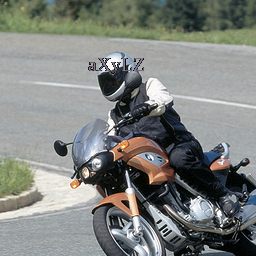} &
\includegraphics[width=0.12\linewidth]{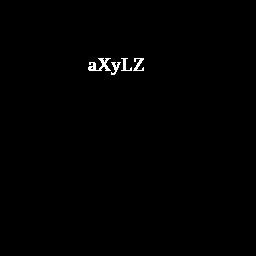} &
\includegraphics[width=0.12\linewidth]{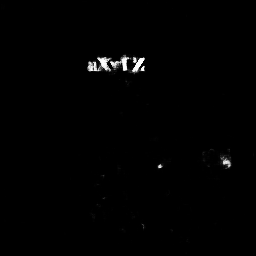} &
\includegraphics[width=0.12\linewidth]{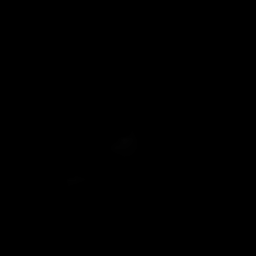} &
\includegraphics[width=0.12\linewidth]{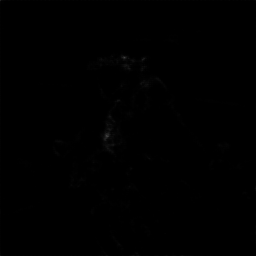} &
\includegraphics[width=0.12\linewidth]{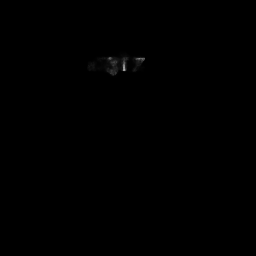} &
\includegraphics[width=0.12\linewidth]{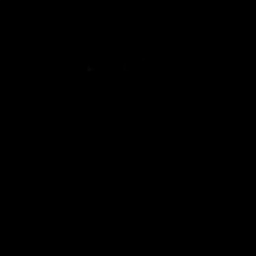} &
\includegraphics[width=0.12\linewidth]{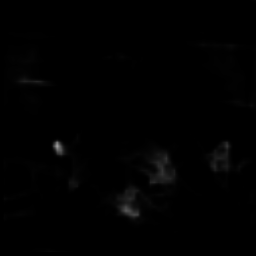} \\

& & & & & \\
(a) Image & (b) GT & (c) Ours & (d) DLV3 & (e) UNet & (f) BASNet & (g) PFANet & (h) SFPN \\
\end{tabular}
\vspace{2ex}
\caption{Visual comparison of the proposed method and the competing methods on text}
\label{fig: test set visualization text}
\end{figure*}

\begin{figure*}[bh]
\setlength\tabcolsep{1pt}
\renewcommand{\arraystretch}{0.5}
\centering
\begin{tabular}{cccccccc}
\includegraphics[width=0.12\linewidth]{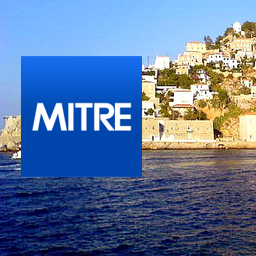} &
\includegraphics[width=0.12\linewidth]{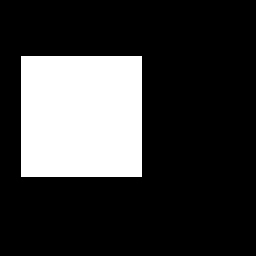} &
\includegraphics[width=0.12\linewidth]{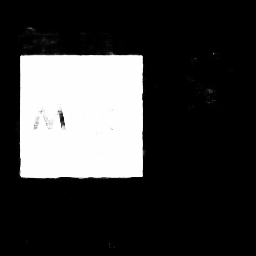} &
\includegraphics[width=0.12\linewidth]{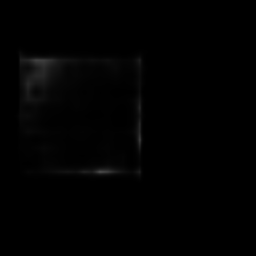} &
\includegraphics[width=0.12\linewidth]{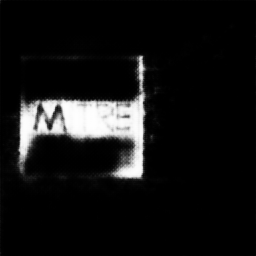} &
\includegraphics[width=0.12\linewidth]{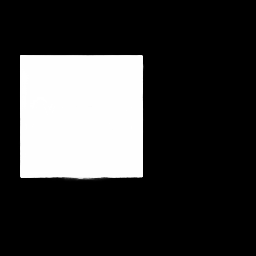} &
\includegraphics[width=0.12\linewidth]{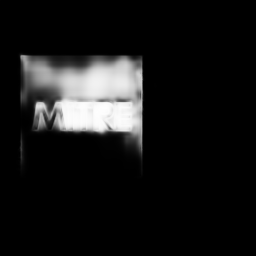} &
\includegraphics[width=0.12\linewidth]{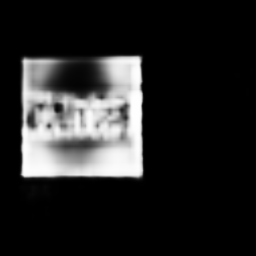} \\
\includegraphics[width=0.12\linewidth]{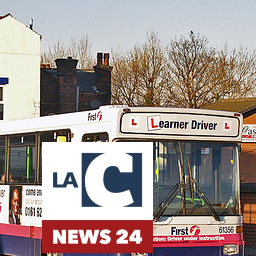} &
\includegraphics[width=0.12\linewidth]{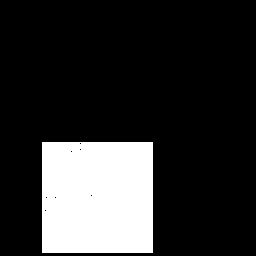} &
\includegraphics[width=0.12\linewidth]{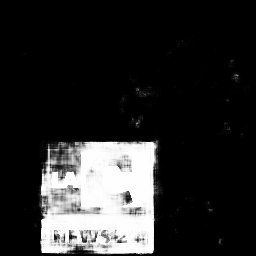} &
\includegraphics[width=0.12\linewidth]{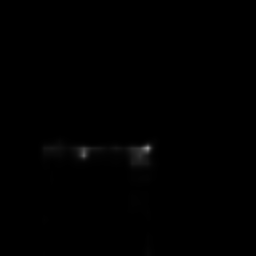} &
\includegraphics[width=0.12\linewidth]{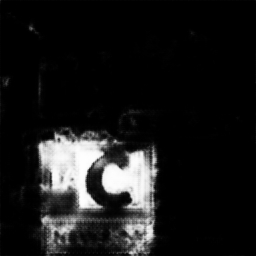} &
\includegraphics[width=0.12\linewidth]{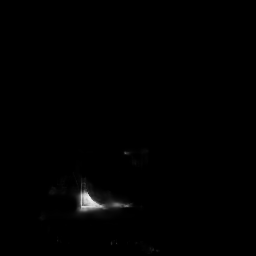} &
\includegraphics[width=0.12\linewidth]{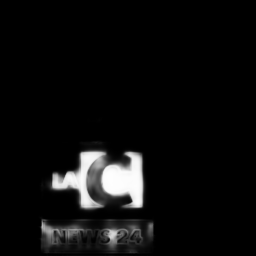} &
\includegraphics[width=0.12\linewidth]{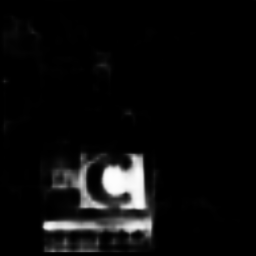} \\
\includegraphics[width=0.12\linewidth]{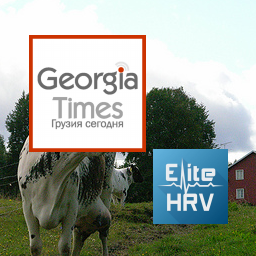} &
\includegraphics[width=0.12\linewidth]{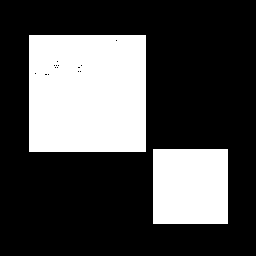} &
\includegraphics[width=0.12\linewidth]{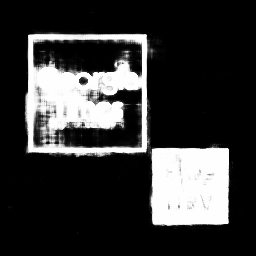} &
\includegraphics[width=0.12\linewidth]{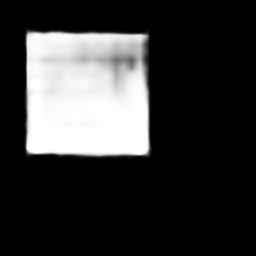} &
\includegraphics[width=0.12\linewidth]{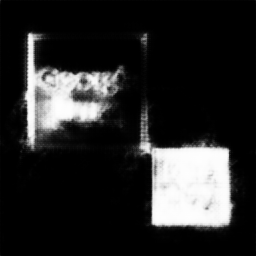} &
\includegraphics[width=0.12\linewidth]{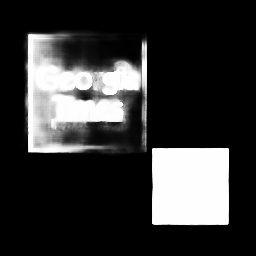} &
\includegraphics[width=0.12\linewidth]{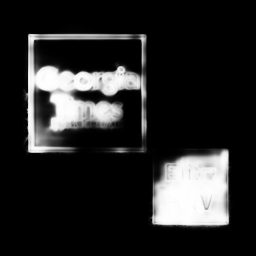} &
\includegraphics[width=0.12\linewidth]{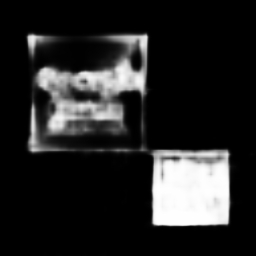} \\
\includegraphics[width=0.12\linewidth]{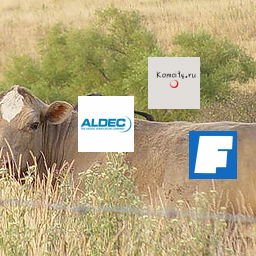} &
\includegraphics[width=0.12\linewidth]{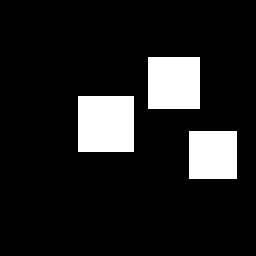} &
\includegraphics[width=0.12\linewidth]{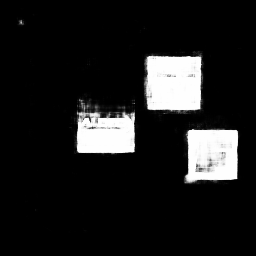} &
\includegraphics[width=0.12\linewidth]{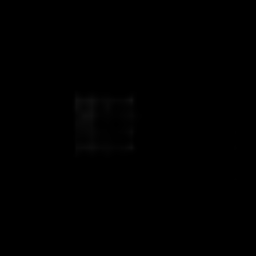} &
\includegraphics[width=0.12\linewidth]{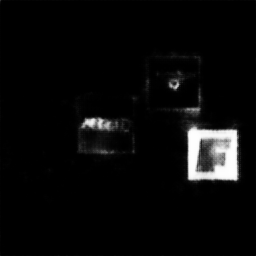} &
\includegraphics[width=0.12\linewidth]{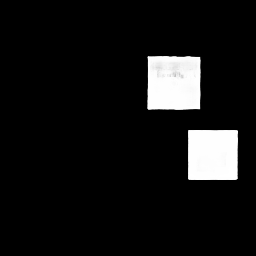} &
\includegraphics[width=0.12\linewidth]{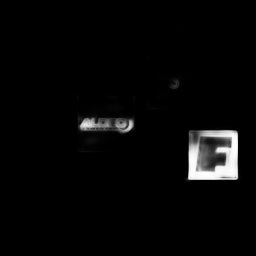} &
\includegraphics[width=0.12\linewidth]{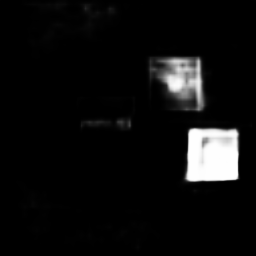} \\
\includegraphics[width=0.12\linewidth]{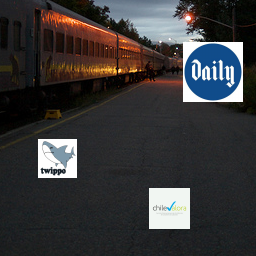} &
\includegraphics[width=0.12\linewidth]{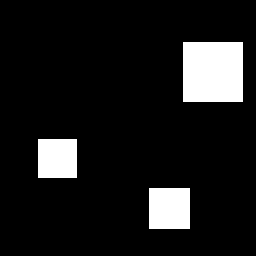} &
\includegraphics[width=0.12\linewidth]{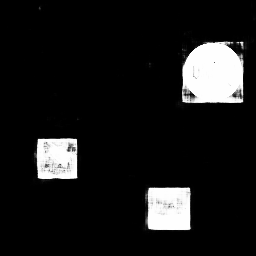} &
\includegraphics[width=0.12\linewidth]{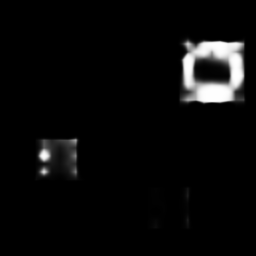} &
\includegraphics[width=0.12\linewidth]{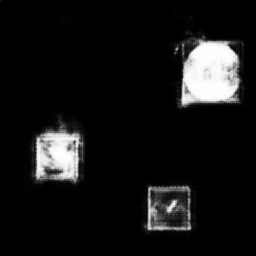} &
\includegraphics[width=0.12\linewidth]{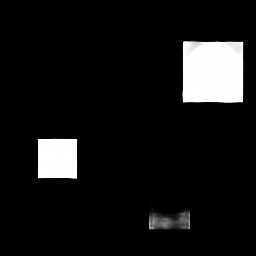} &
\includegraphics[width=0.12\linewidth]{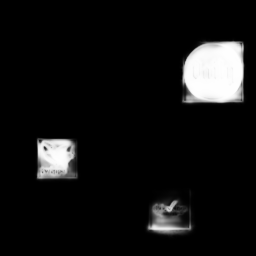} &
\includegraphics[width=0.12\linewidth]{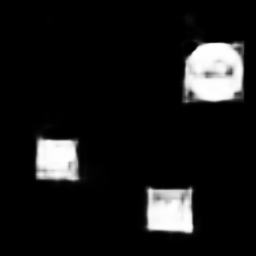} \\
\includegraphics[width=0.12\linewidth]{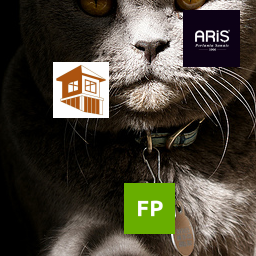} &
\includegraphics[width=0.12\linewidth]{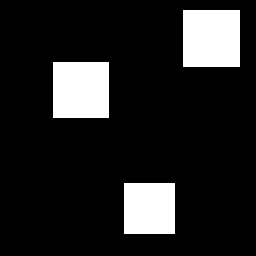} &
\includegraphics[width=0.12\linewidth]{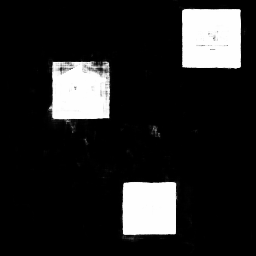} &
\includegraphics[width=0.12\linewidth]{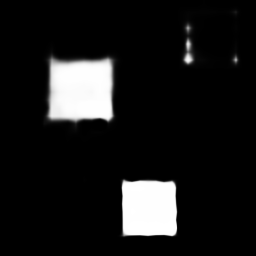} &
\includegraphics[width=0.12\linewidth]{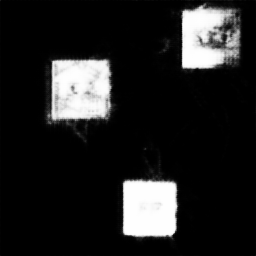} &
\includegraphics[width=0.12\linewidth]{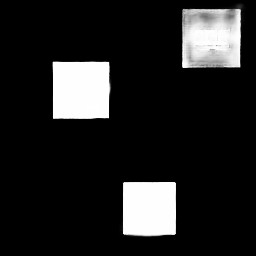} &
\includegraphics[width=0.12\linewidth]{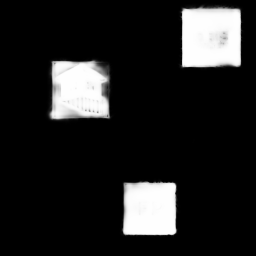} &
\includegraphics[width=0.12\linewidth]{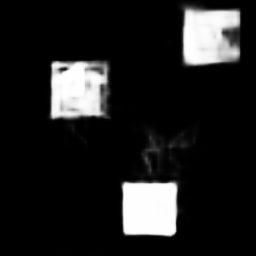} \\
\includegraphics[width=0.12\linewidth]{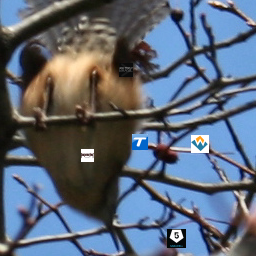} &
\includegraphics[width=0.12\linewidth]{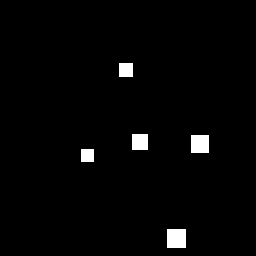} &
\includegraphics[width=0.12\linewidth]{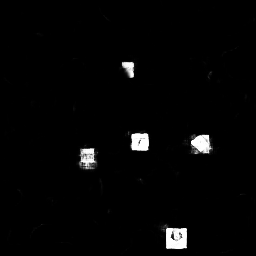} &
\includegraphics[width=0.12\linewidth]{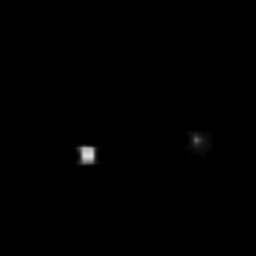} &
\includegraphics[width=0.12\linewidth]{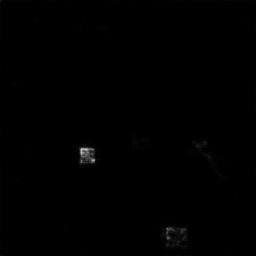} &
\includegraphics[width=0.12\linewidth]{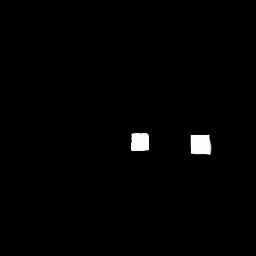} &
\includegraphics[width=0.12\linewidth]{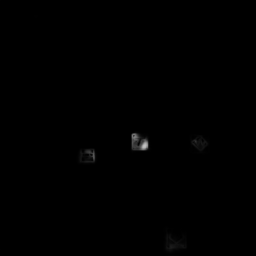} &
\includegraphics[width=0.12\linewidth]{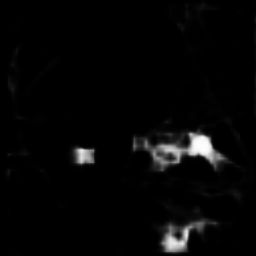} \\
\includegraphics[width=0.12\linewidth]{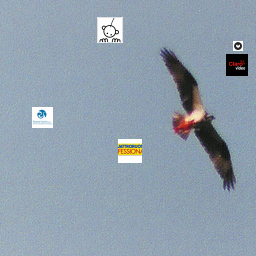} &
\includegraphics[width=0.12\linewidth]{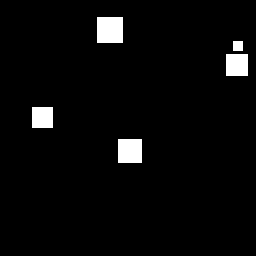} &
\includegraphics[width=0.12\linewidth]{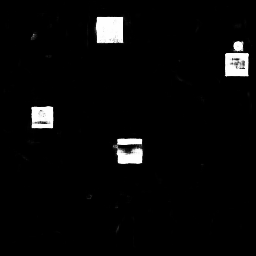} &
\includegraphics[width=0.12\linewidth]{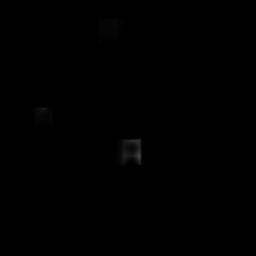} &
\includegraphics[width=0.12\linewidth]{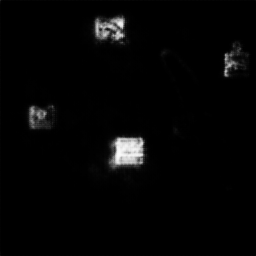} &
\includegraphics[width=0.12\linewidth]{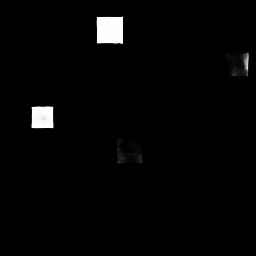} &
\includegraphics[width=0.12\linewidth]{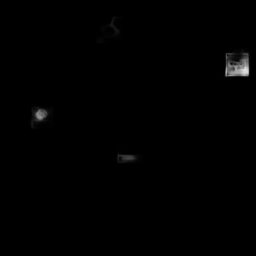} &
\includegraphics[width=0.12\linewidth]{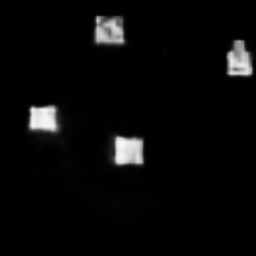} \\
\includegraphics[width=0.12\linewidth]{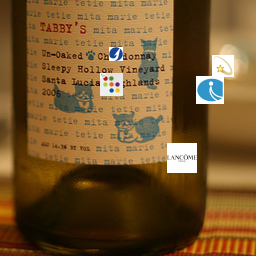} &
\includegraphics[width=0.12\linewidth]{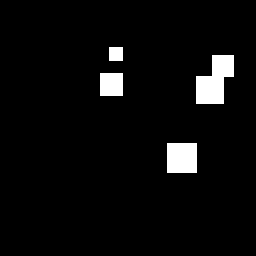} &
\includegraphics[width=0.12\linewidth]{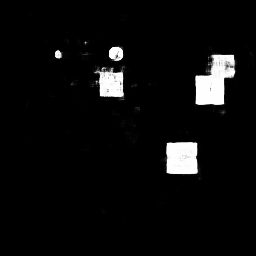} &
\includegraphics[width=0.12\linewidth]{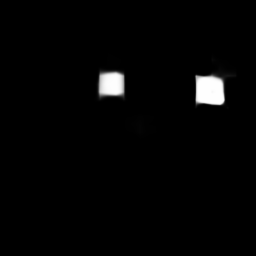} &
\includegraphics[width=0.12\linewidth]{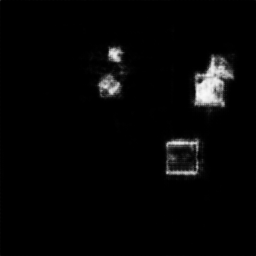} &
\includegraphics[width=0.12\linewidth]{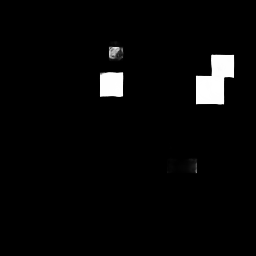} &
\includegraphics[width=0.12\linewidth]{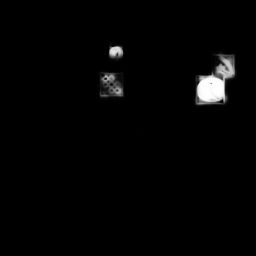} &
\includegraphics[width=0.12\linewidth]{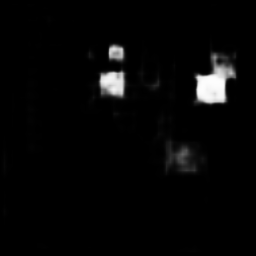} \\

& & & & & \\
(a) Image & (b) GT & (c) Ours & (d) DLV3 & (e) UNet & (f) BASNet & (g) PFANet & (h) SFPN \\
\end{tabular}
\vspace{2ex}
\caption{Visual comparison of the proposed method and the competing methods on logos}
\label{fig: test set visualization logo}
\end{figure*}

Additional results of the visual comparisons are shown in  Figure~\ref{fig: test set visualization stickers}, Figure~\ref{fig: test set visualization text}, Figure~\ref{fig: test set visualization 1}, and Figure~\ref{fig: test set visualization logo}. For each category, we show the perturbations of different sizes, and compare the performance with the competing methods.

\clearpage
\newpage
\section*{Real images downloaded from the internet}

\clearpage
\newpage
\section*{More Test Set Visual Comparisons (1): Stickers}

\clearpage
\newpage
\section*{More Test Set Visual Comparisons (2): Lines}

\clearpage
\newpage
\section*{More Test Set Visual Comparisons (3): Text}

\clearpage
\newpage
\section*{More Test Set Visual Comparisons (4): Logo}
\twocolumn

{\small
\bibliographystyle{ieee_fullname}
\bibliography{ms}
}

\end{document}